\documentclass[10pt,twocolumn,letterpaper]{article}

\usepackage[pagenumbers]{wacv} %

\usepackage{amssymb} %
\usepackage{xcolor} %
\usepackage{booktabs} %
\usepackage{multirow} %
\usepackage{graphicx} %

\makeatletter
\global\expandafter\let\csname ver@everyshi.sty\endcsname\@empty
\makeatother

\usepackage{tikz}
\usepackage{pgfplots}
\pgfplotsset{compat=1.16}
\usepackage{float}

\usepackage{amsmath}
\usepackage{url}

\usepackage{array} %
\usepackage{adjustbox} %

\newcommand{\cmark}{\textcolor{green!60!black}{$\checkmark$}}
\newcommand{\cross}{\textcolor{red!70!black}{$\times$}}

\newlength{\metricfigimgh}
\newcommand{\legdot}[1]{\textcolor[HTML]{#1}{\tikz[baseline=-0.6ex]\fill (0,0) circle (3.2pt);}}
\newcommand{\metriclegend}[1]{%
  \settoheight{\metricfigimgh}{\includegraphics[width=0.27\textwidth]{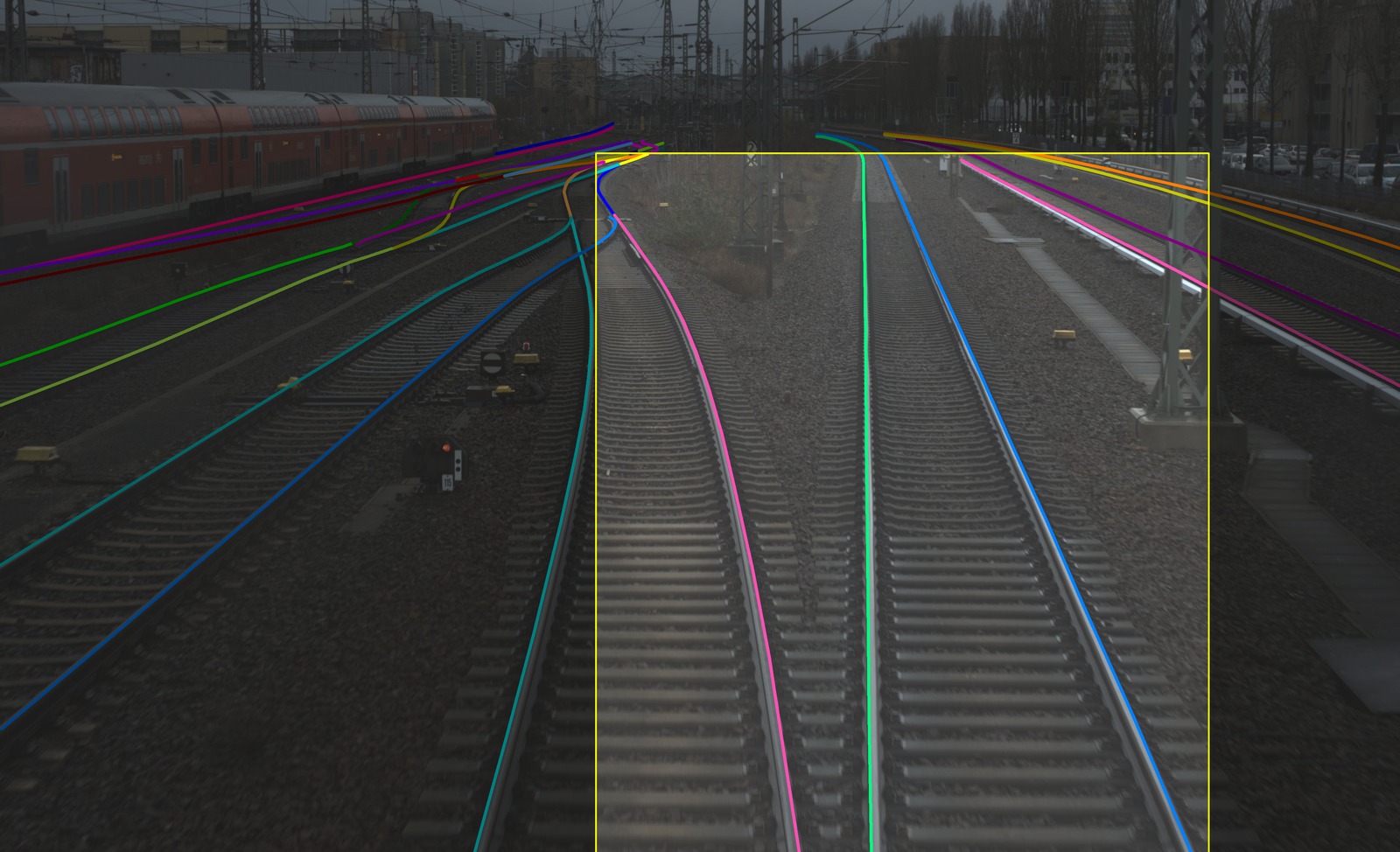}}%
  \smash{\rotatebox{90}{\parbox[b]{3\metricfigimgh}{\centering\small
    \renewcommand{\arraystretch}{1.2}%
    \begin{tabular}{@{}l@{\qquad}l@{}}
      \legdot{2bff00}~TP for LineAP only & \legdot{00a2ff}~TP for both metrics \\
      \legdot{ff04ea}~TP for #1 only & \legdot{ff0000}~FP for both metrics
    \end{tabular}}}}}

\definecolor{wacvblue}{rgb}{0.21,0.49,0.74}
\usepackage[pagebackref,breaklinks,colorlinks,allcolors=wacvblue]{hyperref}

\def\wacvPaperID{1990} %
\def\confName{WACV}
\def\confYear{2027}

\title{Railway Artificial Intelligence Learning Benchmark (RAIL-BENCH): \\ A Benchmark Suite for Perception in the Railway Domain}

\author{Annika B\"{a}tz$^{1}$ \qquad Pavel Klasek$^{2}$ \qquad Seo-Young Ham$^{2}$ \qquad Martin Lauer$^{1}$\\[4pt]
$^{1}$Karlsruhe Institute of Technology, Germany\\
$^{2}$German Centre for Rail Traffic Research at the Federal Railway Authority, Germany\\
{\tt\small \{annika.baetz, martin.lauer\}@kit.edu \quad \{klasekp, hams\}@dzsf.bund.de}
}

\begin{document}
\maketitle
\begingroup\renewcommand\thefootnote{}\footnotetext{Accepted at the IEEE/CVF Winter Conference on Applications of Computer Vision (WACV) 2027.}\endgroup
\begin{abstract}

Benchmark suites are crucial for advancing computer vision research, providing not only curated datasets but also standardized evaluation protocols for reproducible comparison. The railway domain lacks such resources, hindering progress in automated train operation. To address this gap, we present RAIL-BENCH, the first perception benchmark suite designed specifically for railways. RAIL-BENCH comprises five challenges -- rail track detection, object detection, vegetation segmentation, multi-object tracking, and monocular visual odometry -- each tailored to railway-specific characteristics and including a curated dataset from diverse real-world scenarios, evaluation metrics, and public scoreboards; we demonstrate the first three challenges in this paper.
We further introduce LineAP, a novel segment-based average precision metric for rail track detection that decouples geometric accuracy from instance-level grouping and employs a track-width-based distance threshold approximating a fixed real-world distance, addressing key limitations of existing line detection metrics.
Our benchmark suite is publicly available at \url{https://www.mrt.kit.edu/railbench/}.

\end{abstract}

\section{Introduction}
\label{sec:intro}

Automation of passenger and freight transport is increasingly important for future mobility. In railways, driverless operation has existed since the 1970s, mainly on systems fully separated from their surroundings, such as metros, where restricted track access and platform doors allow the system to safely assume clear tracks. Conventional rail infrastructure rarely meets these conditions: level crossings, open platforms, and mixed-traffic urban rail introduce much higher complexity, requiring sensors that can detect obstacles on or near the track, interpret passenger behavior, recognize vegetation encroaching on the clearance gauge, identify blocked tracks, and detect railway signals and their aspects.

\begin{figure}[t]
  \centering
  \setlength{\tabcolsep}{0pt}
  \renewcommand{\arraystretch}{0}
  \begin{tabular}{cc}
    \includegraphics[width=0.24\textwidth,keepaspectratio]{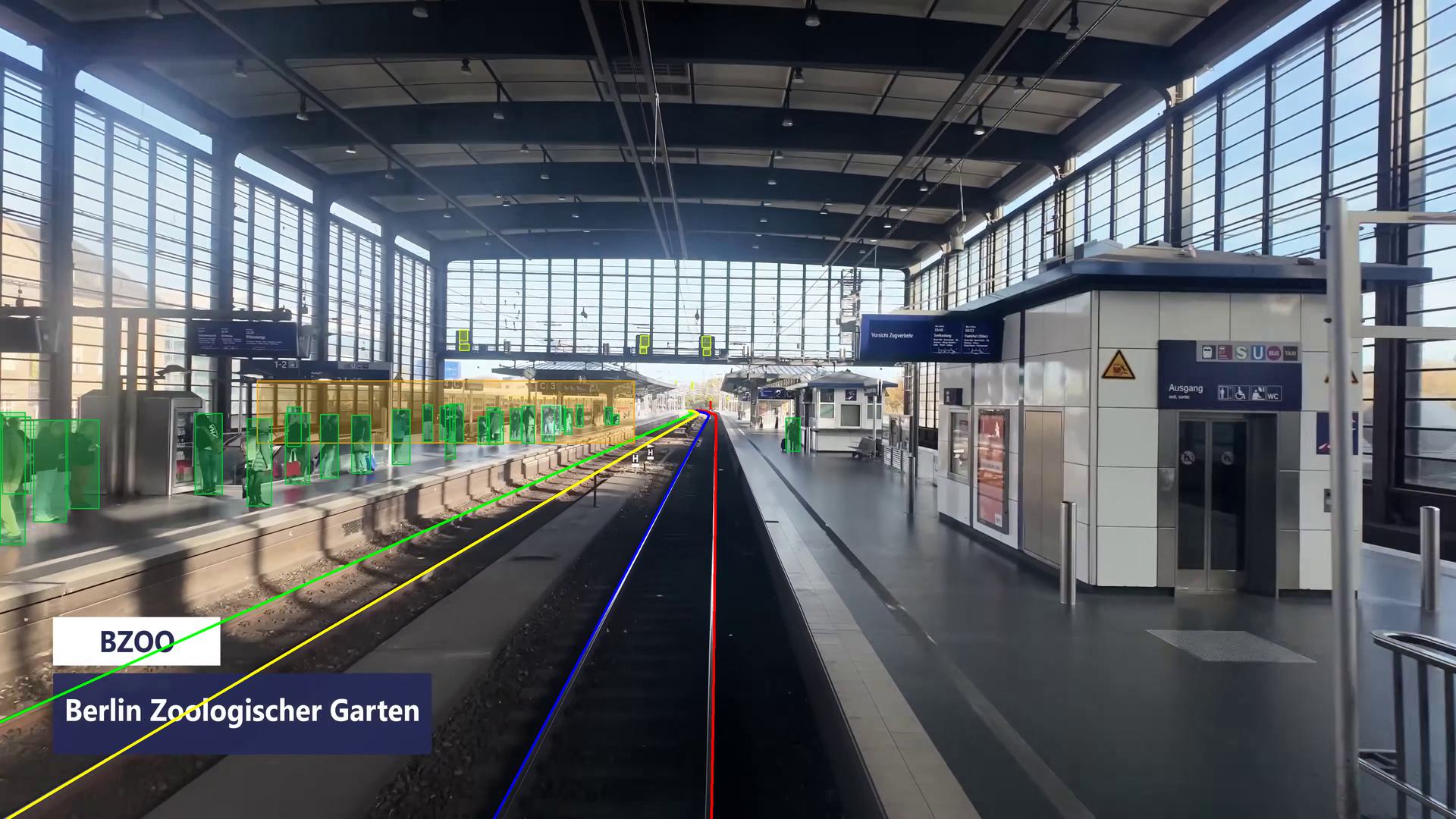} &
    \includegraphics[width=0.24\textwidth,keepaspectratio]{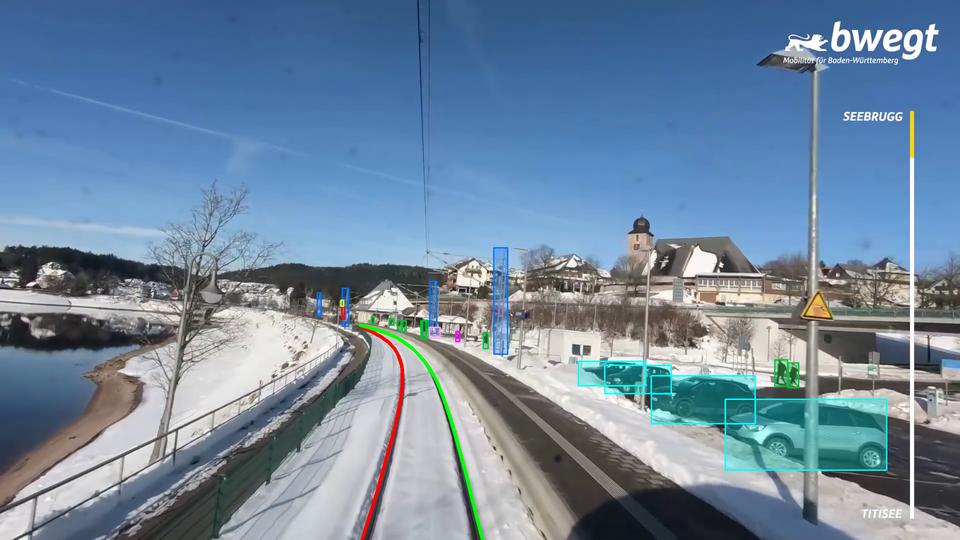} \\
    \includegraphics[width=0.24\textwidth,keepaspectratio]{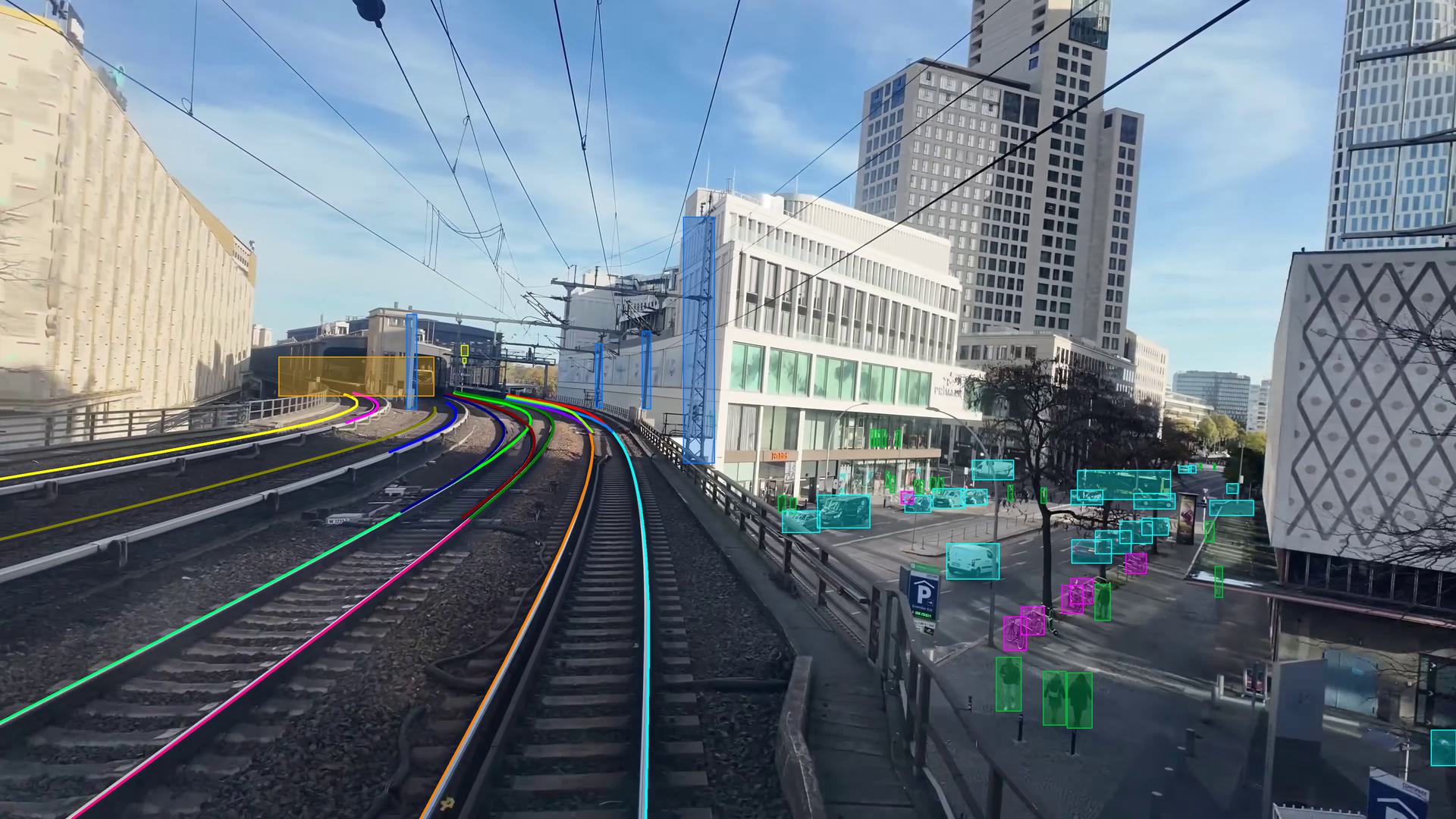} &
    \includegraphics[width=0.24\textwidth,keepaspectratio]{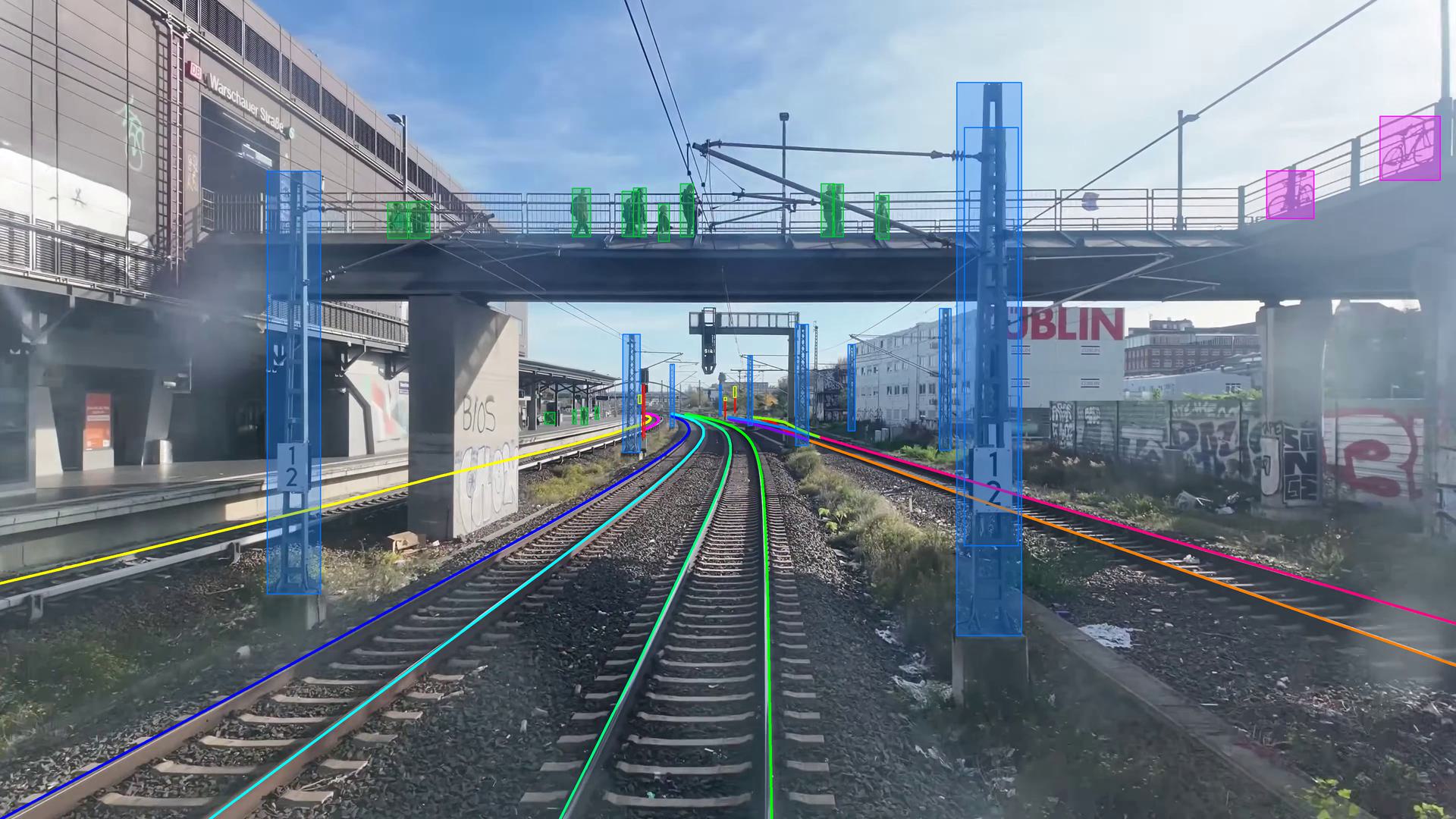} \\
    \includegraphics[width=0.24\textwidth,keepaspectratio]{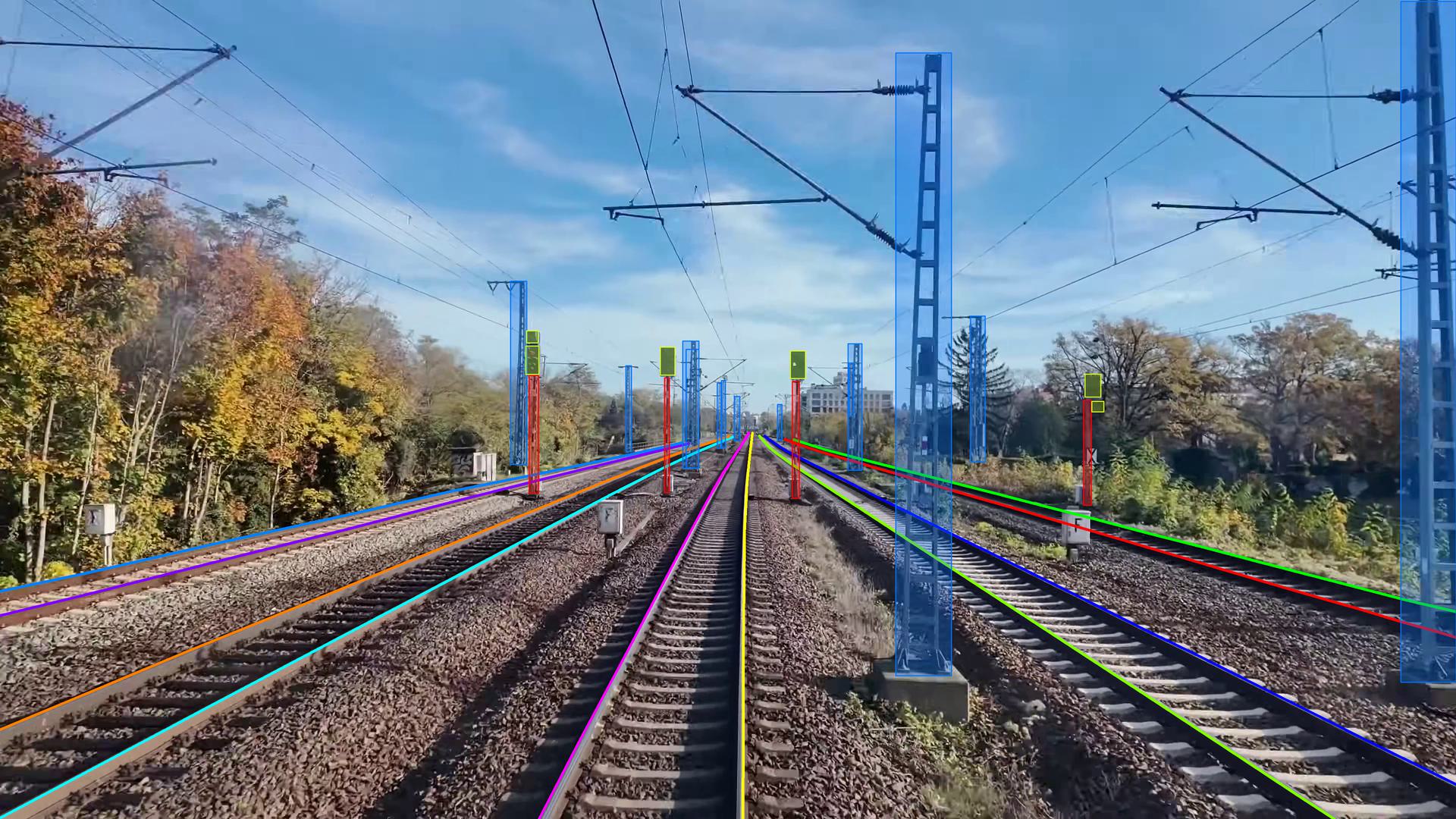} &
    \includegraphics[width=0.24\textwidth,keepaspectratio]{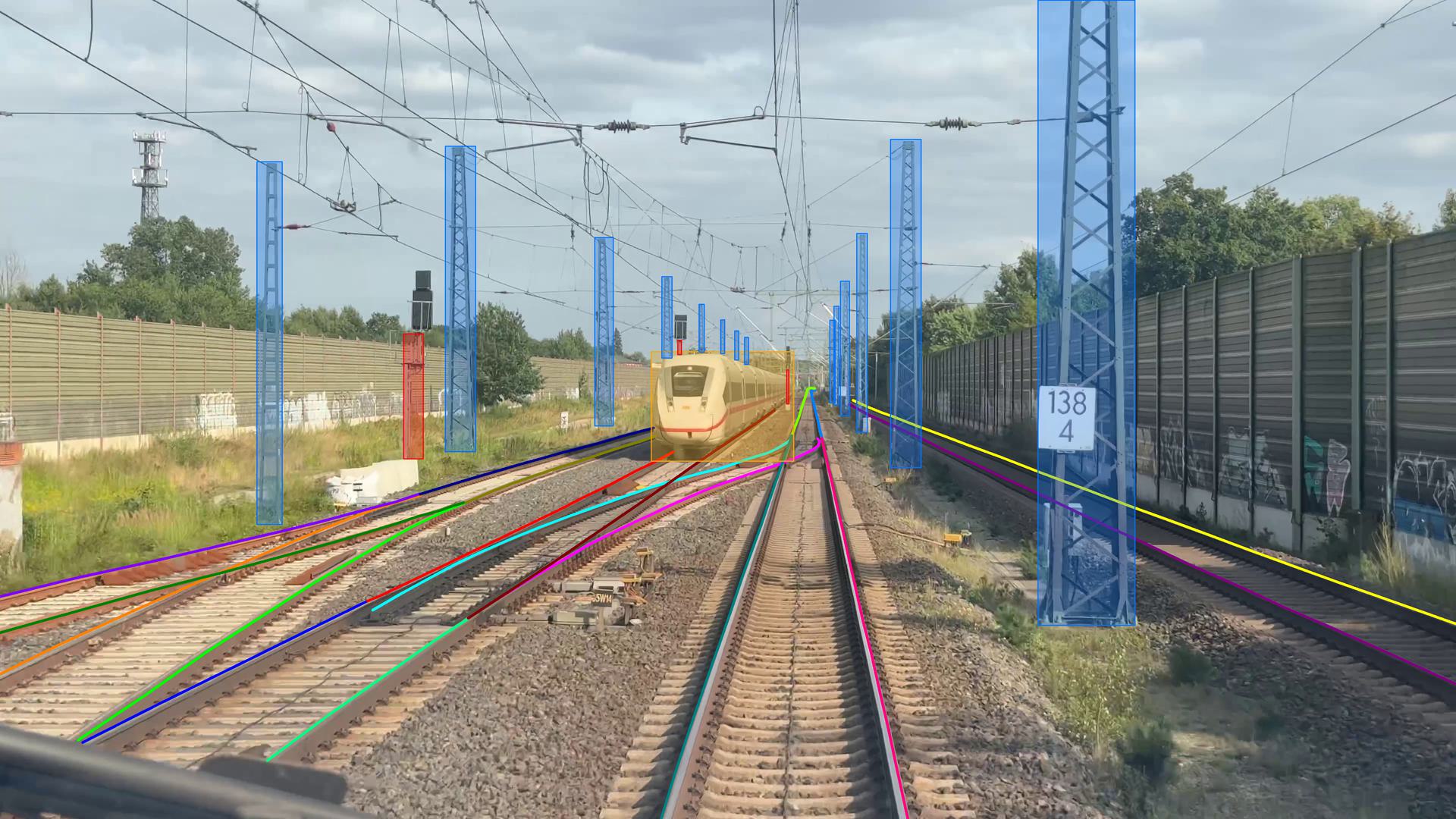} \\
    \includegraphics[width=0.24\textwidth,keepaspectratio]{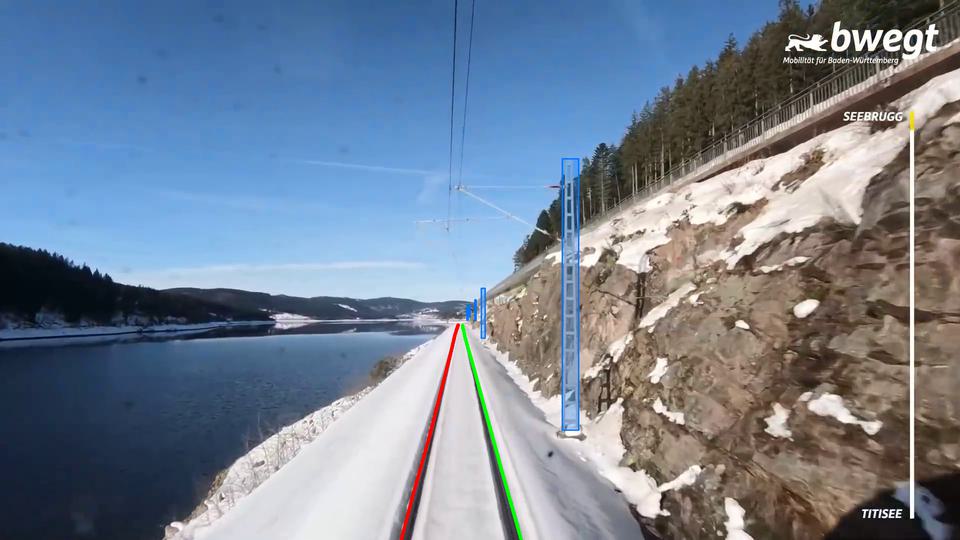} &
    \includegraphics[width=0.24\textwidth,keepaspectratio]{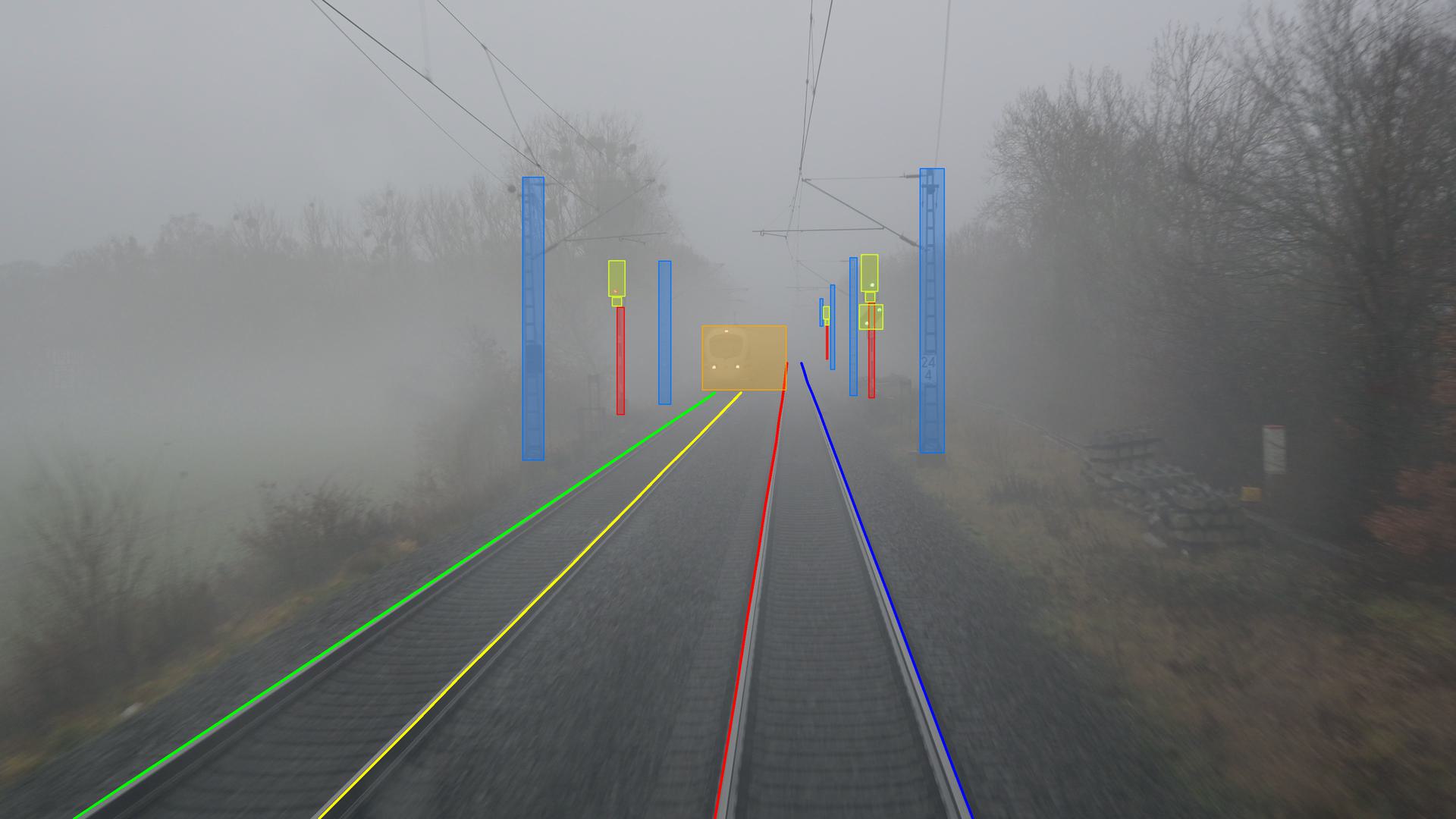} \\
    \includegraphics[width=0.24\textwidth,keepaspectratio]{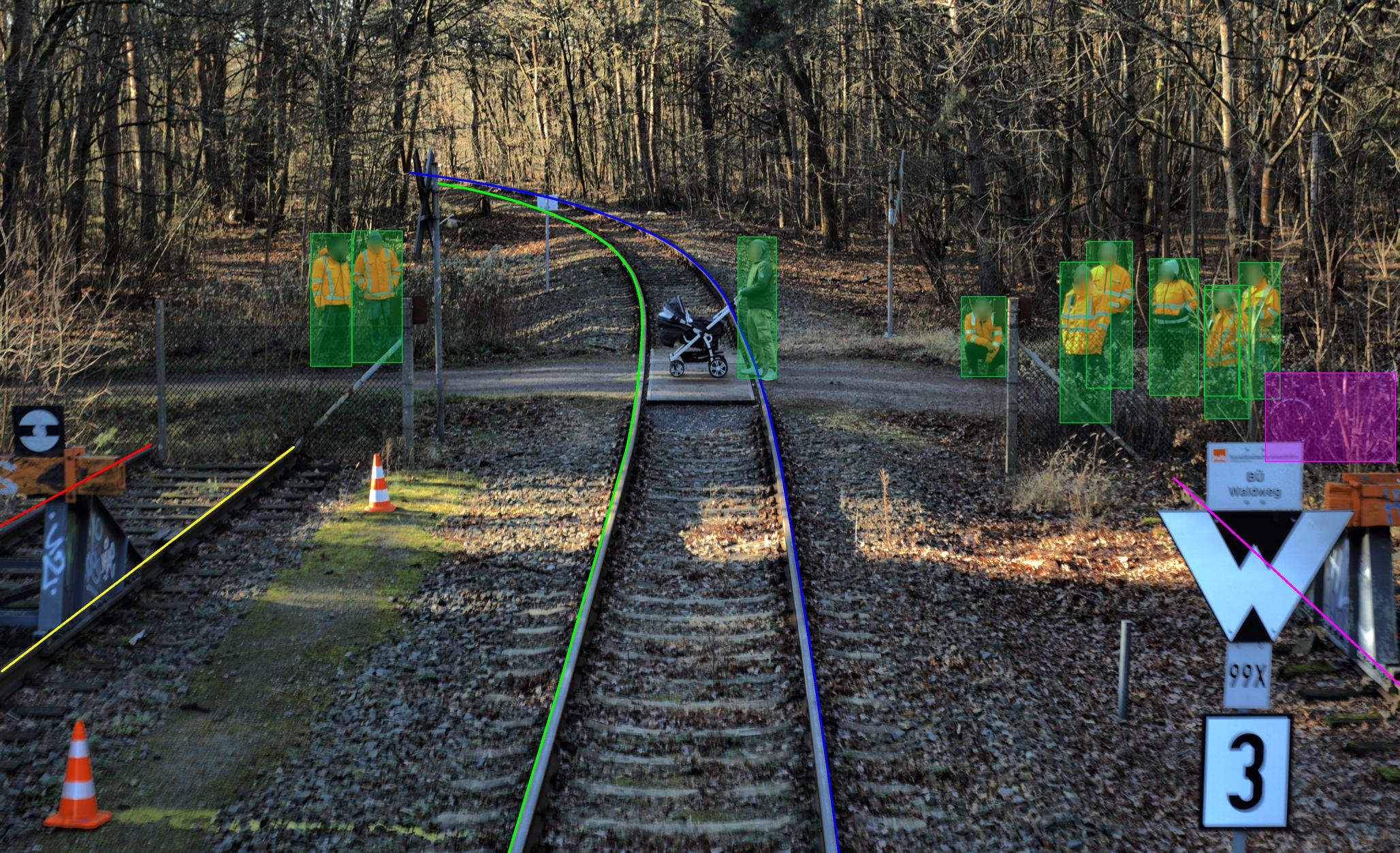} &
    \includegraphics[width=0.24\textwidth,keepaspectratio]{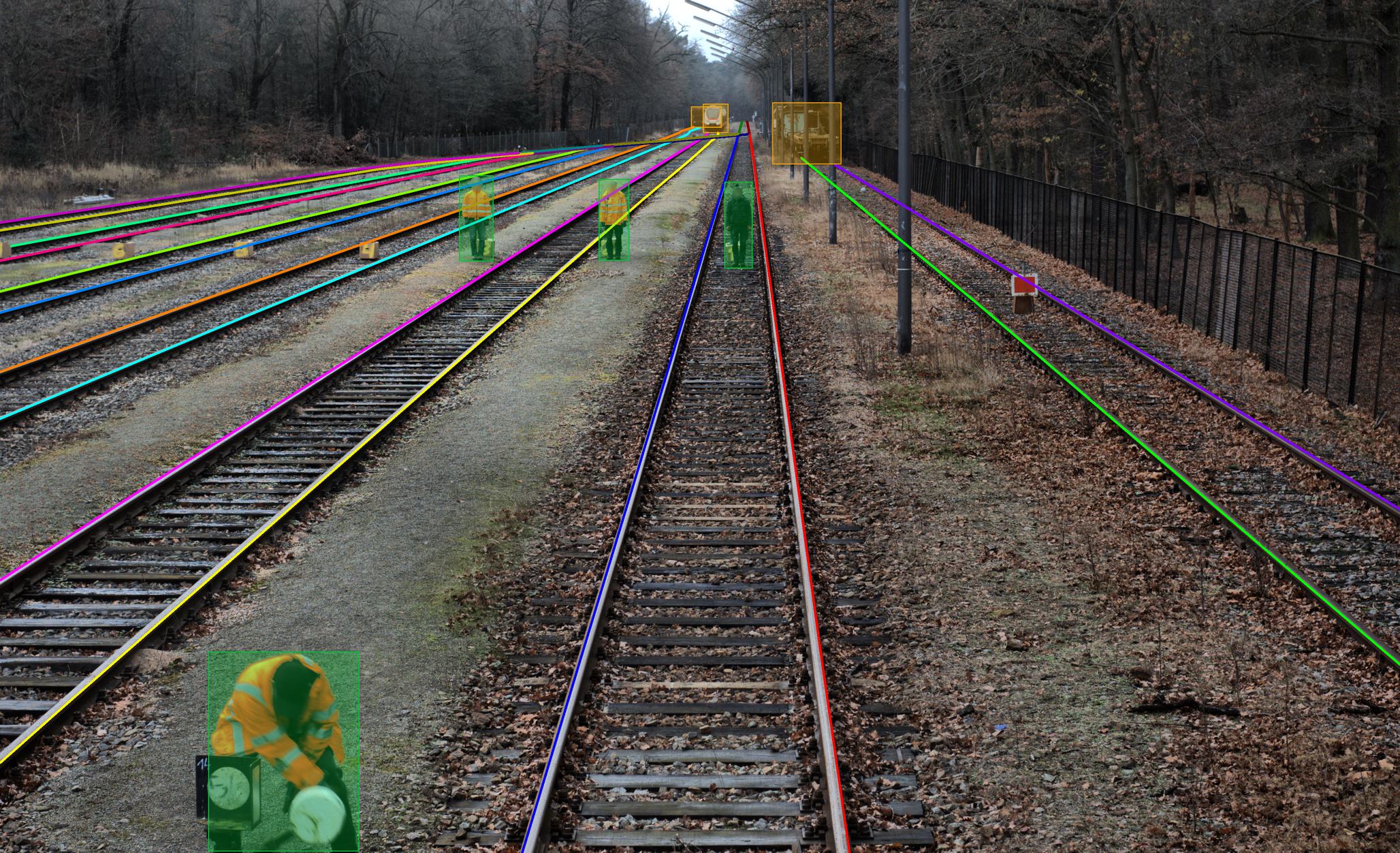}
  \end{tabular}
  \caption{Annotated example images from the RAIL-BENCH Object+Rail dataset displaying diverse scenarios: train stations, urban, suburban/high-speed, rural and staged. Objects are colored as displayed in figure~\ref{fig:objdet-class-dist} and rail colors are randomly assigned. Best viewed in color.}
  \label{fig:anns_objrail}
\end{figure}

Addressing these perception challenges is a fundamental prerequisite for safe, fully automated train operation~\cite{tagiew2023sensor}. State-of-the-art perception systems primarily rely on camera-based imaging, combined with machine learning models, owing to its ability to acquire rich, high-resolution visual data.

In the automotive sector, which faces related perception challenges, model development has been substantially advanced by public benchmark suites such as KITTI~\cite{geiger_arewe_2012}, Waymo~\cite{Sun_2020_CVPR}, and nuScenes~\cite{Caesar_2020_CVPR}. These provide annotated datasets and standardized evaluation protocols that make empirical research reproducible and comparable while lowering the barrier to entry for new contributors -- considerably accelerating progress in the field.

Comparable benchmarks are still missing for the railway domain: publicly available datasets are limited in size and representativeness, and domain-specific evaluation protocols remain poorly defined, hindering unbiased comparison and reproducible evaluation.

To fill this gap, we present RAIL-BENCH, the first environmental perception benchmark suite for the railway domain, providing datasets and evaluation protocols for five challenges: rail track detection, object detection, vegetation segmentation, multi-object tracking, and monocular visual odometry. In this paper, we focus on rail track detection, object detection, and vegetation segmentation, as these exhibit the largest domain gap to automotive perception; multi-object tracking and monocular visual odometry are described in the supplementary material.

Our contributions are outlined as follows:
\begin{itemize}
\item We present the first benchmark suite for environmental perception in the railway domain, comprising five challenges tailored to the specific structural and operational characteristics of railway environments. 
\item We present the novel LineAP metric for evaluating polyline predictions, e.g., for rail tracks. Unlike existing metrics, LineAP handles partial detections, penalizes over-detection consistently, and incorporates joint distance and orientation thresholds, further customized for railway applications via a variable distance threshold that approximates consistent real-world distances.
\end{itemize}

\begin{table*}[t]
\centering
\caption{Public railway object and rail detection datasets with real-world RGB sensor data recorded from the driver's point of view.}
\label{tab:datasets}
{\scriptsize
\setlength{\tabcolsep}{3pt}
\renewcommand{\arraystretch}{1.05}
\resizebox{0.95\textwidth}{!}{%
\begin{tabular}{l*{14}{c}r c@{}}
\toprule
 & \multicolumn{8}{c}{Annotations} & \multicolumn{5}{c}{Scenarios} &  \multicolumn{3}{c}{Data} \\
 \cmidrule(lr){2-9}\cmidrule(lr){10-14}\cmidrule(lr){15-17}

 & \multicolumn{8}{c}{Classes with Instance-Based Dense Annotations} 
 & \multirow{5}{*}{\rotatebox[origin=c]{90}{\shortstack{Train\\[-0.3ex]Stations}}} & \multirow{5}{*}{\rotatebox[origin=c]{90}{Urban}} & \multirow{5}{*}{\rotatebox[origin=c]{90}{\shortstack{Suburban/\\[-0.3ex]High-speed}}} & \multirow{5}{*}{\rotatebox[origin=c]{90}{Rural}} & \multirow{5}{*}{\rotatebox[origin=c]{90}{Staged}} & \multirow{5}{*}{\shortstack{Image\\[-0.3ex]Resolution\\[-0.3ex]in MPx\\[-0.3ex](min - max)}} & \multirow{5}{*}{\# Images} & \multirow{5}{*}{\shortstack{Sparse\\[-0.3ex]Image\\[-0.3ex]Sampling}} \\

 & \multicolumn{4}{c}{Railway Infrastructure} & \multicolumn{4}{c}{Potential Obstacles} \\
\cmidrule(lr){2-5}\cmidrule(lr){6-9}
 & \multirow{2}{*}{Rail} & \multirow{2}{*}{Signal} & \multirow{2}{*}{\shortstack{Signal\\[-0.3ex]Pole}} & \multirow{2}{*}{\shortstack{Catenary\\[-0.3ex]Pole}} & \multirow{2}{*}{Train} & \multirow{2}{*}{Person} & \multirow{2}{*}{\shortstack{Road\\[-0.3ex]Vehicle}} & \multirow{2}{*}{\shortstack{Bicycle/\\[-0.3ex]Cyclist}} \\
 &  &  &  &  &  &  &  &  \\

\midrule
RailSem19~\cite{zendel_railsem19dataset_2019} & \cmark & \cross & \cross & \cross & \cross & \cross & \cross & \cross & \cmark & \cmark & \cmark & \cmark & \cross & $2.07$ & $8{,}500$ & \cmark \\
RailSet$^{*1}$~\cite{zouaoui_railsetunique_2022} & \cmark & \cross & \cross & \cross & \cross & \cross & \cross & \cross & \cmark & \cmark & \cmark & \cmark & \cross & $0.41 - 3.69$ & $729$ & \cmark \\
Rail-DB~\cite{li_raildetection_2022} & \cmark & \cross & \cross & \cross & \cross & \cross & \cross & \cross & \cross & \cross & \cmark & \cross & \cmark & $0.92 - 8.29$ & $7{,}432$ & \cross \\
L4R\_NLB~\cite{hiebert_labels4railsrailway_2025} & \cmark & \cross & \cross & \cross & \cross & \cross & \cross & \cross & \cmark & \cross & \cmark & \cmark & \cross & $2.07$ & $10{,}253$ & \cmark \\
\cmidrule(lr){1-17}
FRSign~\cite{harb_frsignlargescale_2020} & \cross & \cmark & \cross & \cross & \cross & \cross & \cross & \cross & \cmark & \cmark & \cmark & \cmark & \cross & $2.3-3.15$ & $105{,}352$ & \cross \\
GERALD~\cite{leibner_geraldnovel_2023} & \cross & \cmark & \cross & \cross & \cross & \cross & \cross & \cross & \cmark & \cmark & \cmark & \cmark & \cross & $0.92 - 2.07$ & $5{,}000$ & \cmark \\
\cmidrule(lr){1-17}
RailGoerl24~\cite{tagiew_railgoerl24gorlitz_2025a} & \cross & \cross & \cross & \cross & \cross & \cmark & \cross & \cmark & \cross & \cross & \cross & \cross & \cmark & $2.07$ & $12{,}205$ & \cross \\
ESRORAD$^{*2}$~\cite{khemmar_roadrailway_2022} & \cross & \cross & \cross & \cross & \cross & \cmark & \cmark & \cmark & \cross & \cmark & \cross & \cross & \cross & $2.07$ & $\sim100{,}000$ & \cross \\
MRSI$^{*3}$~\cite{chen_mrsimultimodal_2022} & \cross & \cross & \cross & \cross & \cross & \cmark & \cmark & \cmark & \cross & \cmark & \cross & \cross & \cmark & $0.52$ & $2{,}983$ & \cross \\
\cmidrule(lr){1-17}
OSDaR23$^{*4}$~\cite{tagiew_osdar23open_2023} & \cmark & \cmark & \cmark & \cmark & \cmark & \cmark & \cmark & \cmark & \cmark & \cmark & \cmark & \cross & \cmark & $3.94 - 10.3$ & $1{,}534$ m-frames$^{*5}$ & \cross \\
OSDaR26$^{*4}$~\cite{osdar26} & \cmark & \cmark & \cmark & \cmark & \cmark & \cmark & \cmark & \cmark & \cmark & \cmark & \cmark & \cmark & \cmark & $3.94 - 19.96$ & $15{,}646$ m-frames$^{*5}$ & \cross \\
\midrule

RAIL-BENCH Object+Rail & \cmark & \cmark & \cmark & \cmark & \cmark & \cmark & \cmark & \cmark & \cmark & \cmark & \cmark & \cmark & \cmark & $2.07 - 19.96$ & $2{,}500$ & \cmark \\
\midrule
\addlinespace[2pt]
\multicolumn{17}{l}{\scriptsize $^{*1}$ \textit{Published real-world subset of RailSet dataset.}} \\
\multicolumn{17}{l}{\scriptsize $^{*2}$ \textit{Only a small subset of these data shows tram scenes.}} \\
\multicolumn{17}{l}{\scriptsize $^{*3}$ \textit{Subset of the MRSI dataset with instance-based annotations.}} \\
\multicolumn{17}{l}{\scriptsize $^{*4}$ \textit{Our RAIL-BENCH Object+Rail dataset includes specifically sampled images from OSDaR23 and OSDaR26.}} \\
\multicolumn{17}{l}{\scriptsize $^{*5}$ \textit{M-frame stands for multi-sensor frame, which is the collection of several sensor frames.}} \\
\bottomrule
\end{tabular}%
}
}
\end{table*}

\section{Related work}

Several datasets containing camera-based perception data from the railway domain have been published in recent years. This survey focuses on datasets that are publicly available without restrictions and include annotations, e.g., for tracks or objects. Table~\ref{tab:datasets} provides a synopsis.

RailSem19~\cite{zendel_railsem19dataset_2019} provides images extracted from 530 video sequences from the train driver's perspective, exhibiting substantial inter-image variability across a wide range of operational and environmental scenarios. Annotations include geometric shapes (bounding boxes, polylines, polygons) for 21 object classes and dense semantic masks for 20 classes. Except for the rail class, geometric annotations are highly sparse, with only a few objects labeled per image.

RailSet~\cite{zouaoui_railsetunique_2022}, Rail-DB~\cite{li_raildetection_2022}, and L4R\_NLB~\cite{hiebert_labels4railsrailway_2025} provide railway track annotations. L4R\_NLB is the largest and most diverse of the three but does not address complex urban scenarios.

FRSign~\cite{harb_frsignlargescale_2020} and GERALD~\cite{leibner_geraldnovel_2023} provide railway signal annotations for France and Germany, respectively. Although FRSign nominally comprises $105{,}352$ images, its effective size is substantially smaller due to high inter-frame redundancy.

Some datasets focus on dynamic objects such as persons and traffic participants. ESRORAD~\cite{khemmar_roadrailway_2022} provides images, LiDAR data, and 3D annotations for persons, road vehicles, and bicycles; recorded from a car, it contains only a few railway-specific scenarios. MRSI~\cite{chen_mrsimultimodal_2022} collects visible- and infrared-spectrum images, with different subsets annotated by semantic masks or geometric shapes across varying class categories. Considering only the instance-based RGB annotations, MRSI comprises 2983 images labeled for persons, road vehicles, and cyclists, with high inter-frame similarity and low environmental variance. RailGoerl24~\cite{tagiew_railgoerl24gorlitz_2025a} stages potentially dangerous scenarios involving persons and bicycles, making it unique among these datasets, though all images were recorded at the same location, limiting environmental diversity. The RAilWay PEdestrian Dataset (RAWPED)~\cite{toprak_conditionalweighted_2020} contains $22{,}302$ annotated images for the person class. However, the annotations are systematically incomplete across all $135$ scenes. Consequently, this dataset is not included in Table~\ref{tab:datasets}.

OSDaR23~\cite{tagiew_osdar23open_2023} is the first multi-sensor railway dataset, containing RGB and infrared camera, LiDAR, and radar data, with 20 object classes annotated across all sensors via geometric shapes. It comprises 45 sequences totaling $1{,}534$ recording time steps, including complex station scenes and staged dangerous situations. The same authors constructed OSDaR26~\cite{osdar26}, a similarly recorded and annotated multi-sensor dataset that expands the number and duration of sequences.

To the best of our knowledge, RailSem19 is the only public dataset providing vegetation annotations for the railway domain (see Table~\ref{tab:vegetation}). Annotation quality is comparatively low, and vegetation directly on the track is not labeled as such, since these regions are preferentially assigned track-specific classes. Given the particular relevance of reliable vegetation detection in the rail area, these annotations are insufficient for training a robust segmentation model.

\begin{table}[t]
\centering
\caption{Public railway datasets for vegetation segmentation.}
{\scriptsize
\renewcommand{\arraystretch}{1.1}
\setlength{\tabcolsep}{3pt}
\resizebox{0.95\columnwidth}{!}{%
\begin{tabular}{@{}l*{5}{c}@{}}
\toprule
 & \multicolumn{3}{c}{Annotated Vegetation} & \multirow{2}{*}{\# Images} \\
\cmidrule(lr){2-4}
 & High Growing & Low Growing & On Tracks & \\
\midrule
RailSem19~\cite{zendel_railsem19dataset_2019} & \cmark & \cmark & \cross & $8{,}500$ \\
\midrule
RAIL-BENCH Vegetation$^{*1}$ & \cmark & \cmark & \cmark & $740$ \\
\midrule
\addlinespace[2pt]
\multicolumn{5}{l}{{\scriptsize $^{*1}$ \textit{Contains frames sampled from OSDaR23 and OSDaR26.}}} \\
\bottomrule
\end{tabular}%
}
}

\label{tab:vegetation}
\end{table}

None of these datasets provide the specific challenges and evaluation metrics of a benchmark suite. Postprocessing, filtering, selecting error-free subsets, and splitting into training and test sets is left to the user, making empirical comparisons between approaches difficult.

\section{Design of the Benchmark Suite}

\subsection{Basic Concept}

The concept of the RAIL-BENCH benchmark suite was inspired by KITTI~\cite{geiger_arewe_2012} and other benchmarks~\cite{Caesar_2020_CVPR, Sun_2020_CVPR, cordts_cityscapesdataset_2016, Argoverse2}. We identified five core perception challenges for automated train operation, three of which are covered in this paper.

For each of the three challenges, RAIL-BENCH provides an image dataset split into training, validation, and test sets. While the training and validation data come with ground-truth, the ground-truth for the held-out test set is not published. RAIL-BENCH provides an evaluation protocol for each challenge and allows users to upload their predictions on the test data, which are then automatically evaluated. Public scoreboards let users compare their results against other approaches.

The three RAIL-BENCH challenges discussed here are:

\textbf{RAIL-BENCH Object}: the recognition of objects belonging to one of seven categories: train, signal, signal pole, catenary pole, road vehicle, bicycle, and person. Objects must be identified by specifying a corresponding bounding box together with an associated class label.

\textbf{RAIL-BENCH Rail:} the recognition of rail tracks. As rail tracks are elongated and poorly suited to bounding-box description, we model them as polylines marking the left and right rail. 

\textbf{RAIL-BENCH Vegetation:} the recognition of vegetation in and around the track area. Vegetation encroaching on the structural clearance gauge poses a safety-critical hazard. The task is formulated as a semantic segmentation problem, requiring pixel-wise classification into two categories: low-growing vegetation (e.g., grasses and herbs) and high-growing vegetation (e.g., trees and shrubs).

Information on the other two RAIL-BENCH challenges -- RAIL-BENCH Tracking and RAIL-BENCH Odometry -- can be found in the supplements.

\begin{figure}[t]
  \centering
  \includegraphics[width=0.9\linewidth,height=\textheight,keepaspectratio]{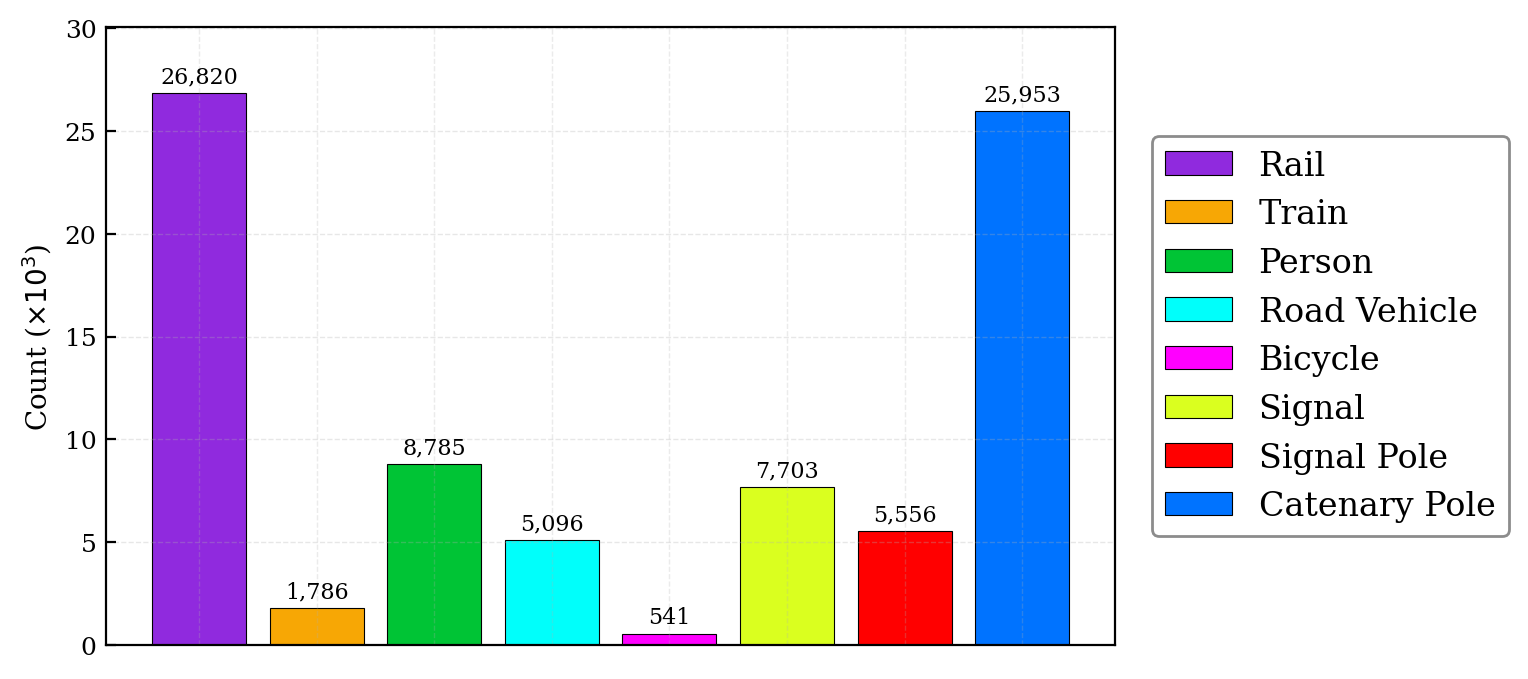}
  \caption{Class distribution in the RAIL-BENCH Object+Rail dataset. All annotations flagged as \textit{iscrowd} and \textit{ignore} are excluded.}
  \label{fig:objdet-class-dist}
\end{figure}

\subsection{Data Collection}

The datasets comprise real-world RGB images gathered from heterogeneous sources. The authors of OSDaR23 and OSDaR26 granted us permission to fully utilize their datasets and also provided additional unpublished recordings. Several YouTube channel owners further granted us permission to use their videos.

For RAIL-BENCH Object+Rail and RAIL-BENCH Vegetation, video frames were selected manually to maximize scenario diversity with respect to the intended task. In total, $2{,}500$ frames were extracted for RAIL-BENCH Object+Rail and $740$ frames for RAIL-BENCH Vegetation.

\subsection{Data Annotation and Preprocessing}\label{sec:annotation}

The annotations of OSDaR23 are public, and we additionally received preliminary annotations for OSDaR26. Depending on the task-specific labeling policy, we kept, adopted, extended, or replaced the given annotations. All other data was labeled from scratch. For all annotation jobs, we used the publicly available Computer Vision Annotation Tool (CVAT)~\cite{cvat}.

For railway tracks, we label the outer rail head of both the left and right rail using polylines. At turnouts, the polylines terminate at the horizontal line connecting the two tips of the switch blades. Polylines are continued across occlusions, provided that the rail position can still be inferred reliably; otherwise, ignore regions are added to mark the affected area and exclude it from evaluation.

For the object detection task, we annotated axis-aligned bounding boxes. The \textit{signal} class contains front-facing light signals but excludes backward-facing and static signals. The categories \textit{person}, \textit{road vehicle}, and \textit{bicycle} can occur in dense groups, which are particularly difficult to annotate at the instance level from large distances; in such cases, objects may be annotated collectively as a single entity and flagged as \textit{iscrowd}. Each bounding box is additionally labeled with an occlusion level, and instances can be marked \textit{ignore} when the annotator is uncertain about the underlying object category.

Vegetation masks were generated in two steps. First, we employed SegFormer-B1~\cite{xie_segformersimple_2021}, pretrained on Cityscapes~\cite{cordts_cityscapesdataset_2016} and finetuned on RailSem19~\cite{zendel_railsem19dataset_2019}, to obtain initial vegetation labels. Second, these masks were manually refined to ensure full compliance with our labeling policy.

\subsection{Dataset Analysis}

The assembled RAIL-BENCH Object+Rail dataset covers diverse railway scenarios and weather conditions (see Fig.~\ref{fig:anns_objrail}). The predominant object categories are \textit{rail} and \textit{catenary pole}, each with approximately $26{,}000$ annotations (see Fig.~\ref{fig:objdet-class-dist}). The \textit{person} category comprises $8{,}785$ annotated instances, averaging $3.5$ instances per image, with a maximum of $71$ annotated persons in a single image.

The RAIL-BENCH Vegetation dataset has around $28\%$ of pixels belonging to the high-growing vegetation class -- almost three times the $11\%$ share of the low-growing class. Example images can be found in the supplementary material.

\section{Evaluation}
\label{sec:eval}

We use a geographical split~\cite{lilja2024localization} to divide both the RAIL-BENCH Object+Rail and Vegetation datasets into non-overlapping training, validation, and test sets ($1{,}500$/$500$/$500$ and $520$/$110$/$110$ images, respectively). In the following, we describe the evaluation metrics used for object detection and rail track detection. For vegetation segmentation, we use the standard Intersection-over-Union (IoU) score for each individual class as well as a combined vegetation class.

\subsection{Evaluation of Object Detections}
\label{sec:mAP}

For RAIL-BENCH Object we adopt the implementation of Average Precision (AP) from the COCO detection challenge~\cite{coco_detection_challenge_codalab, padilla_surveyperformance_2020}. As is common practice, we compute AP over a set of varying IoU thresholds and report the mean value across all thresholds. Furthermore, we use the size and occlusion attributes of each ground-truth bounding box to define three levels of difficulty: \textit{easy}, \textit{moderate}, and \textit{hard}. Details on the difficulty levels are provided in the supplementary material.

\subsection{Evaluation of Rail Track Detections}

For line-based detection tasks like rail or road lane recognition, several evaluation metrics have been suggested, including the metrics used in the TuSimple~\cite{yoo2020endtoendlanemarkerdetection} and CULane~\cite{pan2026spatialdeepspatialcnn} benchmarks, as well as the Chamfer Distance (CD) integrated into an AP framework~\cite{li_hdmapnetonline_2022}, which we refer to as ChamferAP.

The TuSimple metric samples both the predicted and ground-truth lines uniformly at fixed vertical positions in the image. A predicted line is treated as a correct detection if a sufficient fraction of sampled points lies close to sample points from the ground-truth line. While the metric works well for vertical lines, it fails for horizontal ones due to the fixed vertical sampling grid. Furthermore, predictions without nearby ground-truth lines are completely ignored.

The CULane metric transforms each ground-truth and predicted line into a polygon of fixed width and calculates the IoU afterward. It penalizes both false positives (FP) and false negatives (FN). A shortcoming of the CULane metric is that local deviations along a predicted line are absorbed as long as they remain within the polygon boundary; hence, an oscillating prediction may achieve the same IoU as a smooth one.

ChamferAP was developed for evaluating HD map construction models in 2D world coordinates. It provides a more geometrically meaningful distance measure: a fixed number of points are sampled along both predicted and ground-truth lines, and their geometric proximity is evaluated directly using the CD. Moreover, it benefits from the general AP framework, such as integrating prediction confidence into the evaluation process.

Although ChamferAP offers various advantages over the other two metrics, all three metrics share a common shortcoming. A line detector must solve two sub-tasks: (i) the precise geometric localization of lines, and (ii) the correct grouping of detections into individual line instances that correspond to ground-truth definitions. The metrics described so far try to evaluate both tasks simultaneously. As a result, they fail to adequately account for several common cases: (a) partially correct predictions, (b) incomplete predictions, and (c) cases where a predicted instance corresponds to several ground-truth instances or vice versa. In all three cases, even if all or part of the prediction is highly accurate, the metric might disregard the entire prediction as a FP. To address these shortcomings, we propose LineAP, a new metric that evaluates geometric accuracy of predictions without requiring segments to be correctly grouped into complete line instances.

\begin{figure}[t]
    \centering
    \resizebox{0.48\textwidth}{!}{\includegraphics{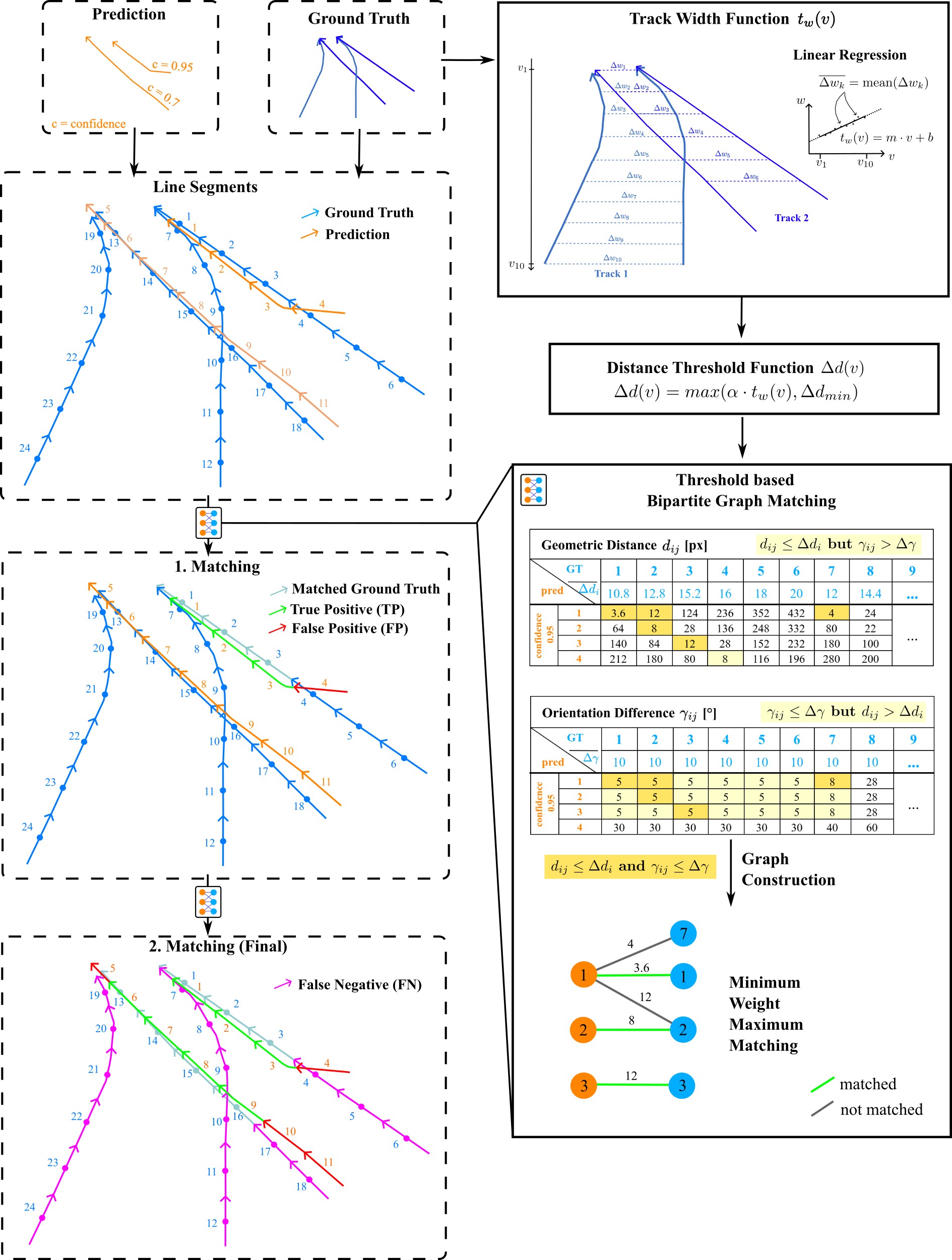}}
    \caption{Illustration of the proposed LineAP matching. Ground-truth (\textcolor[HTML]{00acff}{blue}) and predicted (\textcolor[HTML]{ff8400}{orange}) rails are split into fixed-length segments. Segment pairs are considered matches only if both their geometric distance and their orientation difference fall below specific thresholds. The orientation threshold $\Delta \gamma$ is fixed to $10$° for all ground-truth segments, whereas the distance threshold $\Delta d$ is defined with respect to the track width $t_w(v)$ at the position $v$ of the individual segment center point on the vertical image axis. In the given example, a scaling factor of $\alpha = 0.1$ is applied. 
    The number of matching steps corresponds to the number of unique confidence scores in the predictions.
    }
    \label{fig:lineap-toy-example}
\end{figure}

A further shortcoming shared by all three metrics -- including ChamferAP when applied to images rather than the world coordinates for which it was designed -- is that distance thresholds are defined in pixel space. As a consequence, lines closer to the camera are evaluated against much stricter real-world distances than those in the background, which is particularly problematic for railway applications where safety requirements demand high geometric accuracy even at large distances. Therefore, we propose replacing the fixed pixel threshold with one that varies across the image, using the track width $t_w$ -- which is constant in the real world -- as a reference. The resulting threshold approximates a consistent real-world distance regardless of the position in the image.

\subsubsection{Distance Threshold Function}

We define the track width $t_w$ as the distance between the outside edges of the rail heads, which is the exact position where rails in our dataset are annotated. The track width $t_w$ combines the track gauge, which is a fixed constant, and the width of the rail heads, which can vary in general but can be assumed constant within each image~\cite{holzfeind_handbucheisenbahninfrastruktur_2025}.

We define a track width function $t_w(v)$ for each individual image using the observation that the horizontal distance between two rails corresponds well to the track width. We sample $10$ equidistant points $v_k$ along the vertical image axis within the range of available tracks and, at each position, measure the horizontal distance $\Delta w_k$ between the left and right rail of each visible track. Using linear regression, we obtain the linear function $t_w(v) = m\cdot v + b$ that approximates the track width in pixels as a function of the vertical image position (see Fig.~\ref{fig:lineap-toy-example}). Despite its simplicity, this heuristic yields a robust estimate of $t_w$ across the image.

As we now have an estimate for the track width in image space, we can combine it with a simple scaling factor $\alpha$ to obtain a variable distance threshold $\Delta d(v)$ in pixel units that corresponds approximately to a fixed real-world distance.
\begin{equation}
    \Delta d(v) = \max(\alpha \cdot t_w(v), \Delta d_{\text{min}})
\end{equation}
A scaling factor of $\alpha = 0.1$ corresponds, for instance, to $10\%$ of the real-world track width, which is around $158~\text{mm}$ depending on the rail type. A minimum distance threshold $\Delta d_{\text{min}}$ prevents overly strict thresholds in the far background.

\subsubsection{LineAP Metric}
LineAP adopts the general AP framework but customizes both the matching process and the definition of predicted and ground-truth instances to be matched (see Fig.~\ref{fig:lineap-toy-example}). To this end, both ground-truth and predicted lines are partitioned into short segments of fixed length (the final segment of each line may be shorter). Each predicted segment inherits the confidence score of its parent line. Predicted segments are then matched to ground-truth segments in descending order of confidence score, so that in each matching step, all segments sharing the same confidence score are processed simultaneously. Throughout this process, each segment can be matched at most once.

A predicted segment $j$ and a ground-truth segment $i$ are matched only if two conditions hold: (i) the Euclidean distance $d_{ij}$ between the center point of the ground-truth segment and the closest point on the predicted segment is smaller than the distance threshold $\Delta d(v_i)$, where $v_i$ is the vertical image coordinate of the ground-truth center point, and (ii) the difference in orientation $\gamma_{ij}$ between the two segments is smaller than $\Delta \gamma$.

To obtain the optimal matches, we build a bipartite graph of all valid matches. Predicted and ground-truth segments form the two vertex sets, and edges indicate valid matches, weighted by the geometric distance between the segments. We then use the \textit{minimum-weight maximum matching} algorithm to find the optimal assignment between segments.

LineAP addresses the limitations of the other metrics discussed above in several ways. The segment-based decomposition allows partially detected lines to contribute true positive (TP) matches rather than being penalized as failed instance detections. At the same time, predicted segments without a corresponding ground-truth match are penalized as false positives (FP), ensuring that over-detection is reflected in the evaluation (see Fig.~\ref{fig:metrics-lineap-vs-chamfer}). The track-width-based distance threshold combined with the orientation threshold, enables fine-grained evaluation of geometric accuracy at an approximately consistent real-world scale. Finally, incorporating confidence scores via the AP framework avoids reliance on a single, arbitrary detection threshold.

\subsubsection{Benchmark Metrics}
Since detecting rails as complete instances remains important, e.g., for identifying the ego-track of the train, we incorporate both ChamferAP and LineAP in our rail track detection challenge. We integrate the track-width-based distance threshold $\Delta d(v)$ into the ChamferAP metric, using the center point of each ground-truth rail along the vertical image axis as the reference. For each metric, we pick five different scaling factors $\alpha$ for the distance threshold and report the individual results as well as the mAP across all thresholds. We additionally report the average of both metrics' mAP scores as $\text{mAP}_{\text{avg}}$. We further report the scores for the TuSimple and CULane metrics in the supplements for broader context.

\begin{table}[!tb]
  \centering
  \caption{Object detection results on the RAIL-BENCH Object test set. The first three rows show mAP scores averaged across all classes and IoU thresholds for the difficulty levels e (easy), m (moderate), and h (hard). AP@50 corresponds to an IoU of $0.5$. \textbf{Best results} are shown in bold, \underline{second-best results} are underlined.}
  \label{tab:od_results}
  {\small
  \setlength{\tabcolsep}{4pt}
  \renewcommand{\arraystretch}{1.05}
  \resizebox{\linewidth}{!}{%
  \begin{tabular}{clccccc}
    \toprule
     & & \multirow{2}{*}{\shortstack{YOLOv8L\\World}} & \multirow{2}{*}{\shortstack{{\setbox0\hbox{YOLOv11L}\hbox to\wd0{\hss YOLO11L\hss}}\\COCO}} & \multirow{2}{*}{\shortstack{YOLO11L\\RAIL-BENCH}} & \multirow{2}{*}{\shortstack{DINO-DETR\\COCO}} & \multirow{2}{*}{\shortstack{DINO-DETR\\RAIL-BENCH}} \\
     \\
    \midrule
    \multicolumn{2}{l}{mAP@[.50:.95]\textsubscript{e}} & 18.7 & 20.4 & \underline{35.1} & 21.6 & \textbf{39.6} \\
    \multicolumn{2}{l}{mAP@[.50:.95]\textsubscript{m}} & 12.5 & 13.5 & \underline{25.5} & 14.9 & \textbf{30.4} \\
    \multicolumn{2}{l}{mAP@[.50:.95]\textsubscript{h}} & 9.2 & 9.8 & \underline{16.7} & 10.7 & \textbf{20.8} \\
    \midrule
    \multirow{7}{*}{\rotatebox[origin=c]{90}{AP@50 per class}} & Train & 34.9 & 35.8 & \underline{43.9} & 34.7 & \textbf{48.4} \\
     & Person & 43.5 & 45.4 & \underline{58.7} & 54.5 & \textbf{71.2} \\
     & Road Vehicle & 21.2 & 26.9 & \underline{32.0} & 31.5 & \textbf{45.5} \\
     & Bicycle & 8.4 & 6.9 & 4.4 & \underline{11.2} & \textbf{15.3} \\
     & Signal & 7.0 & 9.1 & \underline{26.7} & 14.9 & \textbf{35.1} \\
     & Signal Pole & 0.2 & 0.0 & \underline{19.7} & 0.0 & \textbf{21.2} \\
     & Catenary Pole & 3.5 & 0.0 & \underline{39.9} & 0.0 & \textbf{51.7} \\
    \bottomrule
  \end{tabular}%
  }%
  }
\end{table}

\begin{table*}[ht]
\centering
\caption{LineAP and ChamferAP results on the RAIL-BENCH test set. Each model is evaluated in three training settings: trained only on RailSem19 (pretrained), trained only on RAIL-BENCH Rail (RB only) and finetuned on RAIL-BENCH Rail after pretraining on RailSem19 (finetuned). Distance thresholds @$\alpha$ are given in percentage. mAP$_{\text{avg}}$ is the average between mAP of LineAP and ChamferAP. \textbf{Best results} are shown in bold, \underline{second-best results} are underlined.}
\label{tab:rail_baselines}
{\scriptsize
\begin{tabular*}{0.8\textwidth}{@{\extracolsep{\fill}} l c c c c c c c c c c c c c}
\toprule
 & \multicolumn{6}{c}{LineAP} & \multicolumn{6}{c}{ChamferAP} & \multirow{2}{*}{$\text{mAP}_{\text{avg}}$} \\
\cmidrule(lr){2-7}\cmidrule(lr){8-13}
Model & @3 & @5 & @7 & @10 & @20 & mAP & @5 & @10 & @20 & @30 & @50 & mAP &  \\
\midrule
PINet (pretrained) & 28.3 & 41.5 & 47.5 & 52.9 & 60.2 & 46.1 & 9.5 & 26.4 & 35.2 & 38.9 & 42.4 & 30.5 & 38.3 \\
PINet (RB only) & 30.8 & 40.1 & 45.5 & 50.7 & 58.6 & 45.1 & 10.7 & 25.4 & 33.9 & 38.1 & 42.2 & 30.1 & 37.6 \\
PINet (finetuned) & 35.5 & 45.3 & 51.0 & 55.9 & 63.3 & 50.2 & 14.3 & 27.4 & 36.8 & \underline{40.3} & \underline{43.6} & 32.5 & 41.3 \\
\midrule
YOLinO (pretrained) & 44.4 & 52.9 & \underline{56.5} & \underline{60.1} & 64.6 & 55.7 & 6.4 & 15.1 & 20.8 & 24.4 & 30.3 & 19.4 & 37.5 \\
YOLinO (RB only) & 46.0 & 51.9 & 55.6 & 59.3 & 63.7 & 55.3 & 6.7 & 14.4 & 19.8 & 23.2 & 29.2 & 18.7 & 37.0 \\
YOLinO (finetuned) & \textbf{51.9} & \textbf{56.1} & \textbf{59.2} & \textbf{62.5} & \textbf{66.5} & \textbf{59.2} & 8.8 & 16.6 & 22.6 & 26.7 & 32.8 & 21.5 & 40.4 \\
\midrule
CLRNet (pretrained) & 42.7 & 50.0 & 53.2 & 56.5 & 61.0 & 52.7 & \underline{20.1} & 31.1 & 36.8 & 39.2 & 41.4 & 33.7 & 43.2 \\
CLRNet (RB only) & 46.1 & 51.7 & 55.7 & 60.0 & \underline{65.6} & 55.8 & 19.7 & \underline{31.4} & \textbf{38.9} & \textbf{42.0} & \textbf{45.1} & \textbf{35.4} & \textbf{45.6} \\
CLRNet (finetuned) & \underline{49.4} & \underline{53.1} & 55.7 & 58.5 & 62.5 & \underline{55.9} & \textbf{22.9} & \textbf{32.5} & \underline{37.7} & 39.7 & 41.6 & \underline{34.9} & \underline{45.4} \\
\bottomrule
\end{tabular*}%
}
\end{table*}

\section{Experiments} 
\label{sec:exp}

\subsection{Object Detection}

For the object detection challenge, we evaluate two YOLO architectures, YOLO11L~\cite{yolo11_ultralytics} and YOLOv8L-World~\cite{yolo_world}, as well as DINO-DETR~\cite{dino_detr}. Both YOLO11L and DINO-DETR are pretrained on the COCO dataset~\cite{coco_detection_challenge_codalab} and finetuned on the RAIL-BENCH Object dataset. We evaluate both models before and after finetuning. YOLOv8L-World is a foundation model that is evaluated in a zero-shot manner using text prompts such as \textit{tall pole with overhead wires} for the class \textit{catenary poles}. Tab.~\ref{tab:od_results} displays the results.

For all classes except for bicycle, the finetuned models -- first DINO-DETR, secondly YOLO11L -- achieve the best results. Note that this applies not only for railway specific categories but also for persons and road vehicles. Hence, training on railway-specific datasets is beneficial even for general object categories, likely due to the domain shift between standard benchmark datasets and the visual conditions encountered in railway environments.

\begin{figure*}[!tb]
  \centering
  \setlength{\tabcolsep}{2pt}
  \renewcommand{\arraystretch}{0}
  \begin{tabular}{ccccc}
    & \multicolumn{1}{c}{\small\textbf{PINet (finetuned)}} &
    \multicolumn{1}{c}{\small\textbf{YOLinO (finetuned)}} &
    \multicolumn{1}{c}{\small\textbf{CLRNet (finetuned)}} \\[4pt]
    \rotatebox{90}{\parbox{2.7cm}{\centering\small\textbf{Ground Truth}\\\footnotesize(random colors)}} &
    \includegraphics[width=0.27\textwidth]{images/Metrics/gt/img_2013_test.jpg} &
    \includegraphics[width=0.27\textwidth]{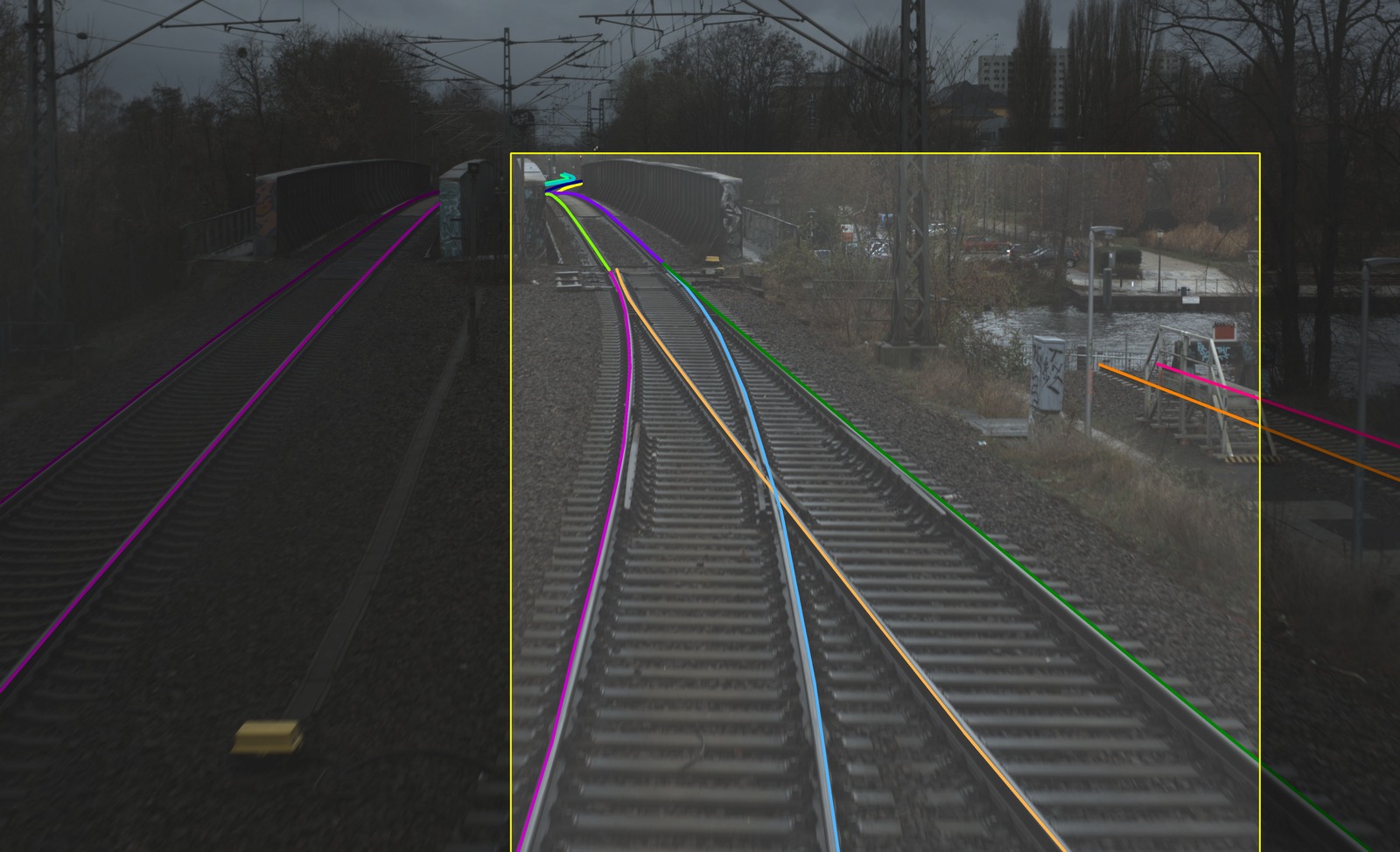} &
    \includegraphics[width=0.27\textwidth]{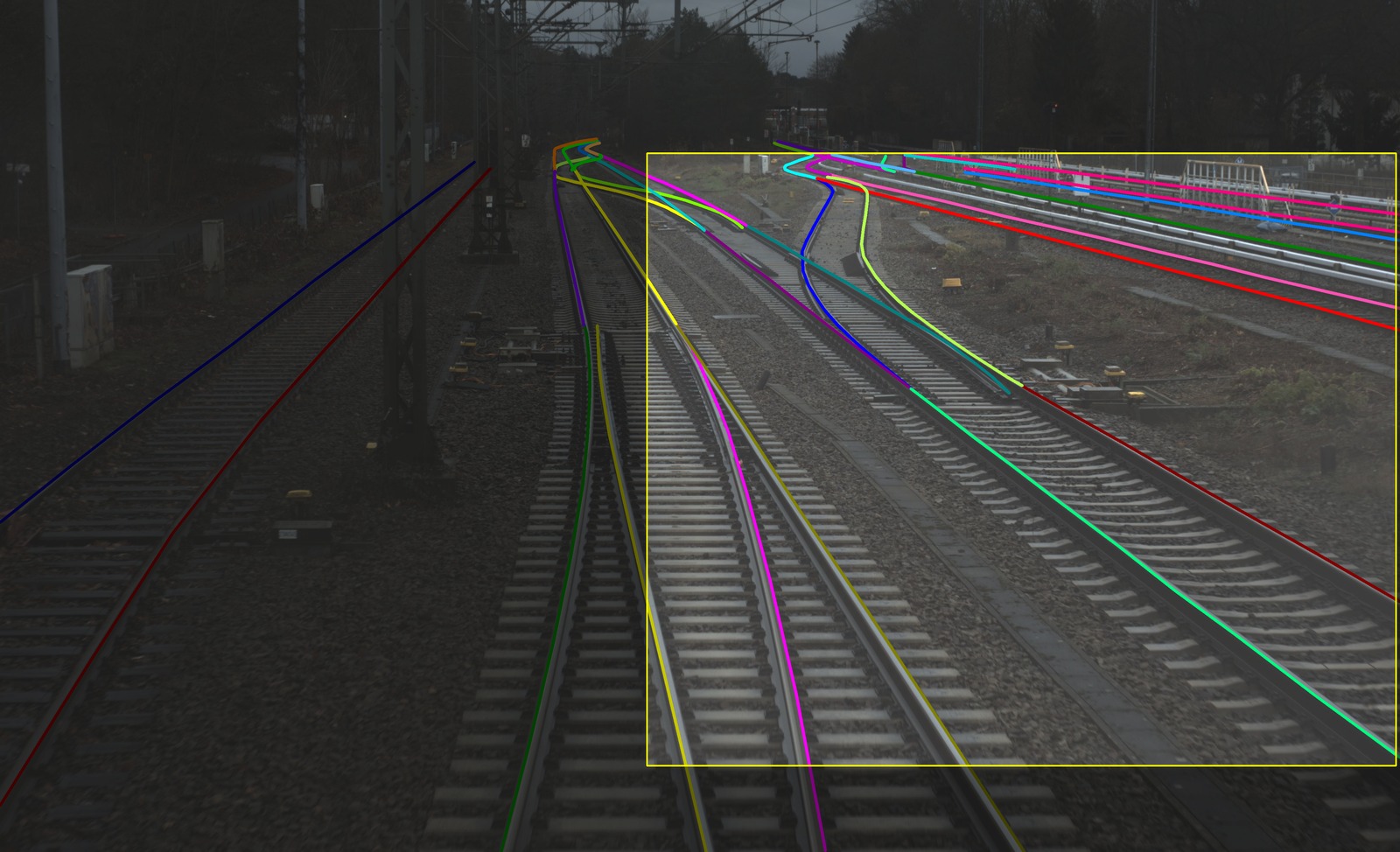} \\
    \rotatebox{90}{\parbox{2.7cm}{\centering\small\textbf{Predictions}\\\footnotesize(random colors)}} &
    \includegraphics[width=0.27\textwidth]{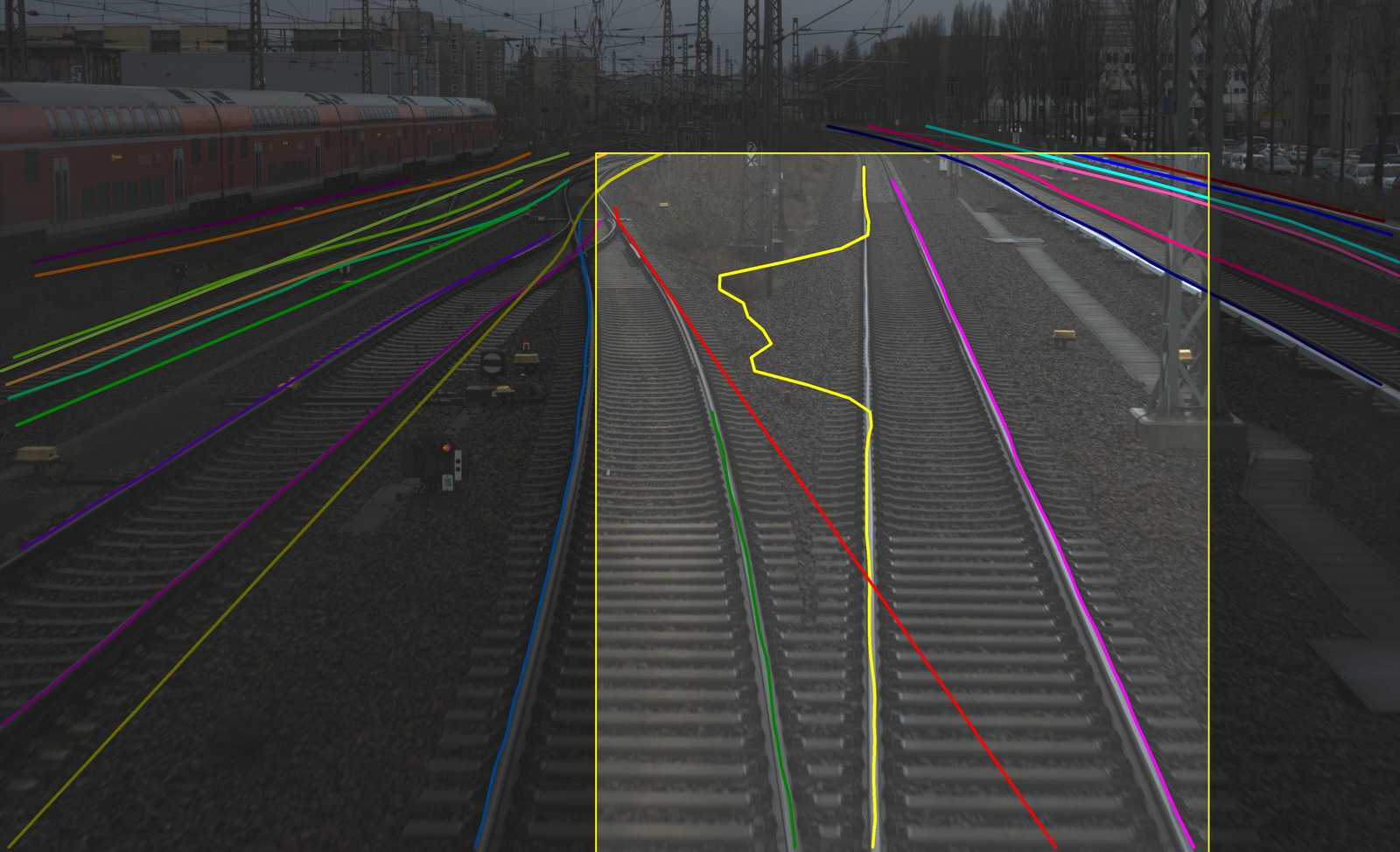} &
    \includegraphics[width=0.27\textwidth]{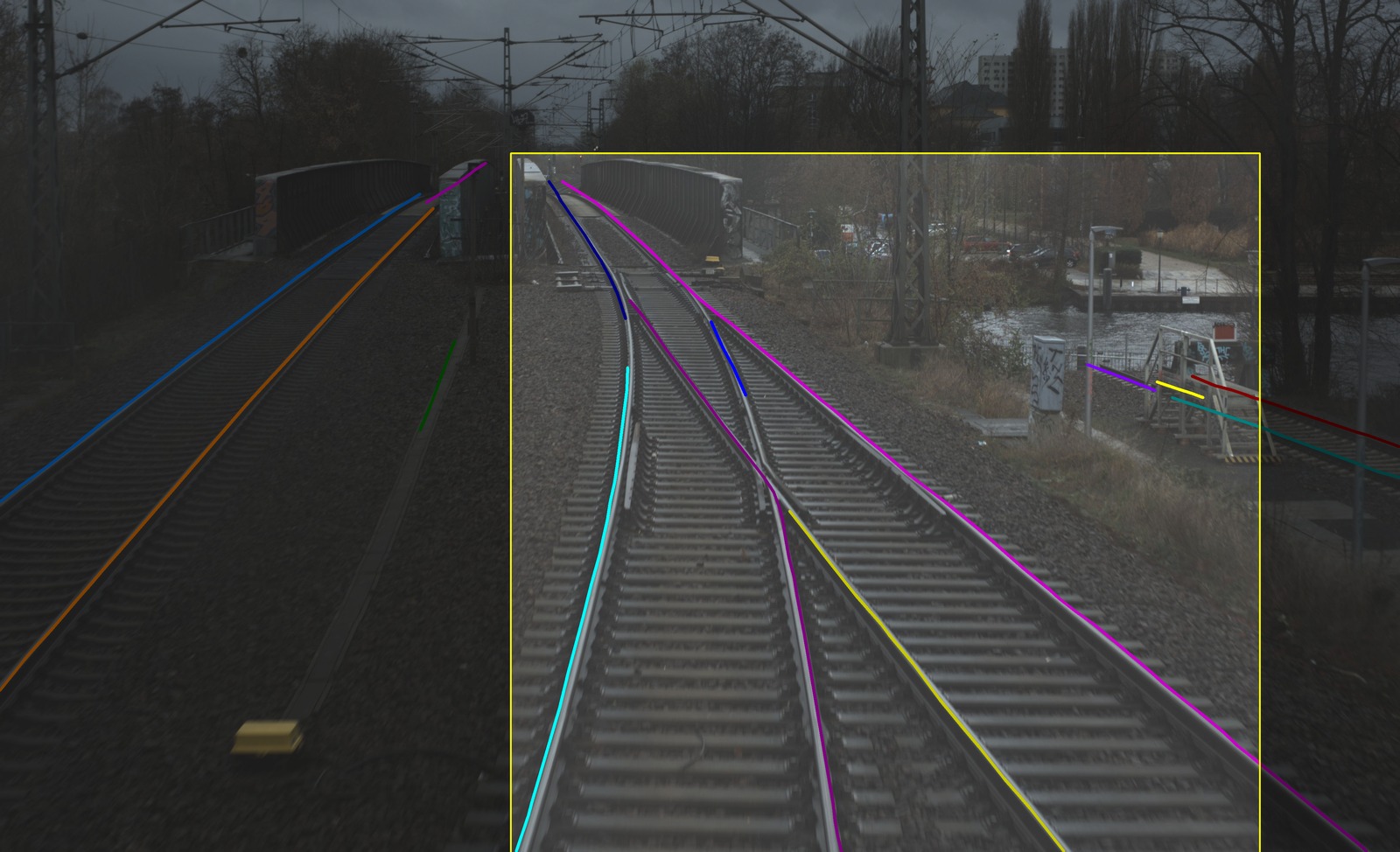} &
    \includegraphics[width=0.27\textwidth]{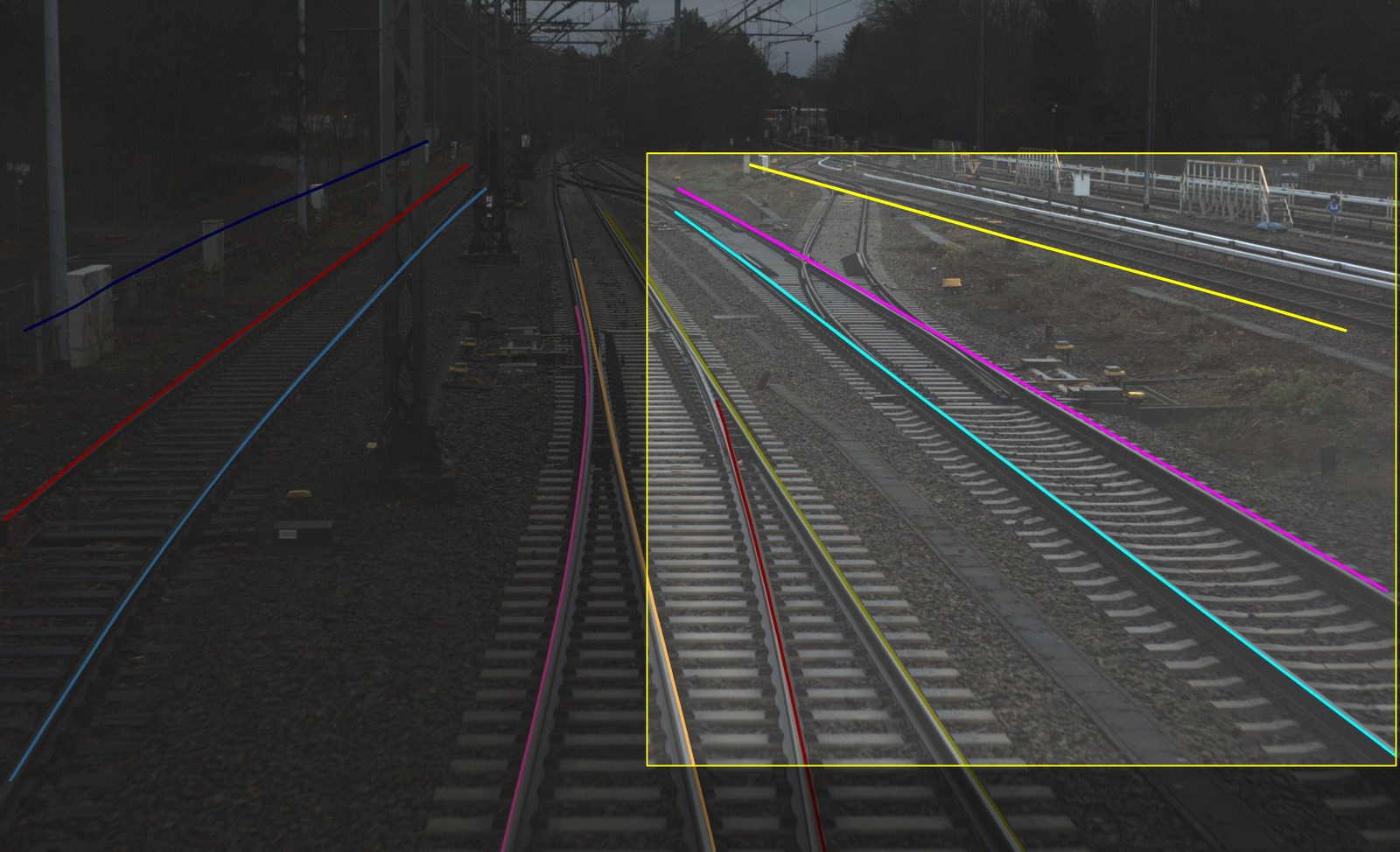} \\[4pt]
    \rotatebox{90}{\parbox{2.7cm}{\centering\small\textbf{easy}}} &
    \includegraphics[width=0.27\textwidth]{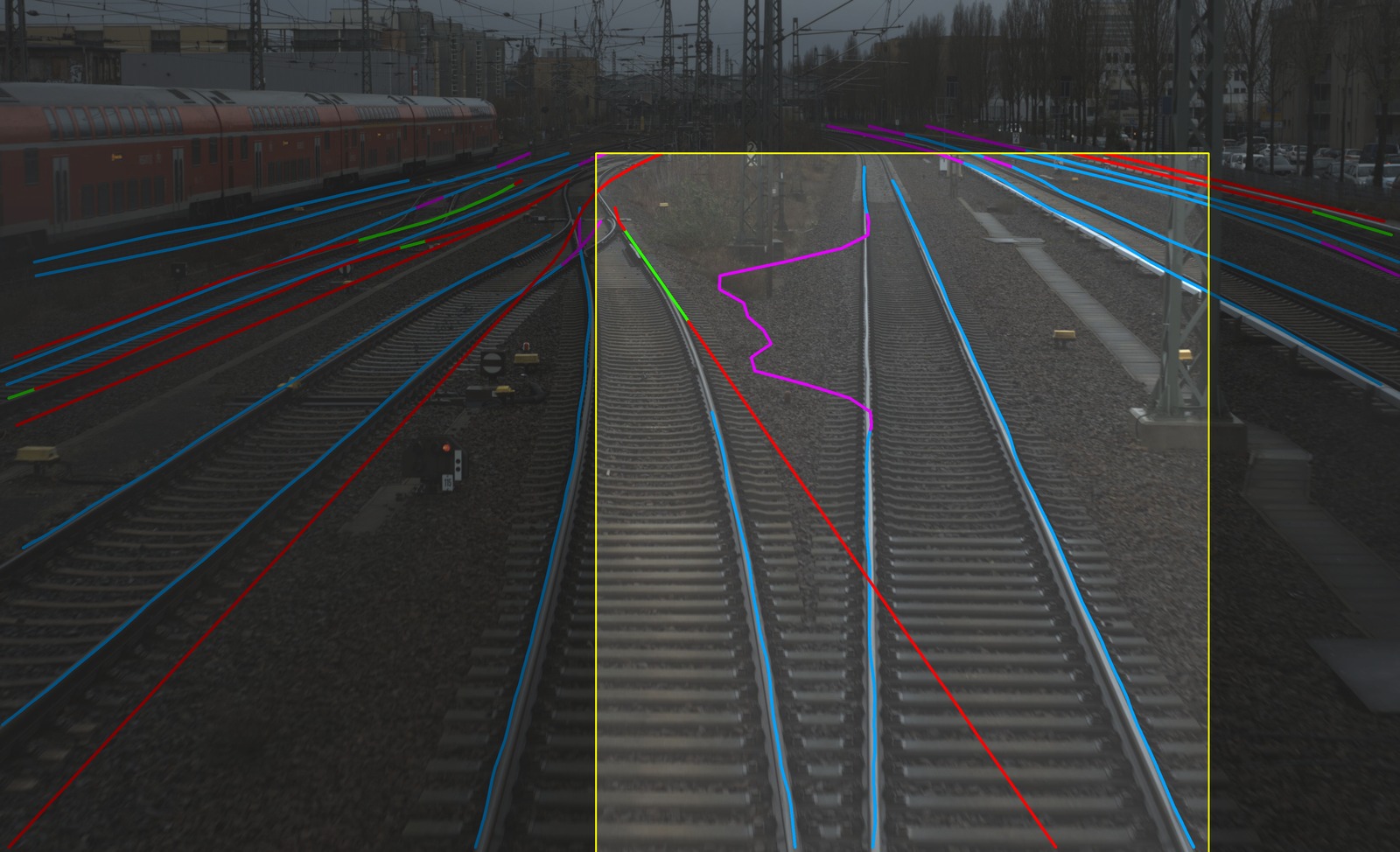} &
    \includegraphics[width=0.27\textwidth]{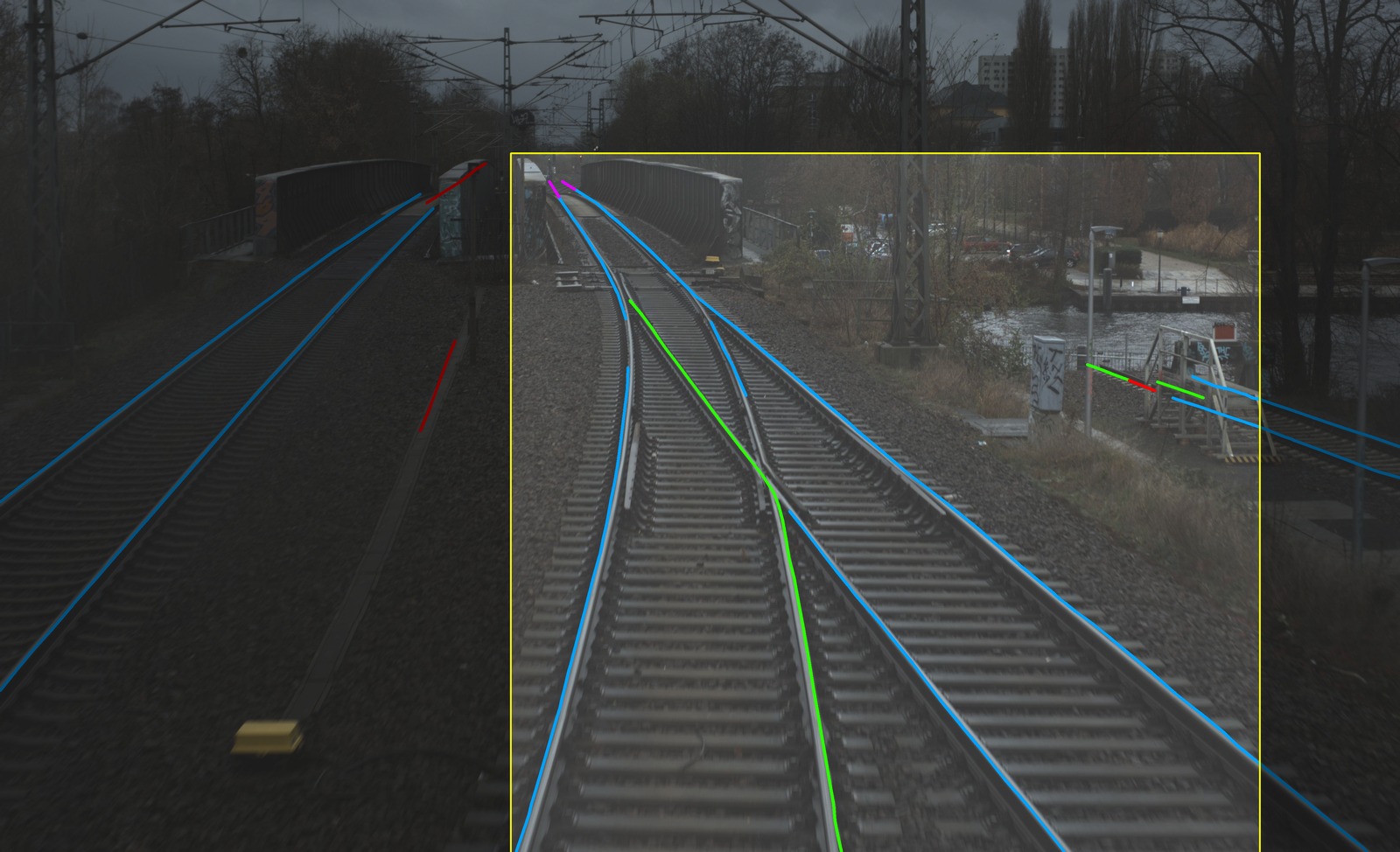} &
    \includegraphics[width=0.27\textwidth]{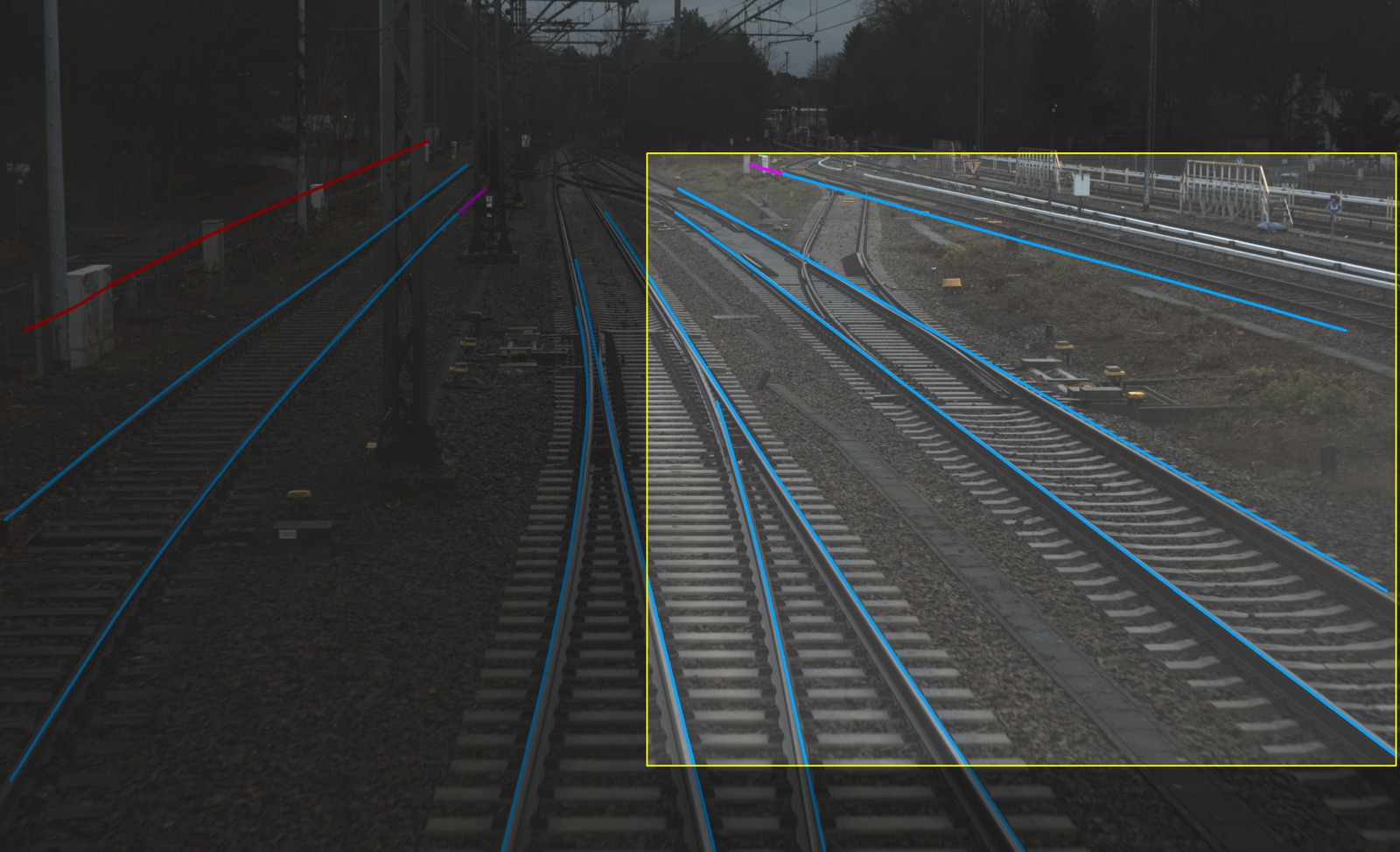} \\
    \rotatebox{90}{\parbox{2.7cm}{\centering\small\textbf{moderate}}} &
    \includegraphics[width=0.27\textwidth]{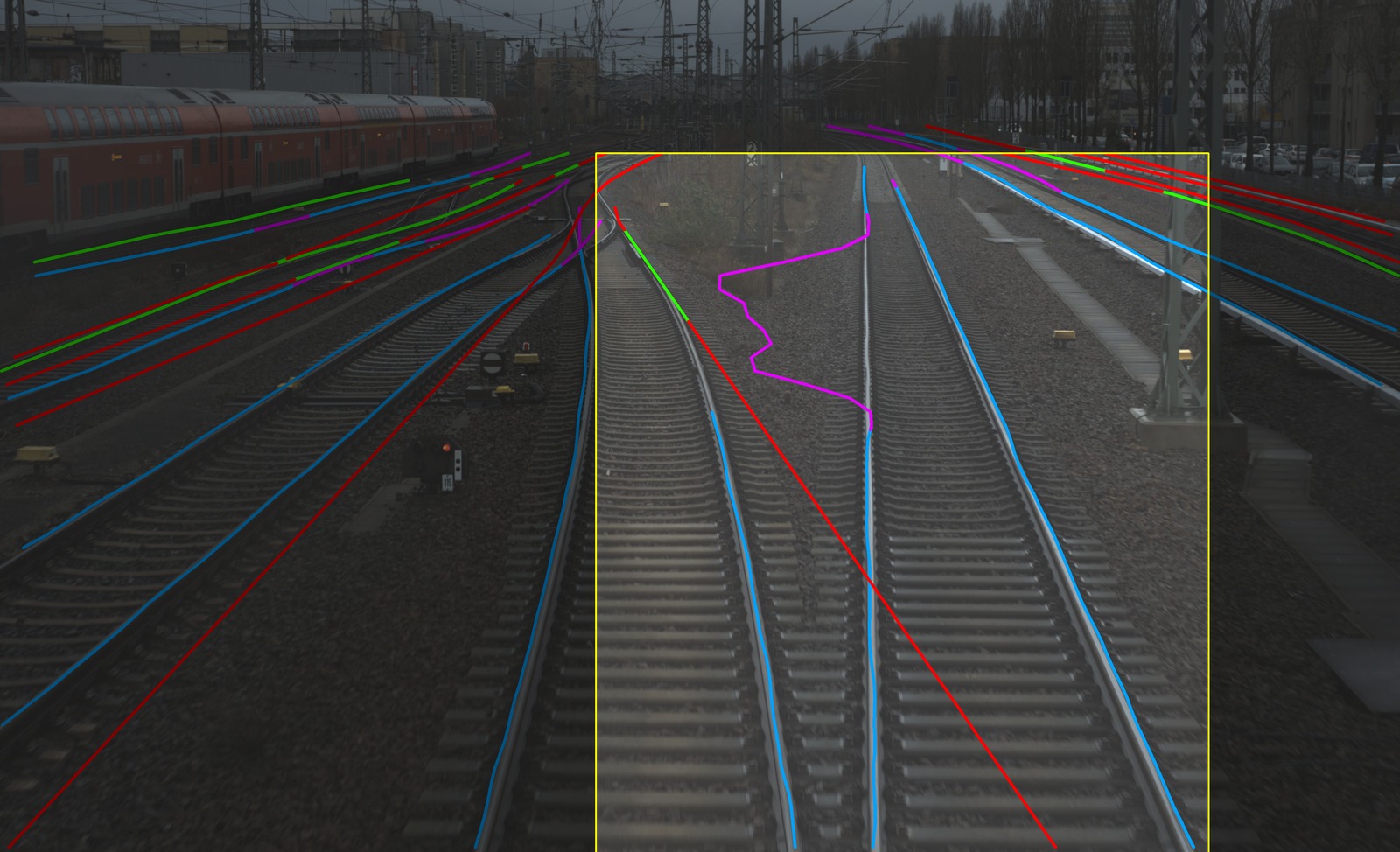} &
    \includegraphics[width=0.27\textwidth]{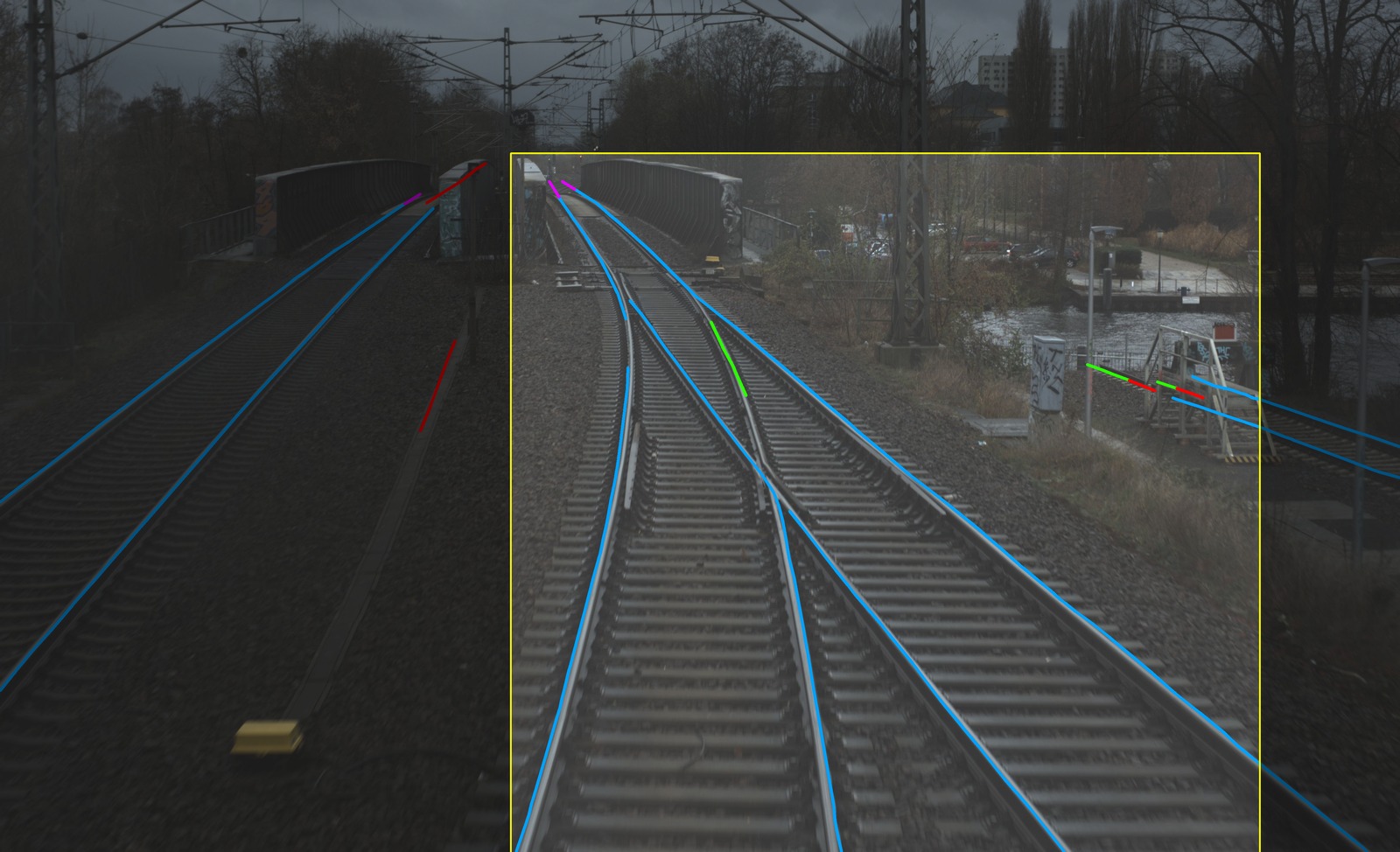} &
    \includegraphics[width=0.27\textwidth]{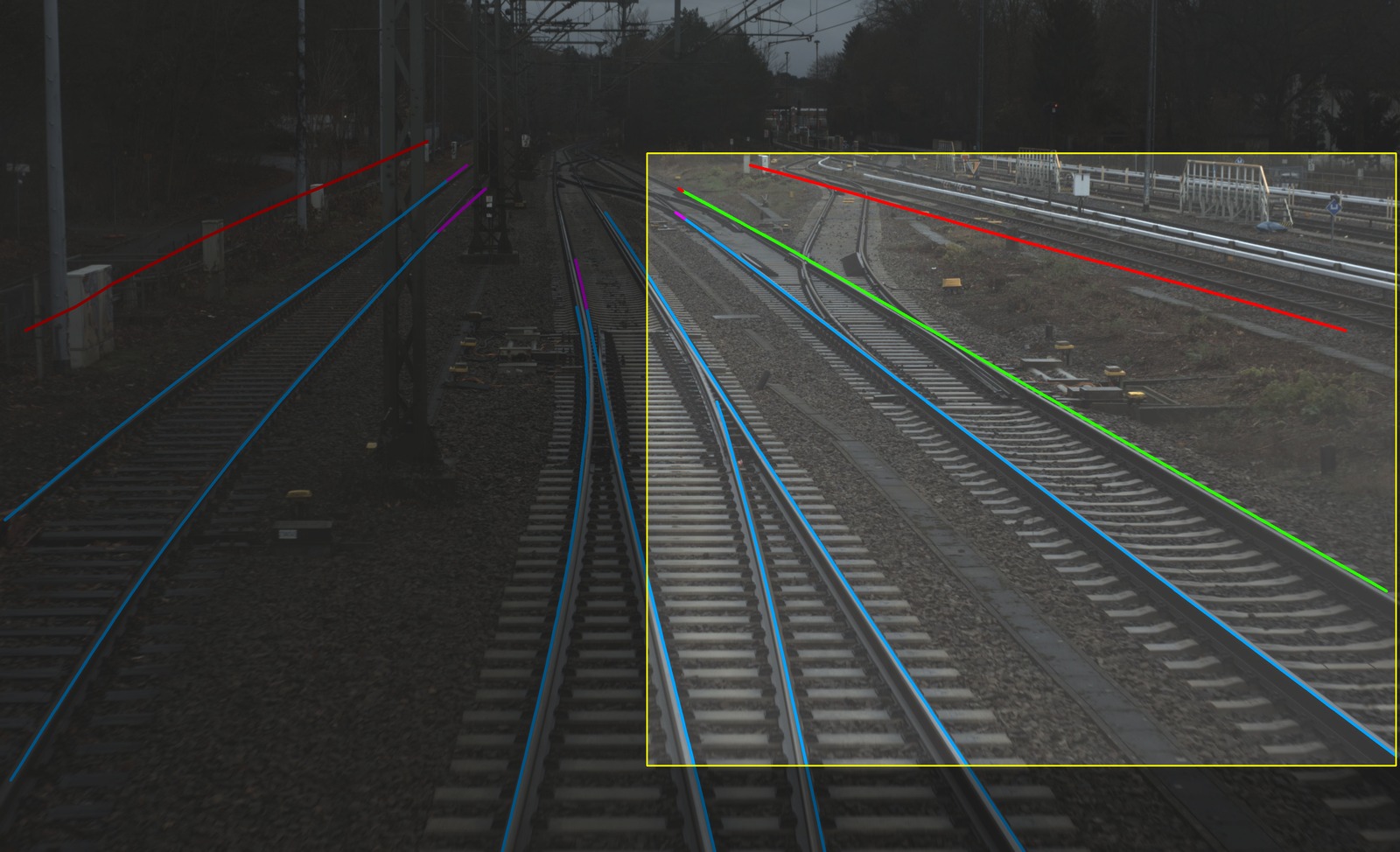} \\
    \rotatebox{90}{\parbox{2.7cm}{\centering\small\textbf{hard}}} &
    \includegraphics[width=0.27\textwidth]{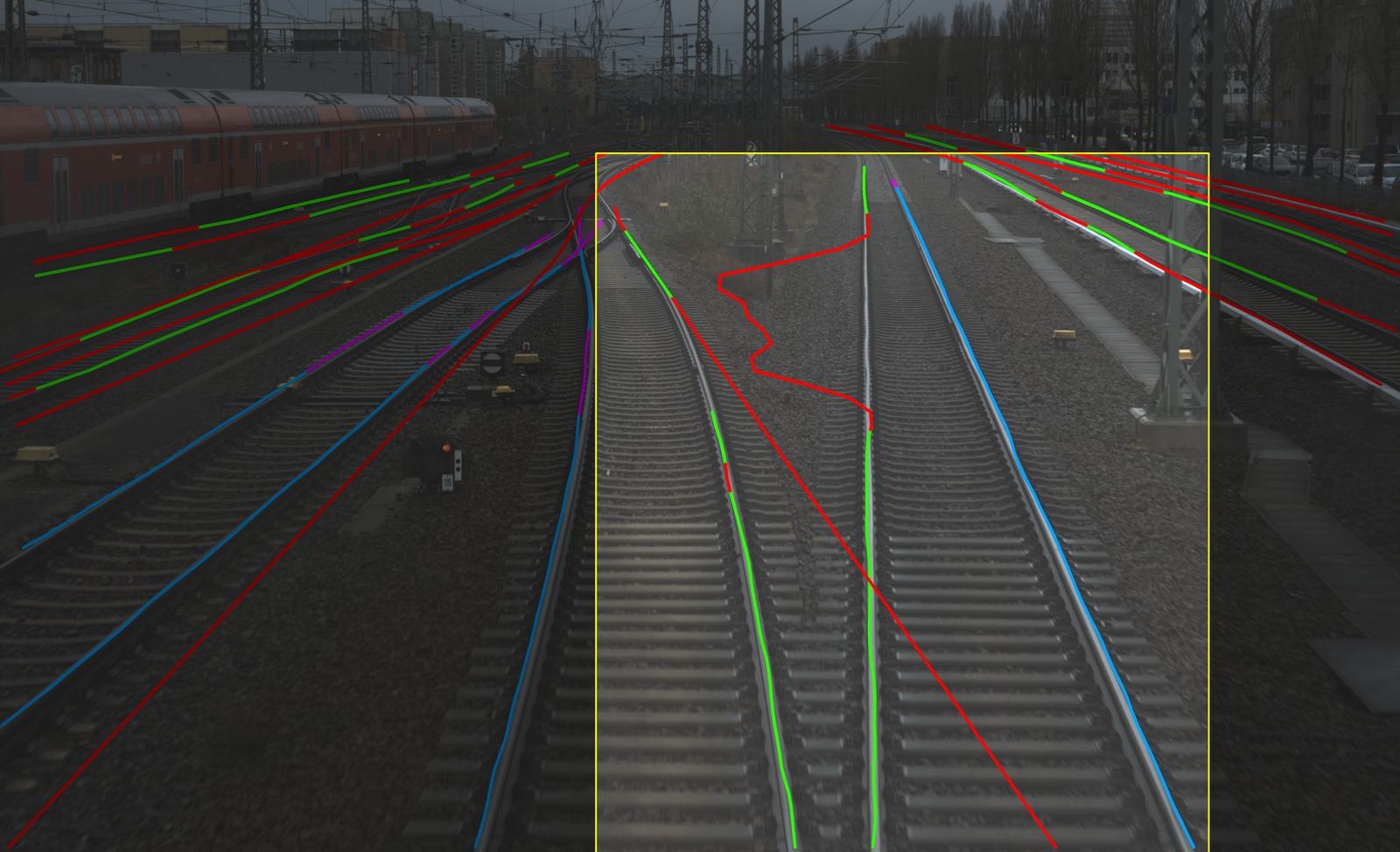} &
    \includegraphics[width=0.27\textwidth]{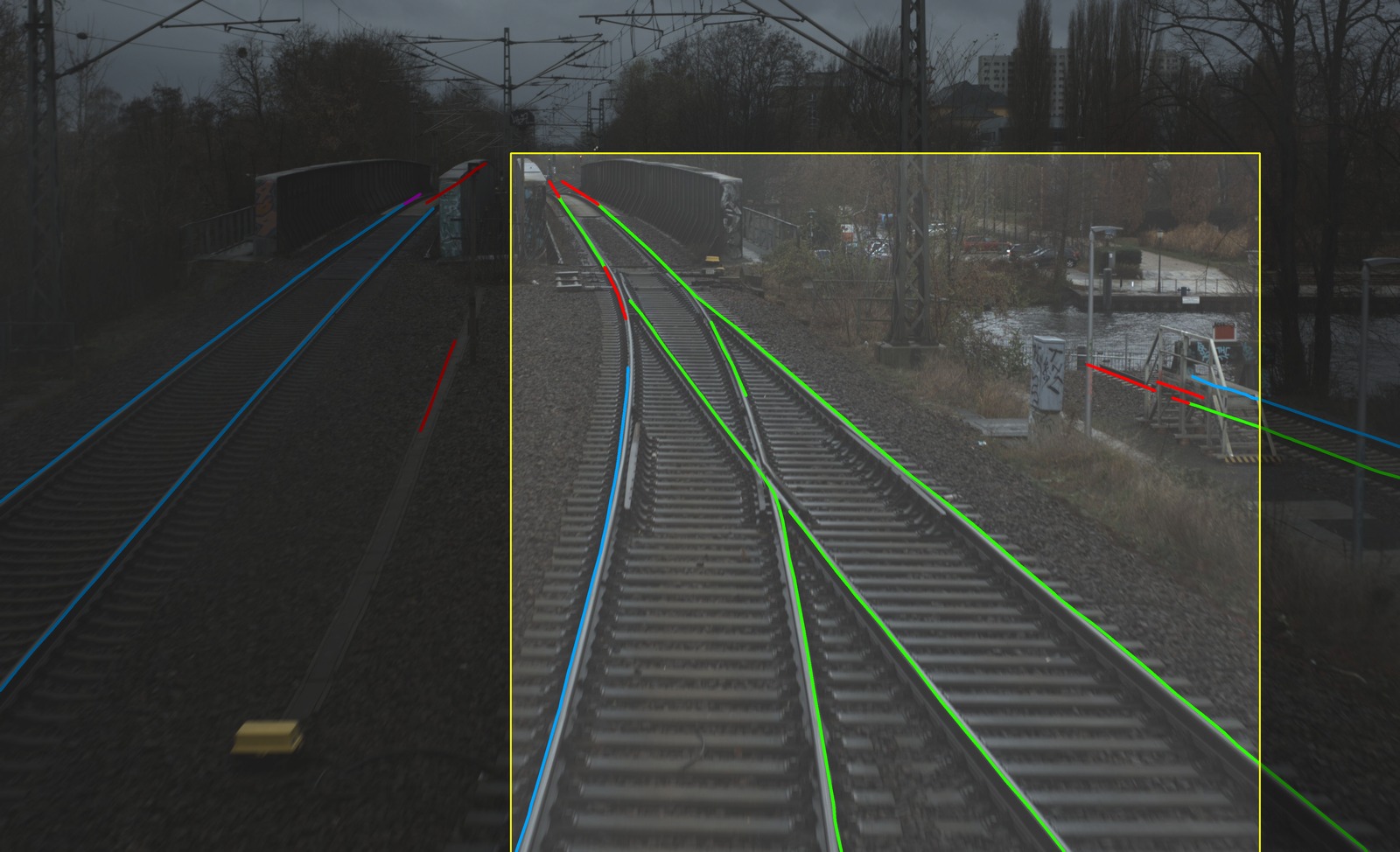} &
    \includegraphics[width=0.27\textwidth]{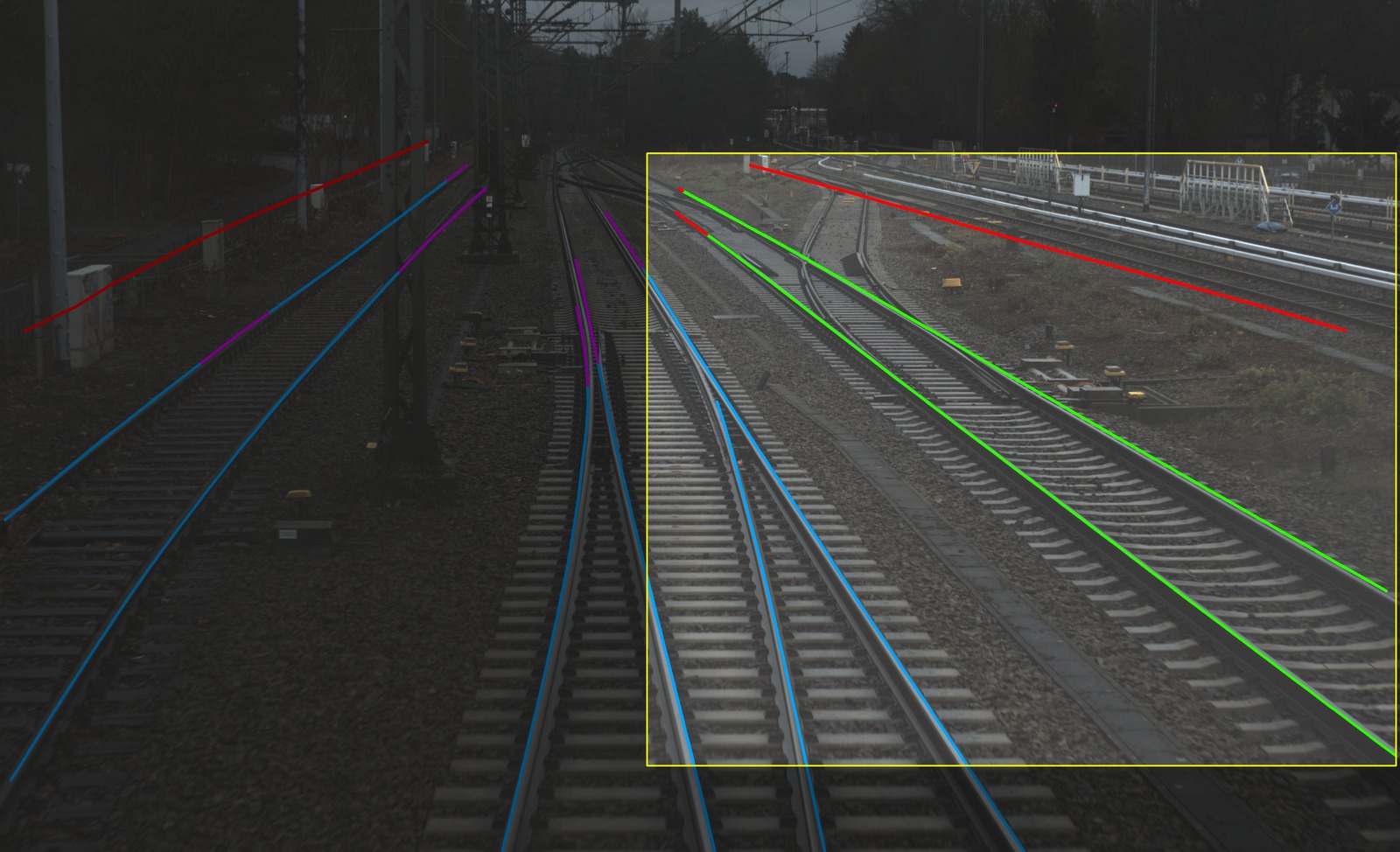} &
    \metriclegend{ChamferAP}
  \end{tabular}
  \caption{Ground truth (first row) and predicted rails (rows 2--5) from finetuned PINet, YOLinO, and CLRNet, evaluated with LineAP and ChamferAP in three difficulty levels: easy (LineAP@20 vs. ChamferAP@50), moderate (LineAP@7 vs. ChamferAP@20), hard (LineAP@3 vs. ChamferAP@5). 
  \textcolor[HTML]{00a2ff}{Blue}: both metrics agree that the prediction is correct.
  \textcolor[HTML]{ff0000}{Red}: both metrics agree that the prediction is incorrect.
  \textcolor[HTML]{2bff00}{Green}: identified as correct by LineAP only.
  \textcolor[HTML]{ff04ea}{Pink}: identified as correct by ChamferAP only.
  Best viewed in color.}
  \label{fig:metrics-lineap-vs-chamfer}
\end{figure*}

\subsection{Rail Track Detection}\label{sec:exp_rails}

We employ three network architectures for rail track detection: PINet~\cite{ko_keypoints_2022}, originally used for lane and railway track detection; YOLinO~\cite{Meyer_2021_ICCV}, a polyline detection framework from autonomous driving; and CLRNet~\cite{zhengCLRNetCrossLayer2022a}, a state-of-the-art lane detection model.
We train each model on the RailSem19 dataset and finetune it thereafter on our RAIL-BENCH Rail dataset. In an additional setting, we train exclusively on our dataset, i.e., without RailSem19 pretraining. Tab.~\ref{tab:rail_baselines} shows the results of these experiments.

Throughout both metrics and all architectures, finetuning on our dataset improves the performance of the pretrained models. 
CLRNet is the overall best architecture, with all three variants outperforming every other model in $\text{mAP}_{\text{avg}}$; however, for CLRNet, pretraining on RailSem19 is not beneficial throughout all evaluation settings. For more lenient thresholds, \mbox{CLRNet} trained exclusively on our RAIL-BENCH dataset is superior in both LineAP and ChamferAP. This may stem from differences in annotation policy between RailSem19 and RAIL-BENCH, such as the exact polyline positioning on the rail head -- a bias that finetuning may not fully overcome.

Beyond CLRNet's overall best scores, the finetuned YOLinO model achieves the best LineAP scores across all distance thresholds. We attribute this to the design of YOLinO, which is optimized to detect line segments rather than complete lines; the reconstruction of final lines is deferred to a subsequent postprocessing step that assembles individual segments. Hence, the model does not produce long, continuous line predictions, but instead yields a larger number of shorter line hypotheses for which local correctness is prioritized over global continuity.

Fig.~\ref{fig:metrics-lineap-vs-chamfer} displays example predictions evaluated with both LineAP and ChamferAP, with a gradually harder distance threshold from top (easy) to bottom (hard). In the left example, only LineAP rewards the partially correct prediction as a TP while disregarding the incorrect part as a FP. As ChamferAP evaluates the predicted line as a single unit, it considers the entire line as either TP or FP depending on the strictness of the evaluation. In the center example, YOLinO identifies almost all relevant switch rails, but as these do not correspond well to the ground-truth line instances, ChamferAP considers them incorrect. In the right example, the predicted rails correspond to several ground-truth lines; again, only LineAP rewards the geometrically correct predictions. 

These examples showcase that evaluating with both metrics provides more insight than each metric alone. LineAP precisely evaluates the geometric accuracy of predictions, while ChamferAP also considers the completeness of predicted line instances.

\subsection{Vegetation Segmentation}\label{sec:exp_vegetation}

\begin{table}[!tb]
\centering
\caption{Vegetation segmentation IoU scores on the RAIL-BENCH test set. mIoU averages the IoU of the Veg low, Veg high and background class. \textbf{Best results} are shown in bold, \underline{second-best results} are underlined.}
\label{tab:railbench_vegetation}
{\small
\setlength{\tabcolsep}{4pt}
\renewcommand{\arraystretch}{1.05}
\resizebox{\linewidth}{!}{%
\begin{tabular}{l c c c c}
\toprule
Model & Veg (low) & Veg (high) & Veg (single-cls) & mIoU \\
\midrule
SegFormer-B2 (Cityscapes zero-shot) & 27.9 & 72.8 & 79.5 & 63.4 \\
SegFormer-B2 (RailSem19) & 39.9 & 79.4 & 84.2 & 70.5 \\
SegFormer-B2 (RAIL-BENCH) & \underline{49.1} & 80.3 & 86.0 & 74.1 \\
\midrule
Mask2Former (Cityscapes zero-shot) & 37.4 & 72.6 & 81.6 & 66.9 \\
Mask2Former (RailSem19) & 42.6 & 79.8 & 85.2 & 71.6 \\
Mask2Former (RAIL-BENCH) & \textbf{52.5} & \textbf{84.4} & \underline{89.2} & \underline{77.1} \\
\midrule
DeepLabV3+ (Cityscapes zero-shot) & 24.4 & 72.2 & 81.3 & 62.2 \\
DeepLabV3+ (RailSem19) & 41.4 & 80.8 & 84.7 & 71.5 \\
DeepLabV3+ (RAIL-BENCH) & \textbf{52.5} & \underline{84.3} & \textbf{90.0} & \textbf{77.3} \\
\bottomrule
\end{tabular}%
}%
}
\end{table}

  \begin{table}[!tb]
  \centering
  \caption{Vegetation segmentation IoU scores with restriction to the track-bed area (the area between a track's two rails) on the 65 test images with a matching rail-track annotation. \textbf{Best results} are shown in bold, \underline{second-best results} are underlined.}
  \label{tab:railbench_vegetation_track_bed}
  {\small
  \setlength{\tabcolsep}{4pt}
  \renewcommand{\arraystretch}{1.05}
  \resizebox{\linewidth}{!}{%
  \begin{tabular}{l c c c c}
  \toprule
  Model & Veg (low) & Veg (high) & Veg (single-cls) & mIoU \\
  \midrule
  SegFormer-B2 (Cityscapes zero-shot) & 4.4 & 28.0 & 12.6 & 43.2 \\
  SegFormer-B2 (RailSem19) & 8.2 & 36.0 & 23.7 & 47.9 \\
  SegFormer-B2 (RAIL-BENCH) & \underline{38.5} & 37.2 & 41.8 & 58.4 \\
  \midrule
  Mask2Former (Cityscapes zero-shot) & 6.1 & 16.0 & 17.5 & 40.2 \\
  Mask2Former (RailSem19) & 3.6 & 35.1 & 20.6 & 46.0 \\
  Mask2Former (RAIL-BENCH) & 33.4 & \underline{47.3} & \underline{49.9} & \underline{60.1} \\  
  \midrule
  DeepLabV3+ (Cityscapes zero-shot) & 5.9 & 25.2 & 20.0 & 43.0 \\
  DeepLabV3+ (RailSem19) & 0.1 & 32.9 & 16.7 & 44.1 \\
  DeepLabV3+ (RAIL-BENCH) & \textbf{39.9} & \textbf{63.1} & \textbf{60.3} & \textbf{67.6} \\
  \bottomrule
  \end{tabular}%
  }%
  }
  \end{table}

We use three networks as baselines for the Vegetation Segmentation challenge: SegFormer-B2~\cite{xie_segformersimple_2021}, Mask2Former~\cite{cheng2021mask2former}, and DeepLabV3+~\cite{deeplabv3plus2018}, all pretrained on Cityscapes~\cite{cordts_cityscapesdataset_2016}. Each model is then finetuned either on RailSem19 or on our RAIL-BENCH Vegetation dataset.

Tab.~\ref{tab:railbench_vegetation} shows that DeepLabV3+ and Mask2Former perform similarly well and better than SegFormer. Finetuning on RAIL-BENCH gives the best mIoU across all three architectures.

As described in Tab.~\ref{tab:vegetation}, the RAIL-BENCH dataset differs from RailSem19 in that vegetation is also annotated when located on tracks. Using the available rail annotations from the RAIL-BENCH Rail dataset, we can analyze the performance of the segmentation models in this particularly safety-critical area. As displayed in Tab.~\ref{tab:railbench_vegetation_track_bed}, training on the RAIL-BENCH Vegetation dataset significantly improves the models' ability to detect vegetation in the track bed and to distinguish low- from high-growing vegetation. This demonstrates the practical value of RAIL-BENCH's dedicated track-bed vegetation annotations for detecting safety-critical clearance-gauge violations.

Training details such as learning rates and number of epochs for all experiments are provided in the supplements.

\section{Conclusions}

We introduced RAIL-BENCH, the first perception benchmark for automated railway operation. RAIL-BENCH not only offers novel, diverse datasets tailored to railway-specific characteristics but also standardized evaluation protocols, including evaluation metrics, automated evaluation routines, and scoreboards for comparing approaches. This includes our newly proposed \mbox{LineAP} metric for polyline-based predictions, which retains key advantages beyond its track-width-based threshold, making it applicable to other polyline detection tasks.

\section*{Acknowledgments}

This work was supported by the Federal Ministry of Transport of Germany within the mFUND program under grant 01F1178A. The authors thank DB InfraGO, in particular Martin K\"{o}ppel and Philipp Neumaier, for providing data for the project and for their support in the development of the benchmark suite. 

The authors gratefully acknowledge the computing time provided on the high-performance computer HoreKa by the National High-Performance Computing Center at KIT (NHR@KIT). This center is jointly supported by the Federal Ministry of Research, Technology and Space and the Ministry of Science, Research and the Arts of Baden-Württemberg, as part of the National High-Performance Computing (NHR) joint funding program. HoreKa is partly funded by the German Research Foundation (DFG).

{
    \small
    \bibliographystyle{ieeenat_fullname}
    \bibliography{main}
}

\clearpage
\twocolumn[
  \centering
  \Large
  \textbf{\thetitle}\\
  \vspace{0.5em}Supplementary Material \\
  \vspace{1.0em}
]
\appendix
\counterwithin{figure}{section}
\counterwithin{table}{section}
\renewcommand{\thefigure}{\thesection\arabic{figure}}
\renewcommand{\thetable}{\thesection\arabic{table}}
\section{Additional RAIL-BENCH Challenges}
\label{sec:challenges}

\begin{figure*}[t]
  \centering
  \setlength{\tabcolsep}{2pt}
  \renewcommand{\arraystretch}{0}
  \begin{tabular}{ccc}
    \includegraphics[width=0.32\textwidth,keepaspectratio]{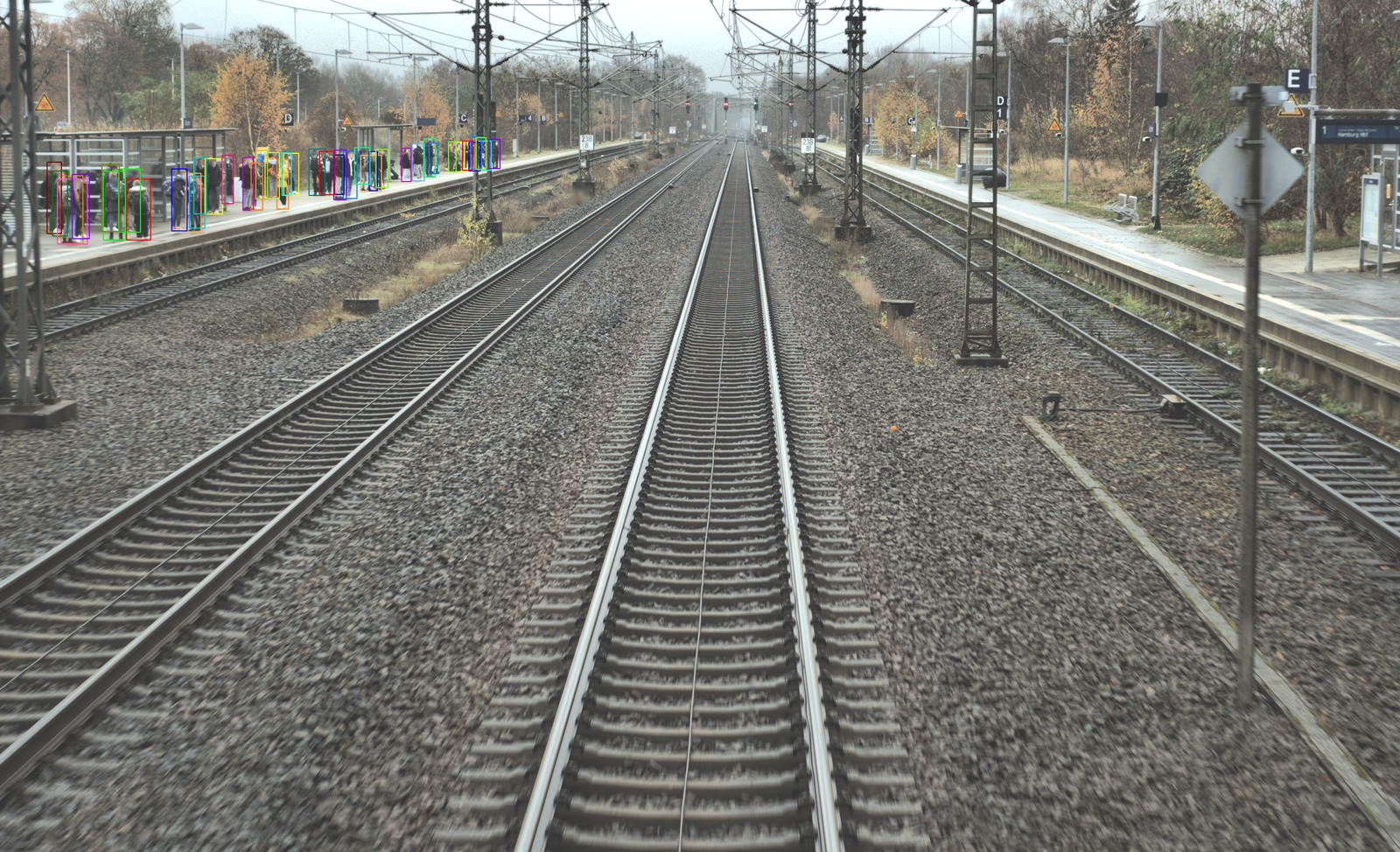} &
    \includegraphics[width=0.32\textwidth,keepaspectratio]{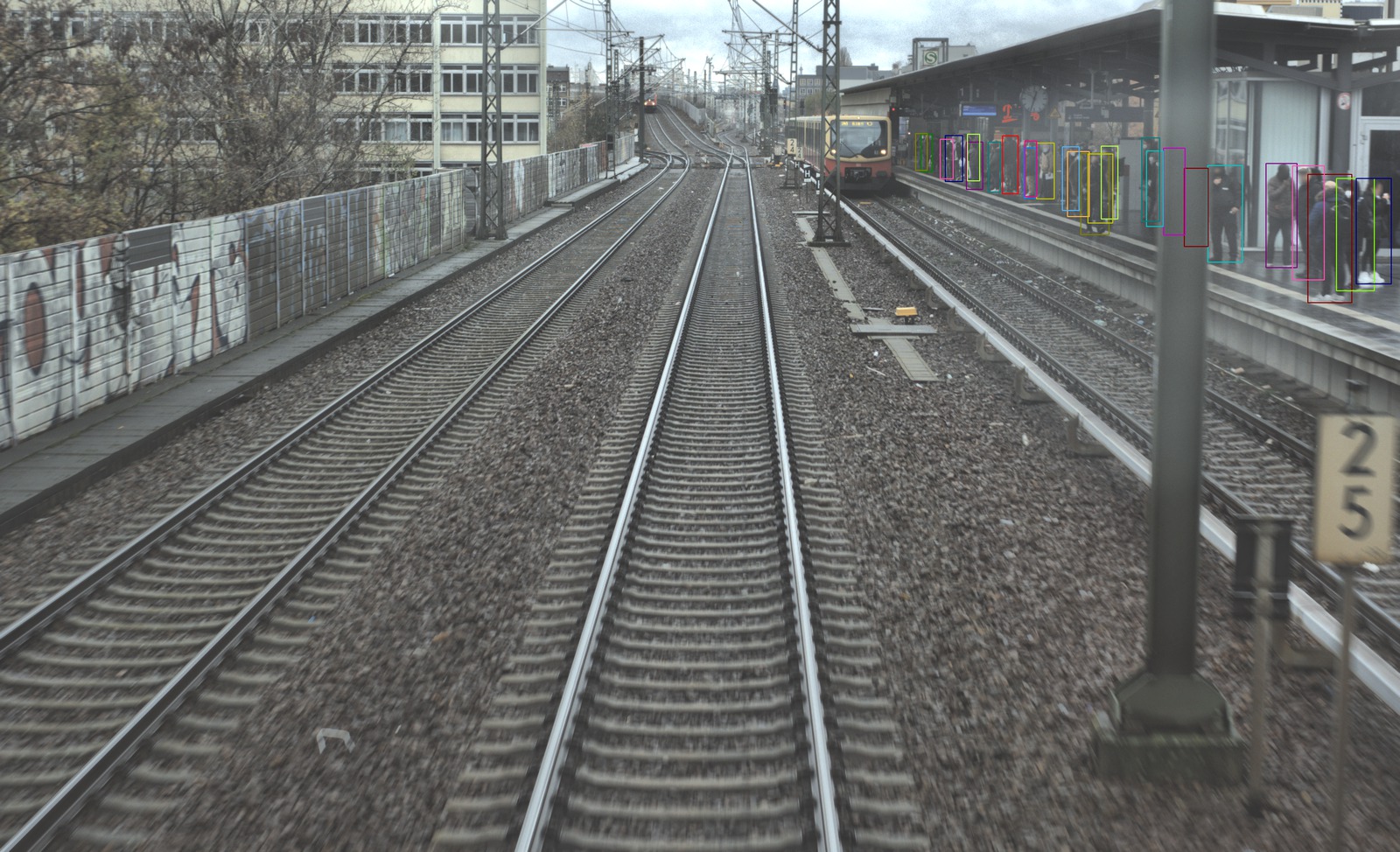} &
    \includegraphics[width=0.32\textwidth,keepaspectratio]{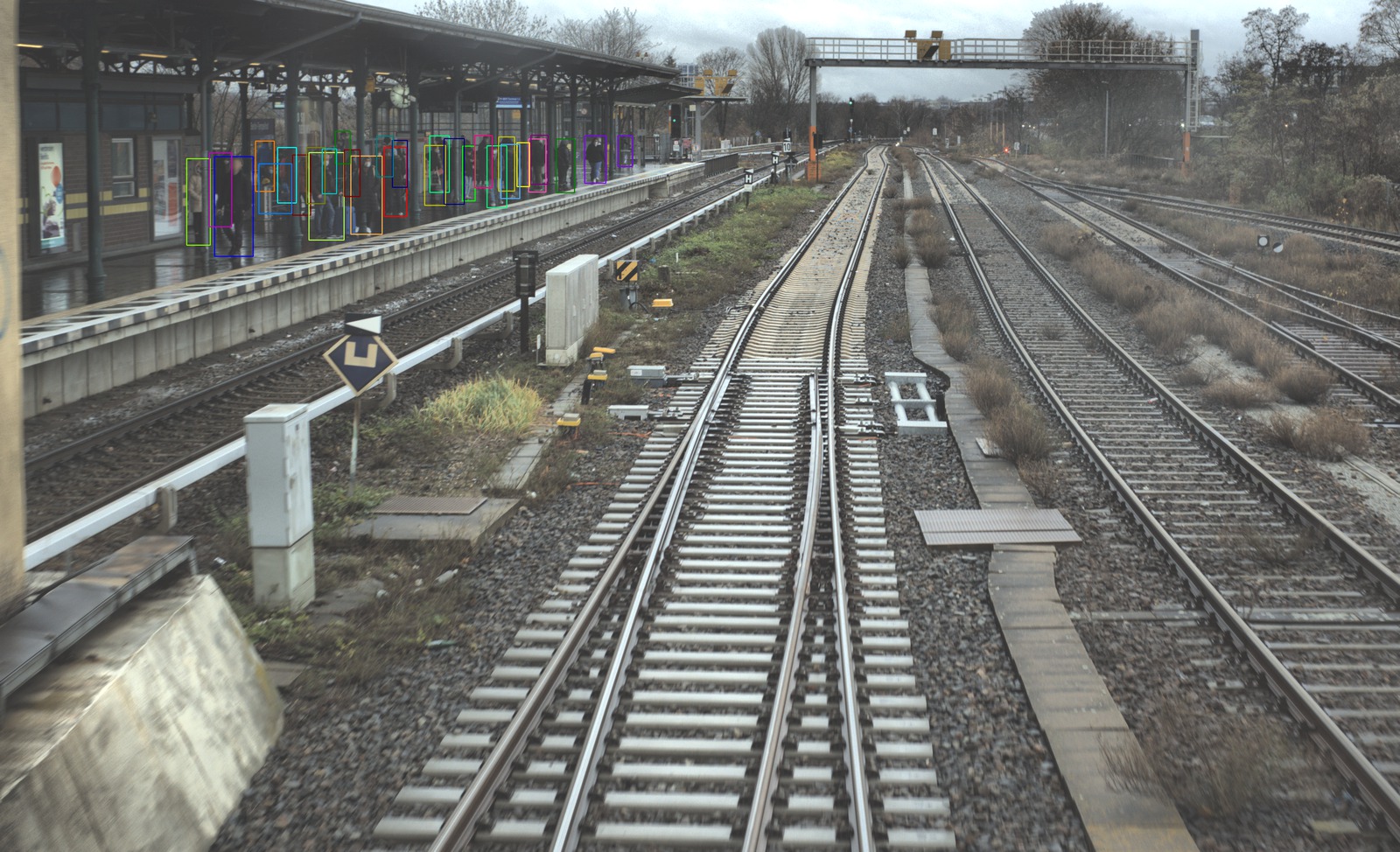} \\
  \end{tabular}
  \caption{Example frames from the RAIL-BENCH Tracking dataset.}
  \label{fig:anns_tracking}
\end{figure*}

\begin{figure*}[t]
  \centering
  \setlength{\tabcolsep}{2pt}
  \renewcommand{\arraystretch}{0}
  \begin{tabular}{ccc}
    \includegraphics[width=0.32\textwidth,keepaspectratio]{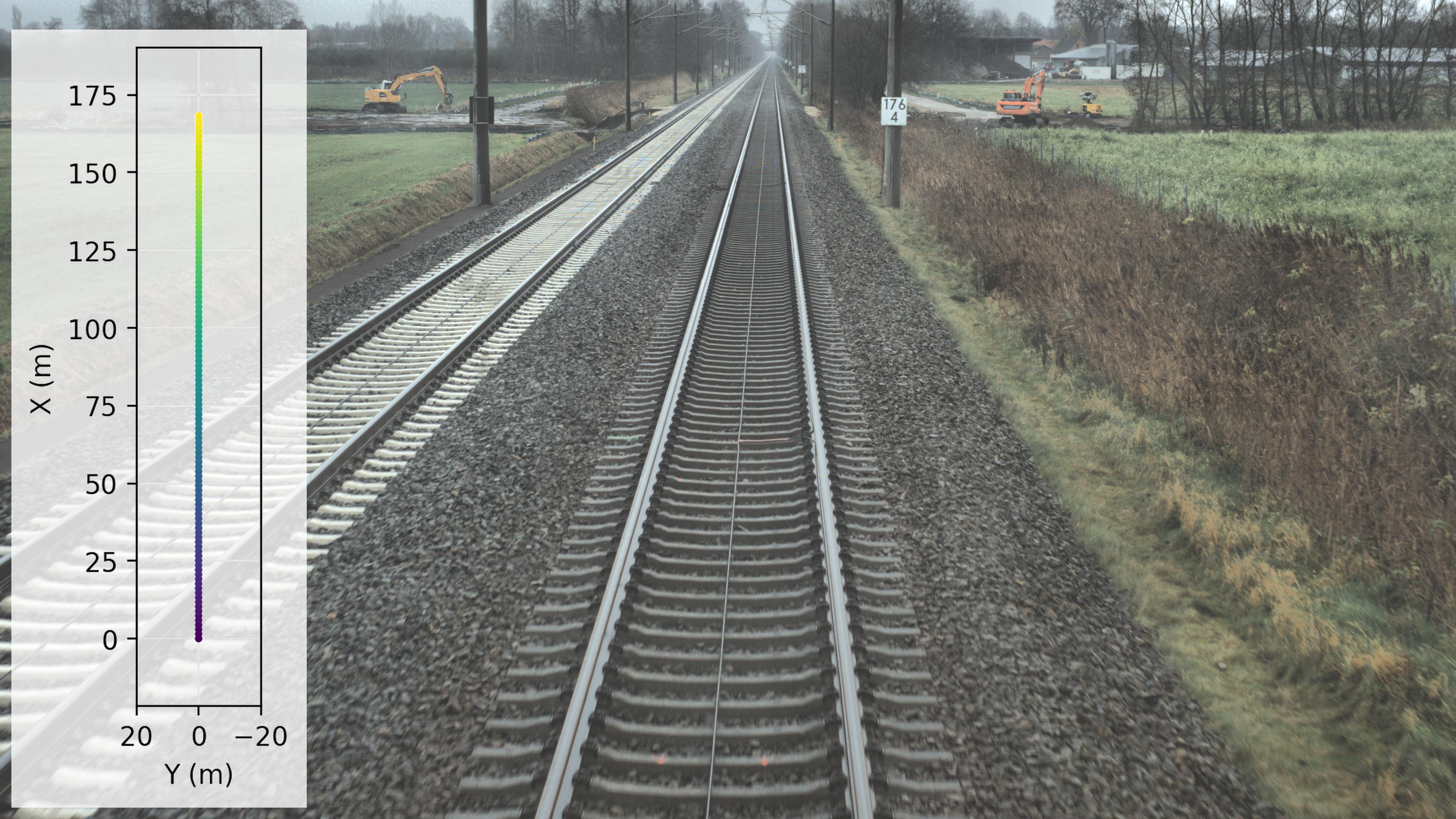} &
    \includegraphics[width=0.32\textwidth,keepaspectratio]{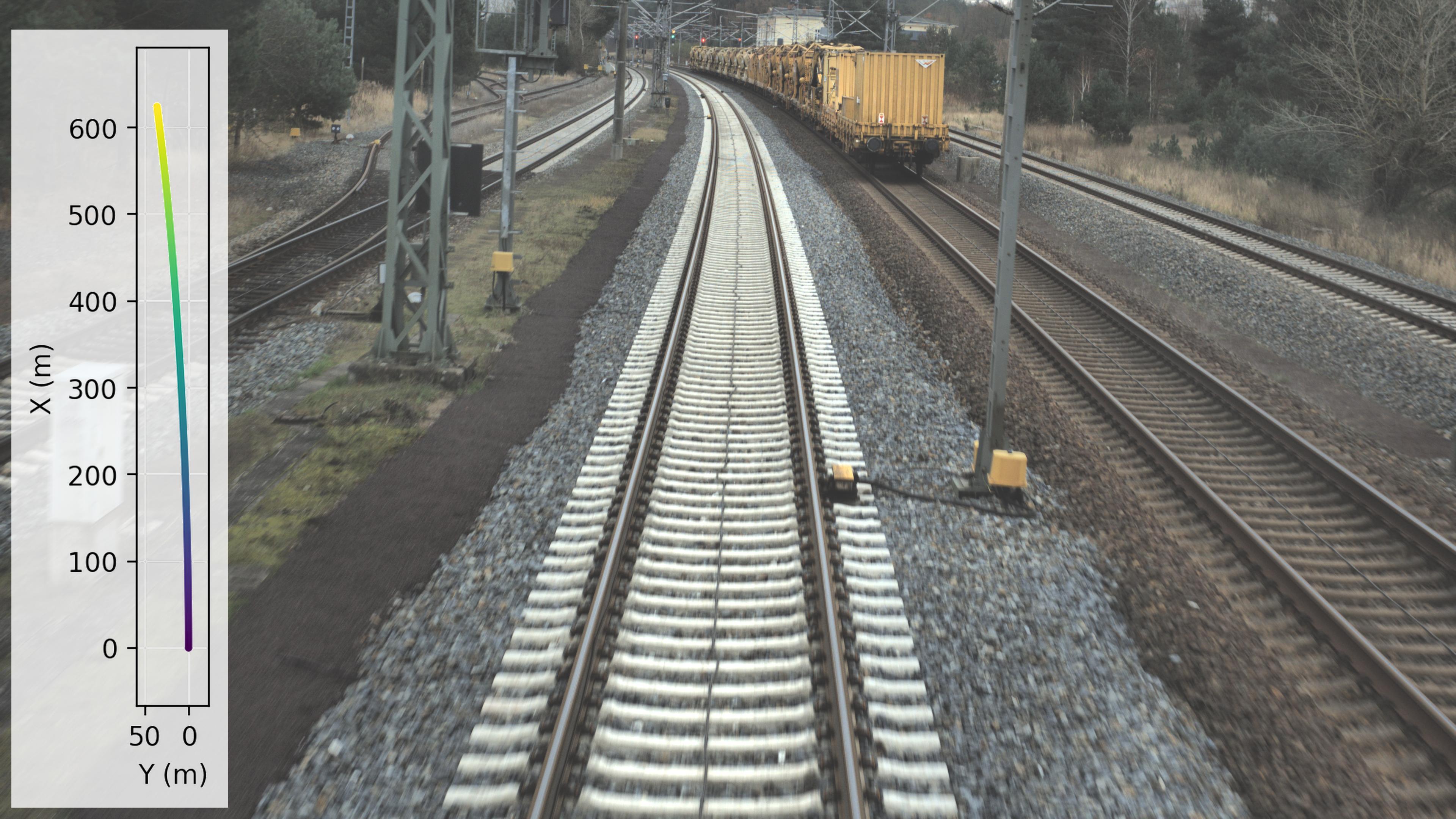} &
    \includegraphics[width=0.32\textwidth,keepaspectratio]{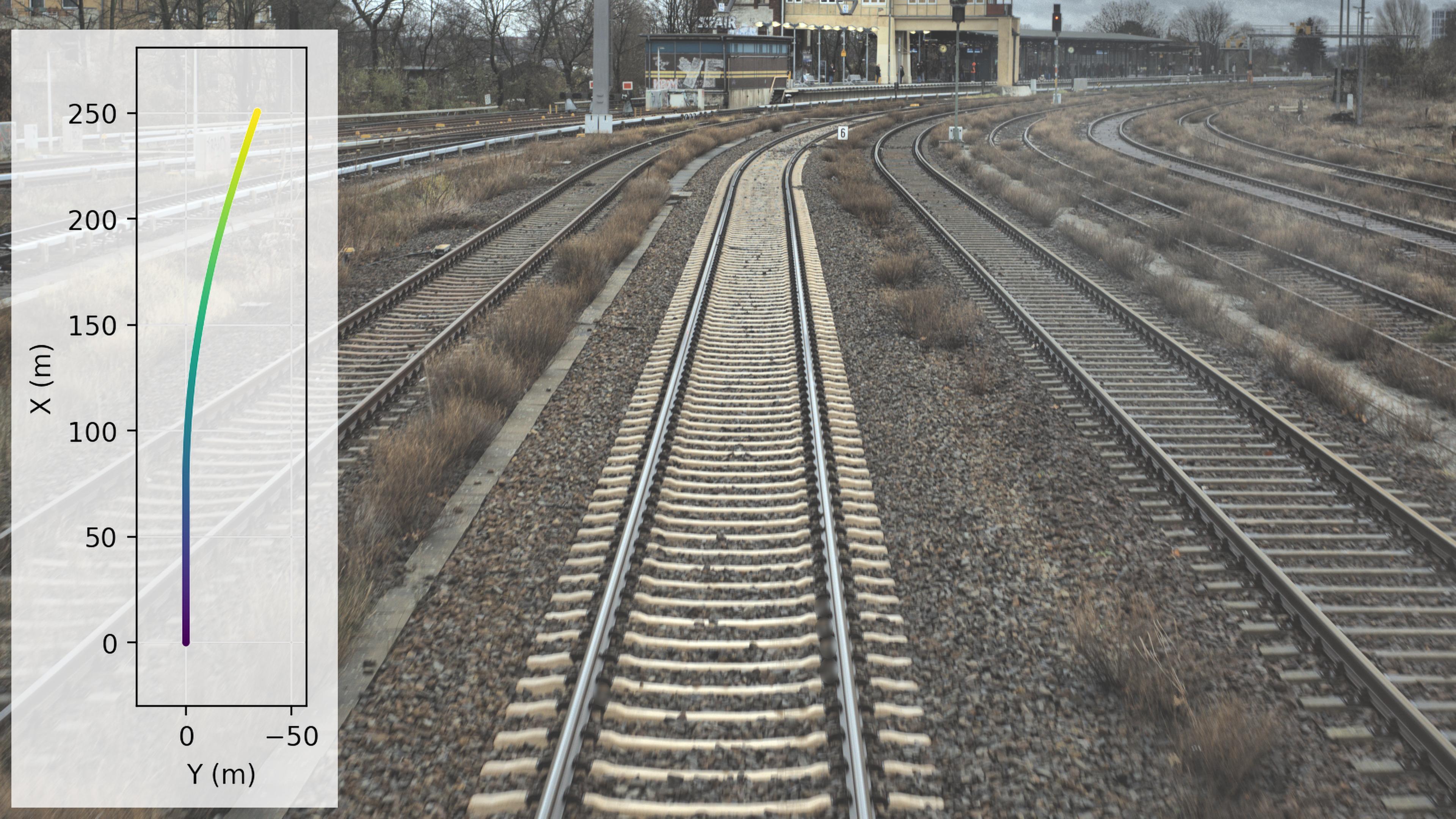} \\
  \end{tabular}
  \caption{Initial frames from three sequences of the RAIL-BENCH Odometry dataset and the corresponding ground truth trajectories of the cameras within each scene.}
  \label{fig:anns_odometry}
\end{figure*}

\subsection{RAIL-BENCH Tracking}
\label{sec:supp_tracking}

Among existing railway datasets with annotations for the class \textit{person}, only OSDaR23~\cite{tagiew_osdar23open_2023} and OSDaR26~\cite{osdar26} can be regarded as suitable for object tracking since they provide persistent object identifiers (IDs) across consecutive frames.
From the additional material provided by the authors of \mbox{OSDaR23} and OSDaR26, we selected four sequences, each containing between $179$ and $249$ frames and recorded at $10$~fps. In all scenes, the ego-vehicle approaches a train station with a large number of persons. All persons were annotated and assigned identifiers that are consistent over the whole sequence. As in the RAIL-BENCH Object+Rail dataset, persons that cannot be annotated individually, because they are too closely packed or too far in the background, are annotated collectively as a single bounding box flagged as \textit{iscrowd}. During evaluation, unmatched predictions that lie more than $50\,\%$ within such a crowd box are ignored, i.e., they are not counted as false positives. These scenes constitute the RAIL-BENCH Tracking dataset (see Tab.~\ref{tab:tracking}).

The RAIL-BENCH Tracking challenge is defined analogous to standard multi-object tracking benchmarks: the objective is to accurately associate person detections across frames within each video sequence. As persons on the platforms appear densely packed and at small scale, multi-object tracking is particularly challenging in our setting (see Fig.~\ref{fig:anns_tracking}).

Due to the comparatively small size of the RAIL-BENCH Tracking dataset, it is used exclusively for evaluation, and no corresponding training split is provided. For evaluation, we use the HOTA~\cite{luiten_hotahigher_2021} metric, which jointly balances detection and association accuracy into a single score and is widely adopted for ranking multi-object tracking predictions.

\subsection{RAIL-BENCH Odometry}
\label{sec:supp_odometry}

Only a few visual odometry datasets for the railway domain are currently available (see Tab.~\ref{tab:odo}). Nordland~\cite{Nordland} originates from a $728\,\mathrm{km}$ long train ride in Norway, recorded four times in different seasons. The dataset does not include the orientation of the train nor the intrinsic and extrinsic camera parameters. OSDaR23 and OSDaR26 provide detailed pose information and camera parameters.

For the RAIL-BENCH Odometry dataset, we selected scenes from OSDaR26 and from the additional data provided by its authors. We excluded scenes with physically implausible trajectories, as well as low velocity scenarios, i.e., average velocity below $2\,\mathrm{m/s}$ or distance traveled below $2\,\mathrm{m}$. In addition, we considered only daylight scenarios. The resulting dataset comprises $50$ distinct monocular video sequences, with a total trajectory length of $11.73\,\mathrm{km}$, corresponding to $16.39\,\mathrm{min}$ of travel time. Ground truth poses of the camera were computed based on the recorded GNSS and IMU measurements and the extrinsic camera parameters. The final RAIL-BENCH Odometry dataset is split into $30$ training and $20$ test sequences (see Fig.~\ref{fig:anns_odometry}).

The RAIL-BENCH Odometry challenge requires the estimation of camera ego-motion from the provided monocular video sequences. Although the motion of railway vehicles is constrained to the rails, large slip as well as unknown branching directions at turnouts pose considerable difficulties for pose estimation.

For evaluation, we align the estimated trajectory with the ground truth using Sim(3) Umeyama alignment~\cite{umeyama_leastsquaresestimation_1991} and then compute the Absolute Trajectory Error (ATE) and the Relative Trajectory Error (RTE)~\cite{sturm_benchmarkevaluation_2012}.

\begin{table}[t]
\centering
\caption{Public railway datasets for multi-object tracking.}
{\scriptsize
\renewcommand{\arraystretch}{1.1}
\setlength{\tabcolsep}{3pt}
\resizebox{\columnwidth}{!}{%
\begin{tabular}{lcccc}
\toprule

 & \multirow{2}{*}{\#scenes} & \multirow{2}{*}{\shortstack{fps\\(min - max)}} & \multirow{2}{*}{\shortstack{\#frames scene\\(min - max)}} & \multirow{2}{*}{\shortstack{scene length (s)\\(min - max)}}\\
 \\
\midrule
OSDaR23~\cite{tagiew_osdar23open_2023} & $45$ & $10$ &  $10 - 100^{*1}$ & $1-10$  \\
OSDaR26~\cite{osdar26} & $77$ & $4.5 - 10$ & $3-299^{*1}$ & $0.3-29.9$  \\
\midrule
RAIL-BENCH Tracking & $4$ & $10$ & $179 - 249$ & $17.9 - 24.9$ \\
\midrule
\addlinespace[2pt]
\multicolumn{5}{l}{\scriptsize $^{*1}$ \textit{Given in m-frames (multi-sensor frames), i.e., time steps}} \\
\multicolumn{5}{l}{\scriptsize \phantom{$^{*1}$} \textit{at which several sensors recorded simultaneously.}} \\
\bottomrule
\end{tabular}%
}
}

\label{tab:tracking}
\end{table}

\begin{table}[t]
\centering
\caption{Public railway datasets for visual odometry.}
{\scriptsize
\renewcommand{\arraystretch}{1.1}
\setlength{\tabcolsep}{3pt}
\resizebox{\columnwidth}{!}{%
\begin{tabular}{@{}l*{8}{c}@{}}
\toprule
 & \multicolumn{3}{c}{Ground truth} & \multicolumn{5}{c}{Sequences} \\
\cmidrule(lr){2-4}\cmidrule(lr){5-9}

  & \multirow{3}{*}{\rotatebox[origin=c]{90}{Position}} & \multirow{3}{*}{\rotatebox[origin=c]{90}{Rotation}} & \multirow{3}{*}{\rotatebox[origin=c]{90}{\shortstack{Camera\\[-0.2ex]Intrinsics}}} & \multirow{3}{*}{\rotatebox[origin=c]{90}{Filtered}} & \multirow{3}{*}{\shortstack{Total\\[-0.2ex]Traveled\\[-0.2ex]Distance}} & \multirow{3}{*}{\shortstack{Total\\[-0.2ex]Recorded\\[-0.2ex]Time}} & \multirow{3}{*}{\# Scenes} & \multirow{3}{*}{\# Frames} \\
  \\
  \\
  
\midrule
Nordland~\cite{Nordland} & \cmark & \cross &  \cross & \cmark & $\sim2{,}912$ km & $\sim40$ h & $4$ & $27{,}592$ \\
OSDaR23~\cite{tagiew_osdar23open_2023} & \cmark & \cmark &  \cmark & \cross & $\sim426$ m & $\sim154$ s & 45 & $1{,}534$ m-frames$^{*1}$ \\
OSDaR26~\cite{osdar26} & \cmark & \cmark &  \cmark & \cross & unknown & $\sim29$ min & $77$ & $15{,}646$ m-frames$^{*1}$ \\
\midrule
\multirow{2}{*}{\shortstack{RAIL-BENCH\\Odometry$^{*2}$}} & \multirow{2}{*}{\cmark} & \multirow{2}{*}{\cmark} & \multirow{2}{*}{\cmark} & \multirow{2}{*}{\cmark} & \multirow{2}{*}{$11.73$ km} & \multirow{2}{*}{$16.39$ min}  & \multirow{2}{*}{$50$}  & \multirow{2}{*}{$8{,}720$} \\
\\
\midrule
\addlinespace[2pt]
\multicolumn{9}{l}{\scriptsize $^{*1}$ \textit{M-frame stands for multi-sensor frame, which is the collection of several sensor frames.}} \\
\multicolumn{9}{l}{\scriptsize $^{*2}$ \textit{Contains sequences sampled from OSDaR26.}} \\
\bottomrule
\end{tabular}%
}
}

\label{tab:odo}
\end{table}

\section{Additional RAIL-BENCH Dataset Details}

\begin{figure*}[t]
  \centering
  \setlength{\tabcolsep}{2pt}
  \renewcommand{\arraystretch}{0}
  \begin{tabular}{ccc}
    \includegraphics[height=0.18\textwidth,keepaspectratio]{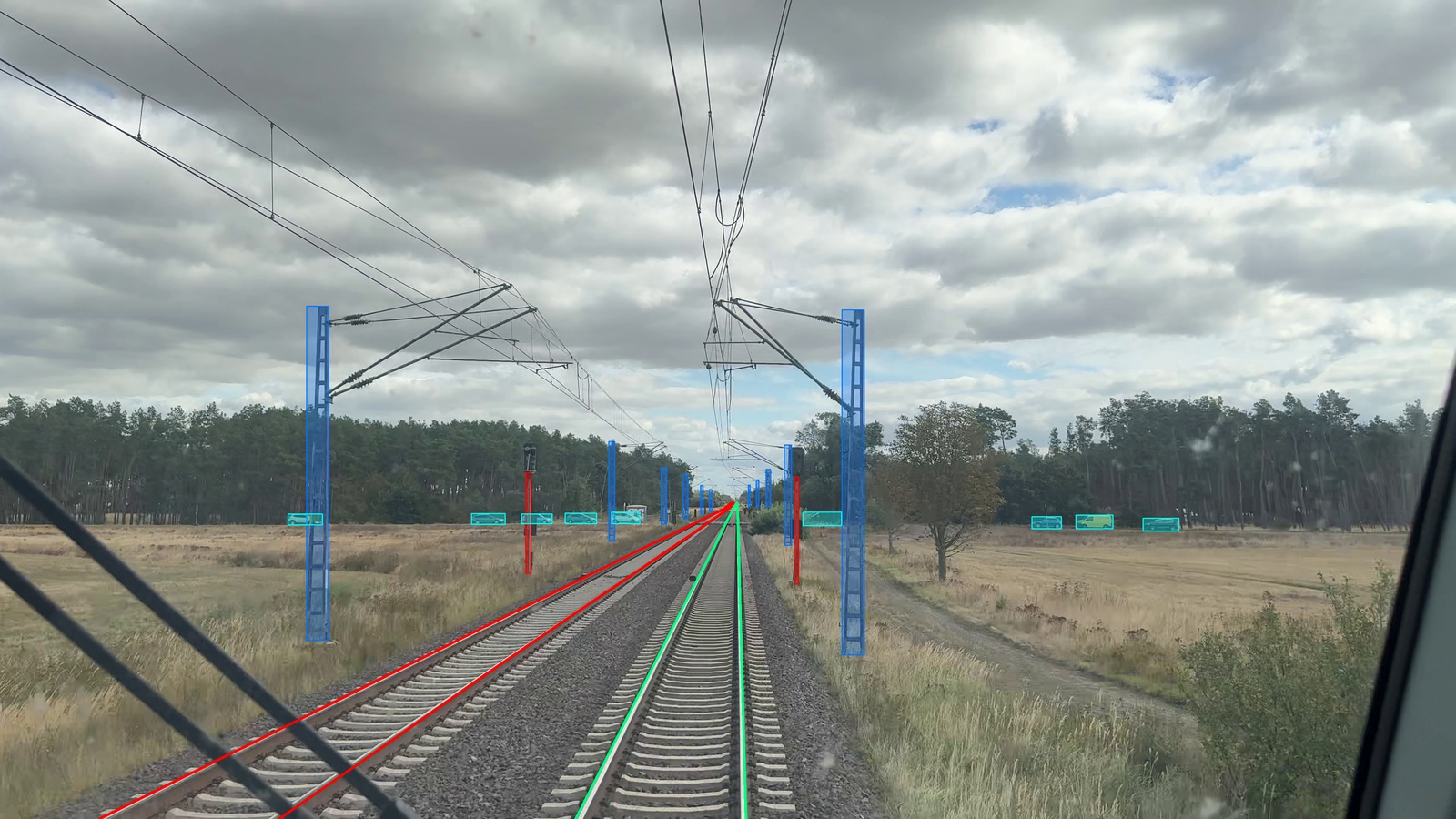} &
    \includegraphics[height=0.18\textwidth,keepaspectratio]{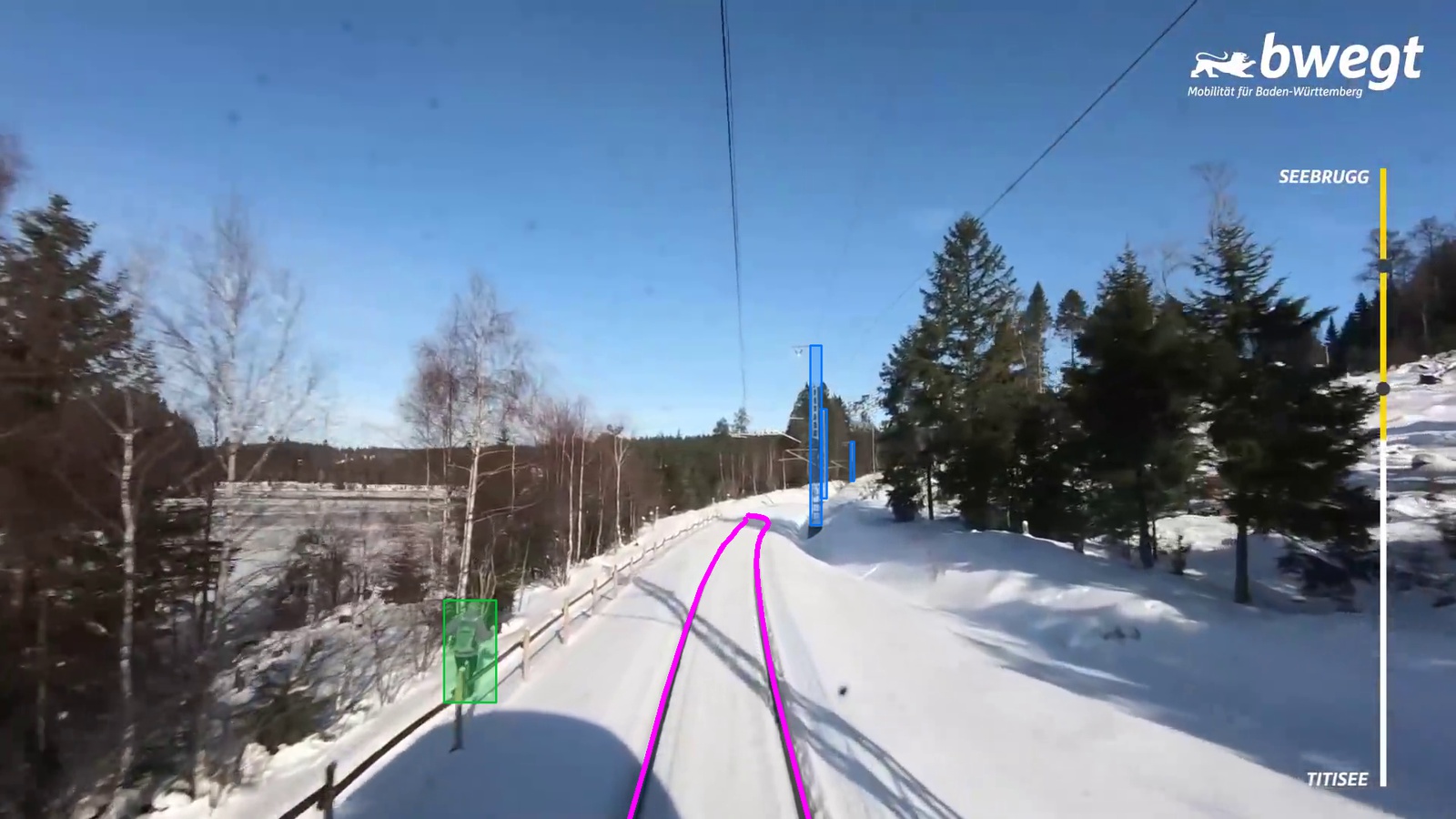} &
    \includegraphics[height=0.18\textwidth,keepaspectratio]{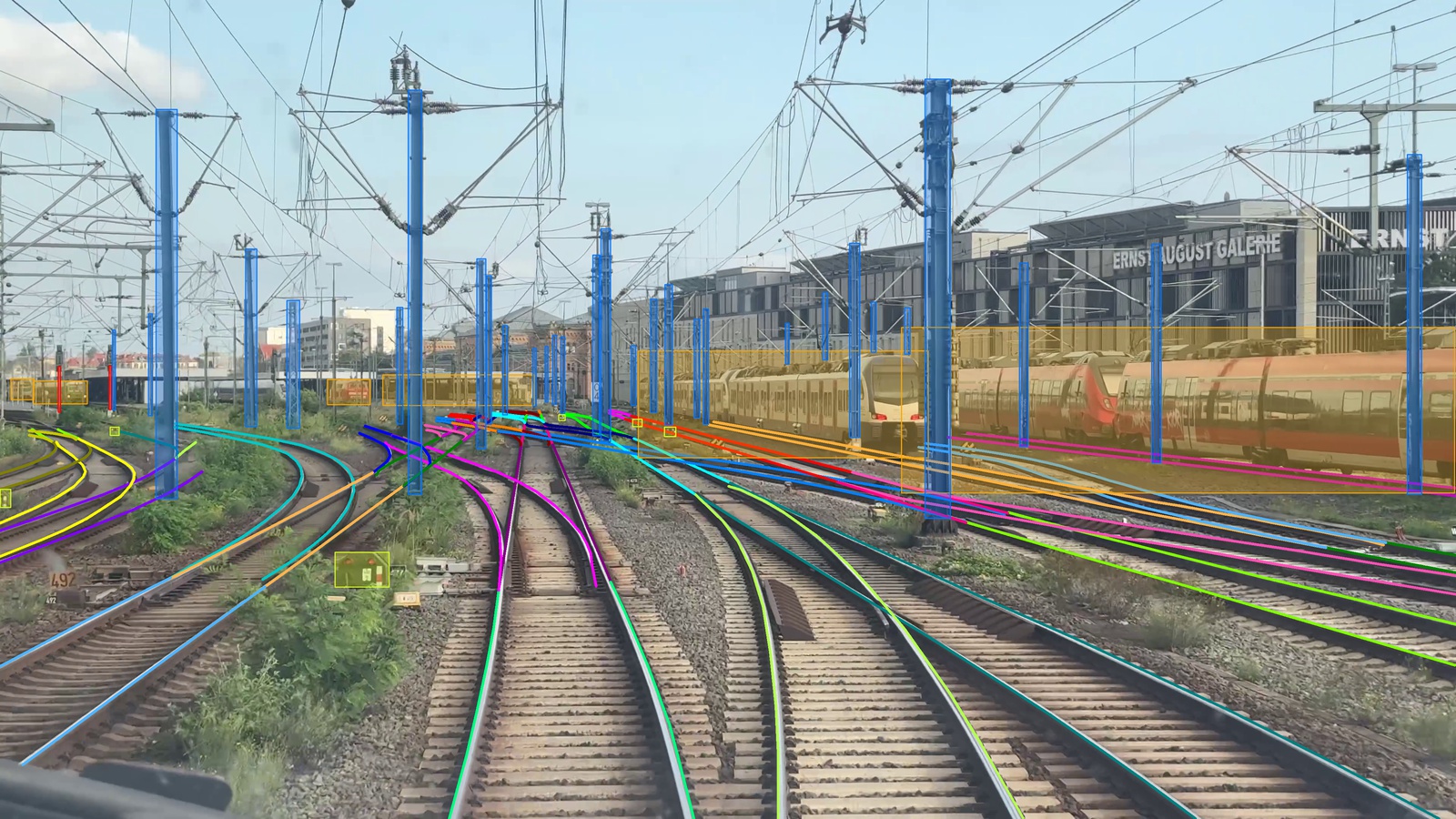} \\
    \includegraphics[height=0.18\textwidth,keepaspectratio]{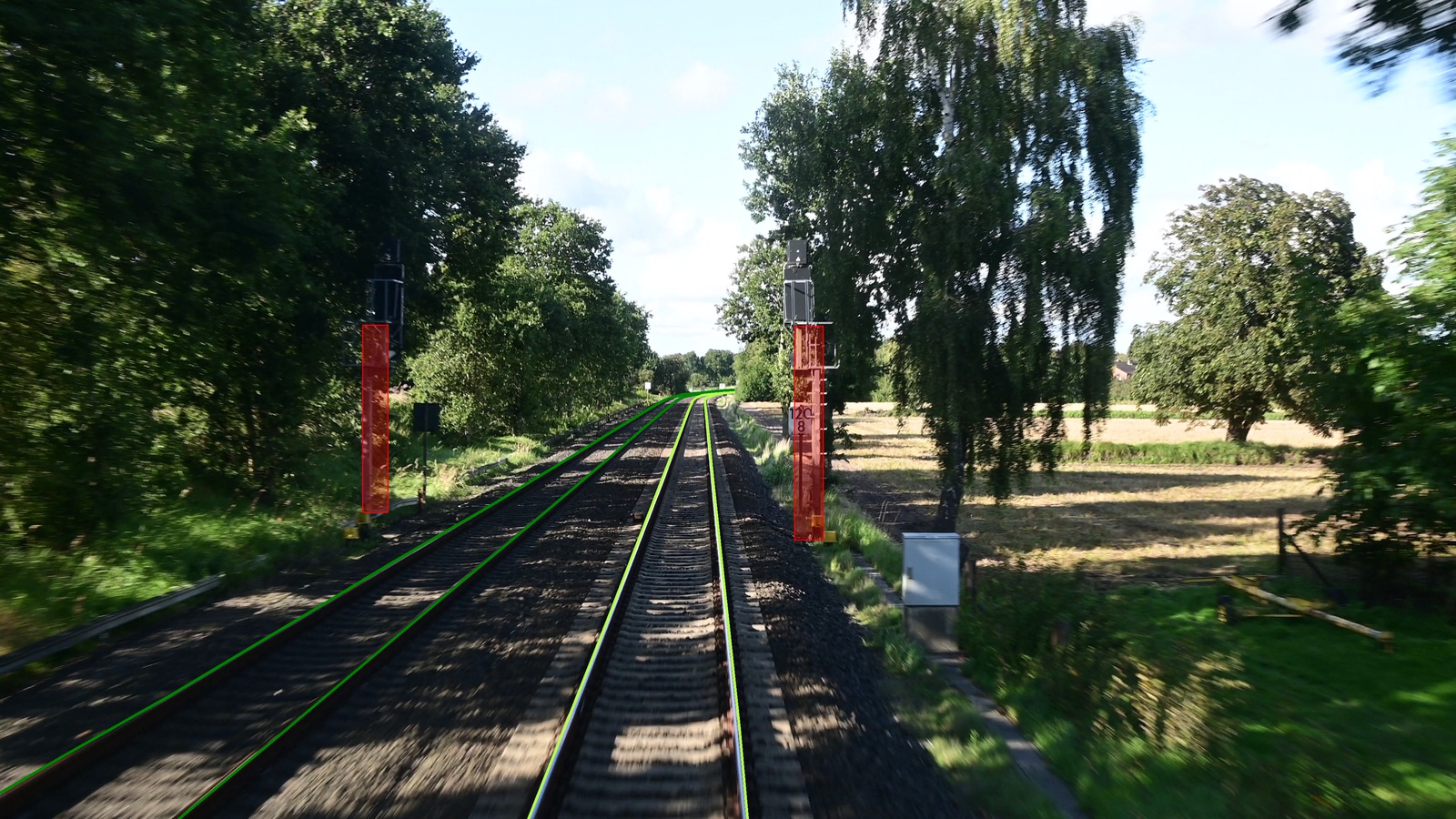} &
    \includegraphics[height=0.18\textwidth,keepaspectratio]{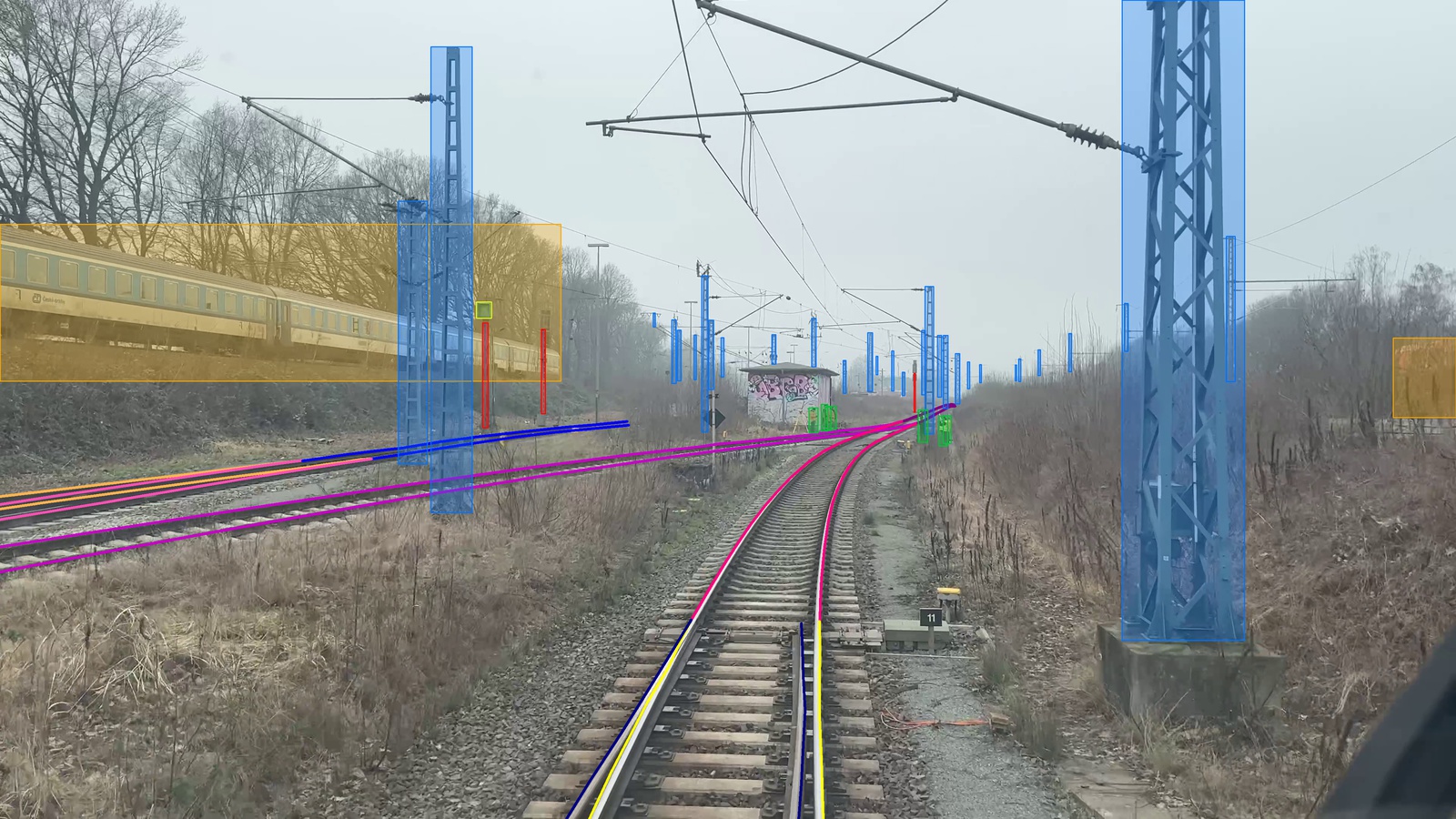} &
    \includegraphics[height=0.18\textwidth,keepaspectratio]{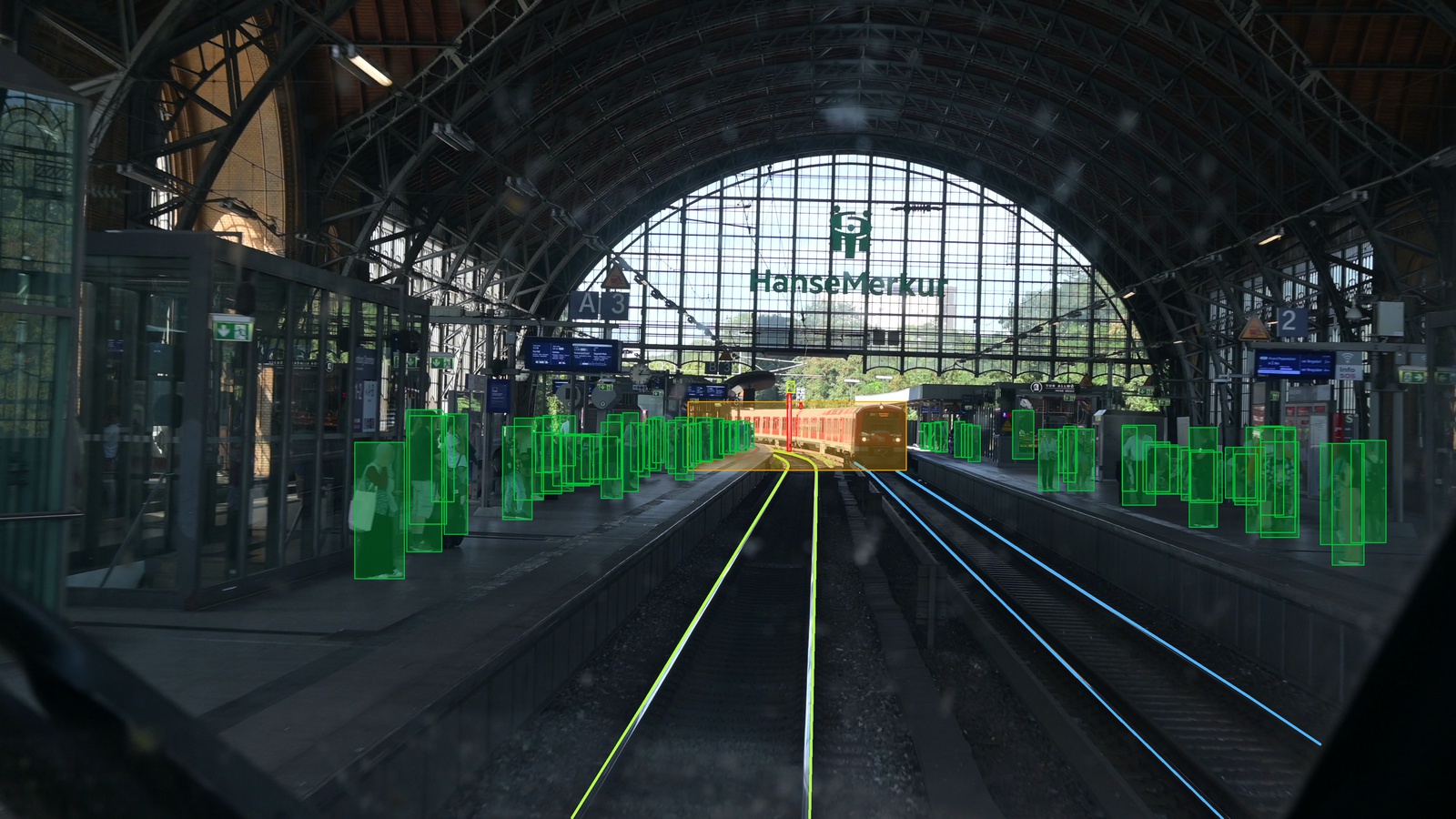} \\
  \end{tabular}
  \caption{Example images of the RAIL-BENCH Object+Rail dataset. Rails of the same track are displayed in the same color, while different tracks are displayed in different colors. Object classes are displayed in the colors introduced by Figure 2 in the main text.}
  \label{fig:anns_obj_rail}
\end{figure*}

\begin{figure*}[t]
  \centering
  \setlength{\tabcolsep}{2pt}
  \renewcommand{\arraystretch}{0}
  \begin{tabular}{ccc}
    \includegraphics[height=0.18\textwidth,keepaspectratio]{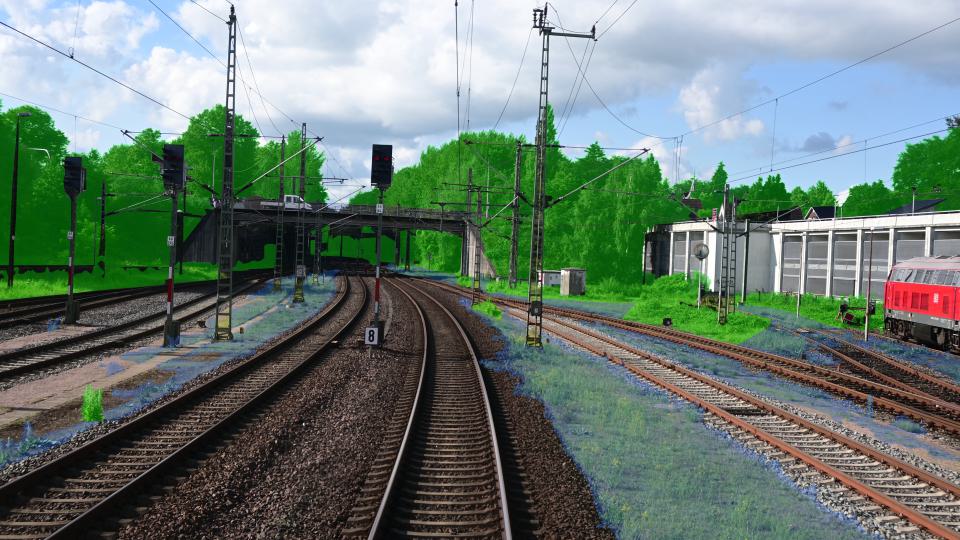} &
    \includegraphics[height=0.18\textwidth,keepaspectratio]{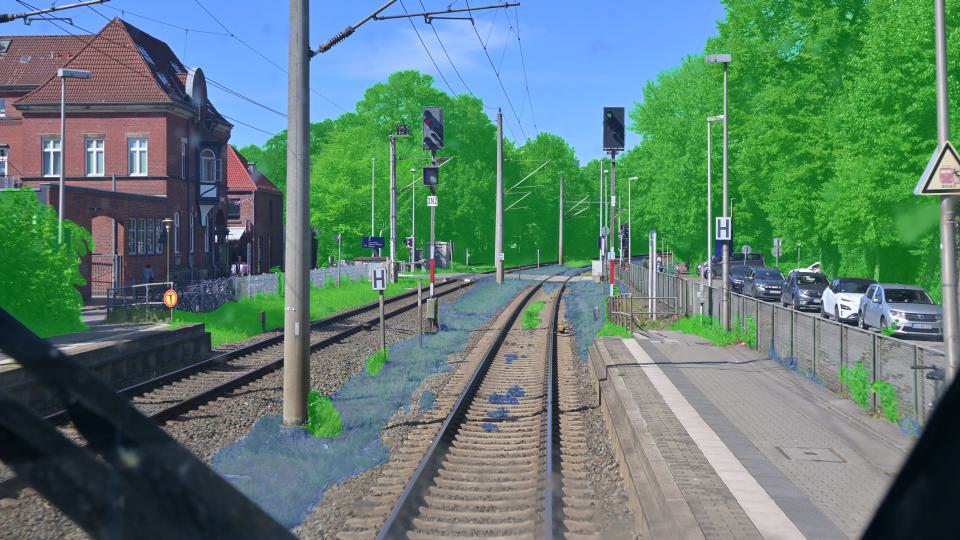} &
    \includegraphics[height=0.18\textwidth,keepaspectratio]{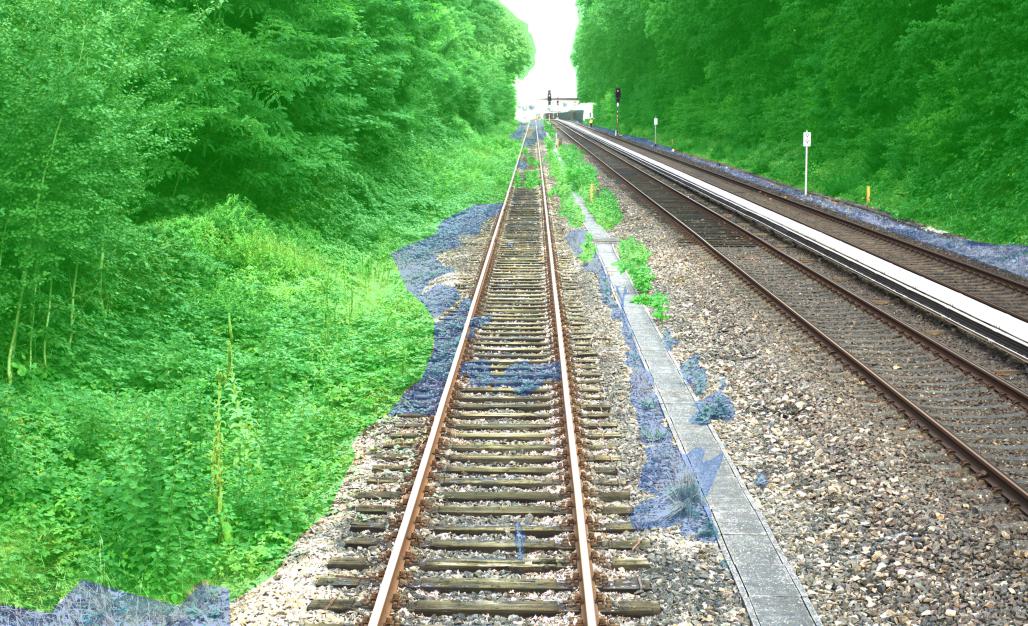} \\
    \includegraphics[height=0.18\textwidth,keepaspectratio]{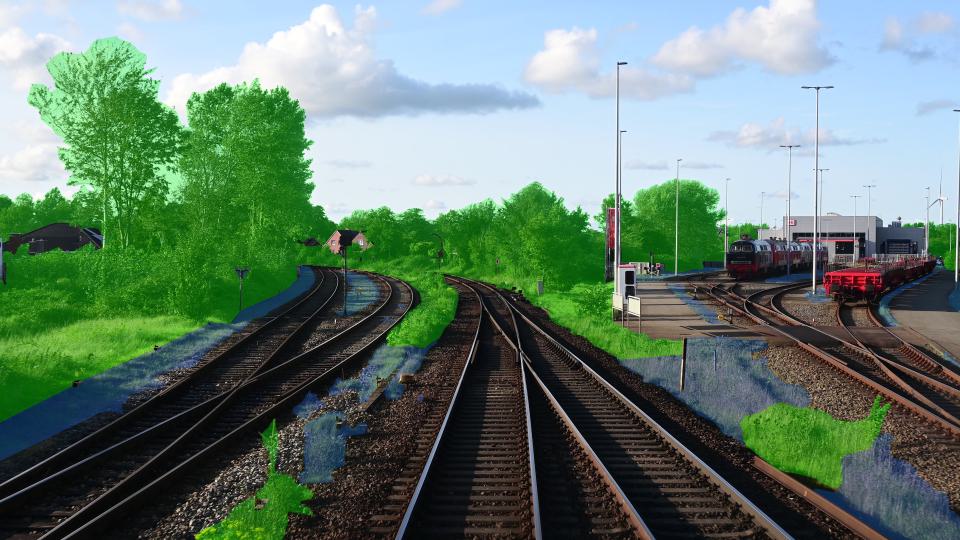} &
    \includegraphics[height=0.18\textwidth,keepaspectratio]{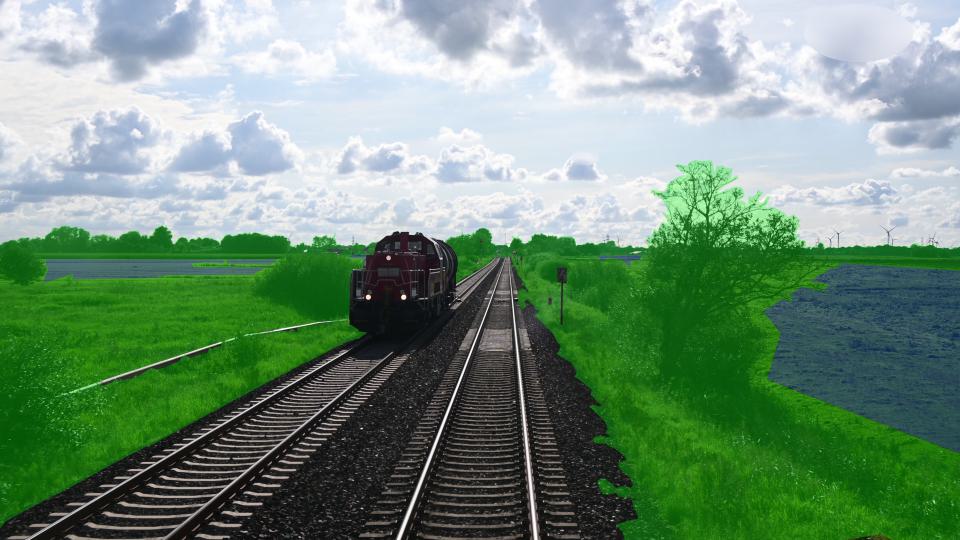} &
    \includegraphics[height=0.18\textwidth,keepaspectratio]{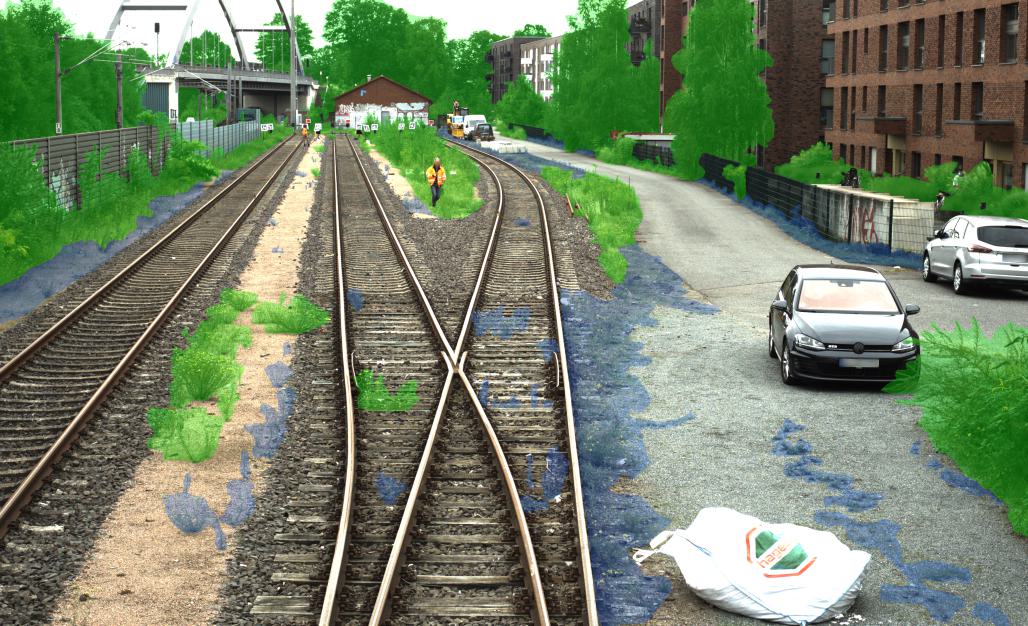} \\
  \end{tabular}
  \caption{Example images of the RAIL-BENCH Vegetation dataset. High-growing vegetation is displayed in green, while low-growing vegetation is displayed in blue.}
  \label{fig:anns_veg}
\end{figure*}

Consecutive frames of railway recordings can be highly redundant, in particular in rural areas with only one or two tracks. Therefore, the images of the RAIL-BENCH Object+Rail and Vegetation datasets were not sampled at fixed temporal intervals but selected manually from the available recordings. We deliberately sampled more densely from complex and highly relevant scenes to maximize diversity with respect to the intended task, prioritizing scenes with multiple tracks and switches, stations, railway workers and level crossings for RAIL-BENCH Object+Rail, and selecting scenes with varying vegetation types, densities, and vegetation within the track bed for RAIL-BENCH Vegetation. Figures~\ref{fig:anns_obj_rail} and \ref{fig:anns_veg} provide additional example images of the RAIL-BENCH Object+Rail and Vegetation datasets, respectively.

\begin{figure*}[t]
  \centering
  \setlength{\tabcolsep}{4pt}
  \renewcommand{\arraystretch}{0}
  \begin{tabular}{cc}
    \multicolumn{1}{c}{\small{(a) RAIL-BENCH Object+Rail}} &
    \multicolumn{1}{c}{\small{(b) RAIL-BENCH Vegetation}} \\
    \includegraphics[height=0.34\textwidth,keepaspectratio]{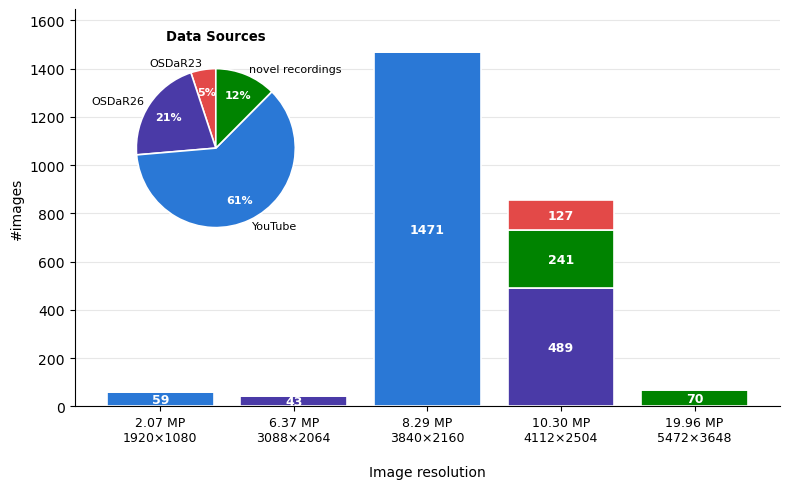} &
    \includegraphics[height=0.34\textwidth,keepaspectratio]{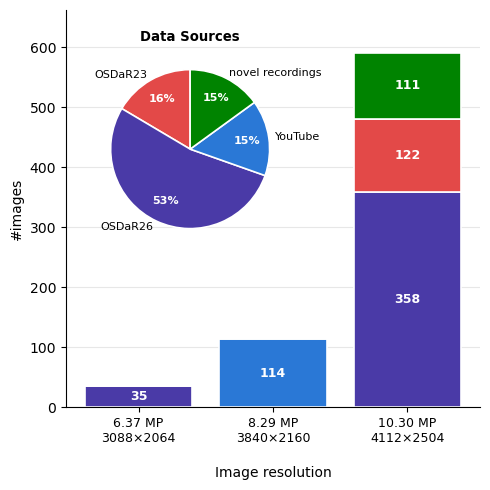} \\
  \end{tabular}
  \caption{Image resolution and data source distribution for the RAIL-BENCH Object+Rail (a) and Vegetation (b) datasets. \textit{Novel recordings} refers to previously unpublished recordings provided by the authors of OSDaR23 and OSDaR26 specifically for the RAIL-BENCH datasets; \textit{YouTube} refers to video frames collected from various YouTube channels, which likewise have not previously been published with annotations.}
  \label{fig:dataset_sources}
\end{figure*}

Figure~\ref{fig:dataset_sources} shows the image resolution and data source distribution for both datasets. By combining images from the OSDaR23 and OSDaR26 datasets with previously unpublished data, we ensure that the RAIL-BENCH datasets contain a diverse set of images that are representative of real-world railway scenarios. With the majority of the images having a resolution of at least $6.37$ megapixels, the RAIL-BENCH datasets provide higher resolution images than many existing datasets, which is important for the detection of small objects and fine details in railway environments.

\section{Details on Experimental Settings}
\subsection{Object Detection}

For YOLO11L and YOLOv8L-World, we use the implementations and pretrained weights provided by Ultralytics~\cite{yolo11_ultralytics, yolo_world}. For DINO-DETR~\cite{dino_detr}, we use the 4-scale deformable-attention variant with a ResNet-50 backbone and pre-trained weights provided by the mmdetection model zoo~\cite{mmdetection}. 

For predictions with YOLO11L and DINO-DETR pretrained on the COCO dataset~\cite{coco_detection_challenge_codalab}, we map the COCO object categories to the RAIL-BENCH classes, e.g., the COCO class \textit{traffic light} is mapped to the RAIL-BENCH class \textit{signal}. All RAIL-BENCH object categories can be covered in this way, except the \textit{signal pole} and \textit{catenary pole} categories. The full mapping is listed in Table~\ref{tab:coco_mapping}.

YOLO11L is finetuned on the RAIL-BENCH Object dataset for $60$ epochs with standard Ultralytics~\cite{yolo11_ultralytics} parameters with the exception of the initial learning rate (lr), which was set to $1\times10^{-4}$. DINO-DETR is finetuned on the RAIL-BENCH Object dataset for $45$ epochs with the same parameters as in the original DINO-DETR code release~\cite{dino_detr}, e.g. $\text{lr} = 1\times10^{-4}$.

YOLOv8L-World is used in a zero-shot setting, where the model is prompted with text prompts, such as \textit{tall pole with overhead wires} for the class \textit{catenary pole} (see Tab.~\ref{tab:prompt_list}). These prompts are derived empirically based on exploratory experiments conducted on the training and validation splits of the dataset. 

The three difficulty levels \textit{easy}, \textit{moderate}, and \textit{hard} used for evaluation are defined based on the size and occlusion attributes of each ground-truth bounding box. The \textit{easy} level considers only bounding boxes with an area of at least $50^2$ pixels and an occlusion below $25\,\%$, the \textit{moderate} level considers bounding boxes with an area of at least $30^2$ pixels and an occlusion below $75\,\%$, and the \textit{hard} level considers all bounding boxes regardless of size and occlusion. Following COCO~\cite{coco_detection_challenge_codalab}, ground-truth boxes outside the respective range are not removed but ignored: predictions matched to them are counted neither as true nor as false positives, and unmatched predictions outside the size range are likewise ignored. Annotations flagged as \textit{ignore} are excluded in all levels.

\begin{table}[t]
\centering
\caption{Mapping from COCO categories to RAIL-BENCH object categories, used for the pretrained YOLO11L and DINO-DETR models.}
{\scriptsize
\renewcommand{\arraystretch}{1.1}
\setlength{\tabcolsep}{3pt}
\begin{tabular}{ll}
\toprule
RAIL-BENCH class &  COCO class(es) \\
\midrule
train &  6 (train) \\
catenary pole  & -- (no COCO equivalent) \\
signal pole  & -- (no COCO equivalent) \\
signal  & 9 (traffic light) \\
person  & 0 (person) \\
road vehicle  & 2, 5, 7 (car, bus, truck) \\
bicycle  & 1 (bicycle) \\
\bottomrule
\end{tabular}%
}
\label{tab:coco_mapping}
\end{table}

\begin{table}[t]
\centering
\caption{Text prompts used for YOLOv8L-World in the zero-shot setting.}
{\scriptsize
\renewcommand{\arraystretch}{1.1}
\setlength{\tabcolsep}{3pt}
\begin{tabular}{ll}
\toprule
RAIL-BENCH class & Text prompt(s) \\
\midrule
train & train \\
catenary pole & tall pole with overhead wires \\
signal pole & tall pole with traffic light \\
signal & railway signal showing light \\
person & person \\
road vehicle & car, bus, truck \\
bicycle & bicycle \\
\bottomrule
\end{tabular}%
}
\label{tab:prompt_list}
\end{table}

\subsection{Rail Track Detection}

All three rail track detection models are pretrained on the RailSem19 dataset~\cite{zendel_railsem19dataset_2019} and finetuned on the RAIL-BENCH Rail dataset. In an additional experiment, we also trained the models exclusively on RAIL-BENCH Rail. The training settings for all three models are summarized below.

PINet~\cite{ko_keypoints_2022} is pretrained, finetuned and trained exclusively on RAIL-BENCH Rail using a learning rate of $1\times10^{-3}$, for $55$, $35$ and $70$ epochs, respectively. The YOLinO~\cite{Meyer_2021_ICCV} model is pretrained for $30$ epochs with a learning rate of $1\times10^{-4}$, finetuned for $15$ epochs with a learning rate of $1\times10^{-5}$, and trained exclusively on RAIL-BENCH Rail for $55$ epochs with a learning rate of $1\times10^{-4}$. For \mbox{CLRNet}~\cite{zhengCLRNetCrossLayer2022a} we use a ResNet-18 backbone. The model is pretrained for $50$ epochs with a learning rate of $2\times10^{-3}$, finetuned on RAIL-BENCH Rail for $40$ epochs with a learning rate of $2\times10^{-4}$, and trained exclusively on RAIL-BENCH Rail for $70$ epochs with a learning rate of $2\times10^{-3}$. For all CLRNet training settings, we use the following loss weights: $w_{cls}=2$, $w_{xytl}=0.2$, $w_{LIoU}=2$, $w_{seg}=0.5$, and a non-maximum-suppression threshold of $5$.

While PINet and CLRNet directly return line instances, the raw output of YOLinO consists of a set of line segments. To process the YOLinO predictions, we applied a custom postprocessing procedure in which each line segment is represented as a node in a graph, and edges encode the distances between the start and end points of pairs of segments. A greedy selection strategy was used to choose the edges with the smallest weights, subject to the constraint that each node has at most one incoming and one outgoing edge. Given the final set of edges, the corresponding line segments were concatenated to form individual lines, which are smoothed afterwards.

For evaluation with ChamferAP~\cite{li_hdmapnetonline_2022} we use $50$ sample points along each predicted and ground truth line and set the minimal distance threshold $\Delta d_{\min}$ to $10~\text{px}$. For \mbox{LineAP} we define the segment length as $2\%$ of the respective image width, fix the orientation threshold to $10$° and choose a minimal distance threshold $\Delta d_{\min}$ of $5~\text{px}$. Following standard evaluation protocols, we limit the number of polyline predictions to the $100$ with the highest confidence scores for both LineAP and ChamferAP. For all metrics, including CULane~\cite{pan2026spatialdeepspatialcnn} and TuSimple~\cite{yoo2020endtoendlanemarkerdetection}, we preprocess predictions to cut off any parts of the predicted rails that reach into ignore areas.

TuSimple was designed for highway scenes with at most four lanes and cannot be applied to rail scenes without modification. We therefore keep the original evaluation protocol wherever possible and adapt it only where necessary to obtain a meaningful evaluation on our data; our goal is not to improve the metric itself. 
TuSimple compares a predicted and a ground truth rail via their point accuracy, i.e., the fraction of sampled image rows in which the predicted rail lies within the distance threshold of the ground truth rail. To account for the very short rails in our dataset, we compute this point accuracy only over the rows in which the ground truth rail is visible, so that rows in which both rails are absent do not count as correct. 
Ground truth and predicted rails are assigned greedily one-to-one in descending order of point accuracy, and an assigned prediction is considered correct if its point accuracy is at least $85\%$, as visualized in Fig.~\ref{fig:metrics-lineap-vs-tusimple}. The accuracy of an image is the mean point accuracy of the assigned pairs over all of its $N_{\text{gt}}$ ground truth rails, where ground truth rails without an assigned prediction contribute zero, and the reported accuracy is the mean over all images. 
In contrast to the original TuSimple evaluation, which normalizes by $\min(4, N_{\text{gt}})$ and discards the worst lane accuracy in images with more than four lanes, we normalize by the actual number of ground truth rails $N_{\text{gt}}$, since our images frequently contain more than four rails.

\subsection{Vegetation Segmentation}

For the vegetation baselines we use the models SegFormer-B2~\cite{xie_segformersimple_2021}, Mask2Former (tiny-sized version) with a Swin backbone~\cite{cheng2021mask2former}, and DeepLabV3+ with a ResNet-101 backbone~\cite{deeplabv3plus2018}, all pretrained on Cityscapes~\cite{cordts_cityscapesdataset_2016} -- either from the mmsegmentation model zoo~\cite{mmseg2020} or the Hugging Face Hub~\cite{wolf_transformers_2020}.

SegFormer-B2 and Mask2Former are finetuned on the RAIL-BENCH Vegetation dataset for up to $50$ epochs with a learning rate of $6\times10^{-6}$, selecting the best checkpoint via early stopping (patience of $5$ epochs) based on the validation mIoU. DeepLabV3+ is finetuned for $30$ epochs with a polynomial learning-rate schedule decaying from $1\times10^{-3}$ to $1\times10^{-4}$.

\section{CULane and TuSimple Scores for Rail Track Detection Baselines}

\begin{table*}[ht]
\centering
\caption{Additional CULane and TuSimple scores for the Rail Track Detection baselines presented in Table~\ref{tab:rail_baselines} of the main paper. Distance thresholds @$\alpha$ are given in percentage with respect to the image width. \textbf{Best results} are shown in bold, \underline{second-best results} are underlined.}
\label{tab:culane_tusimple_results}
{\scriptsize
\begin{tabular*}{0.8\textwidth}{@{\extracolsep{\fill}} l c c c c c c c c c c c c c}
\toprule
 & \multicolumn{6}{c}{CULane (F1)} & \multicolumn{6}{c}{TuSimple (Accuracy)} \\
\cmidrule(lr){2-7}\cmidrule(lr){8-13}
Model & @0.5 & @0.7 & @1 & @2 & @5 & mF1 & @0.3 & @0.5 & @0.7 & @1 & @2 & mAcc \\
\midrule
PINet (pretrained) & 16.6 & 23.6 & 34.7 & 48.2 & 54.3 & 35.5 & 46.7 & 58.2 & 61.0 & 62.3 & 63.8 & 58.4 \\
PINet (RB only) & 17.2 & 26.2 & 36.3 & 46.8 & 53.5 & 36.0 & 50.5 & 58.4 & 60.8 & 62.2 & 63.6 & 59.1 \\
PINet (finetuned) & 23.3 & 33.2 & 42.3 & 50.9 & 56.6 & 41.3 & 55.4 & \underline{62.2} & \underline{64.0} & \underline{65.0} & \underline{66.1} & \underline{62.6} \\
\midrule
YOLinO (pretrained) & 19.6 & 27.3 & 35.1 & 41.0 & 45.6 & 33.7 & 51.8 & 59.3 & 60.3 & 60.7 & 61.2 & 58.7 \\
YOLinO (RB only) & 30.0 & 33.8 & 37.5 & 41.9 & 46.6 & 37.9 & 56.4 & 59.0 & 59.7 & 60.1 & 60.7 & 59.2 \\
YOLinO (finetuned) & 34.8 & 38.0 & 41.3 & 45.7 & 50.0 & 42.0 & \underline{59.6} & 61.6 & 62.1 & 62.5 & 63.0 & 61.8 \\
\midrule
CLRNet (pretrained) & 25.7 & 36.2 & 45.9 & 53.6 & 57.4 & 43.8 & 51.7 & 59.3 & 60.5 & 61.2 & 61.9 & 58.9 \\
CLRNet (RB only) & \underline{35.5} & \underline{41.1} & \underline{47.4} & \textbf{54.5} & \textbf{59.5} & \underline{47.6} & \textbf{59.9} & \textbf{63.4} & \textbf{65.0} & \textbf{66.0} & \textbf{67.0} & \textbf{64.3} \\
CLRNet (finetuned) & \textbf{41.3} & \textbf{45.2} & \textbf{49.2} & \underline{54.2} & \underline{57.8} & \textbf{49.6} & 59.0 & 61.3 & 62.3 & 63.0 & 63.8 & 61.9 \\
\bottomrule
\end{tabular*}%
}
\end{table*}

In addition to the evaluation results with LineAP and ChamferAP presented in the main paper, we provide the evaluation scores of the CULane and TuSimple metrics in Table~\ref{tab:culane_tusimple_results}. For both metrics, we define distance thresholds relative to the respective image width, e.g. AP@1 takes a distance threshold of $1\%$ of the image width.

\begin{figure*}[!tb]
  \centering
  \setlength{\tabcolsep}{2pt}
  \renewcommand{\arraystretch}{0}
  \begin{tabular}{ccccc}
    & \multicolumn{1}{c}{\small\textbf{PINet (finetuned)}} &
    \multicolumn{1}{c}{\small\textbf{YOLinO (finetuned)}} &
    \multicolumn{1}{c}{\small\textbf{CLRNet (finetuned)}} \\[4pt]
    \rotatebox{90}{\parbox{2.7cm}{\centering\small\textbf{Ground Truth}\\\footnotesize(random colors)}} &
    \includegraphics[width=0.27\textwidth]{images/Metrics/gt/img_2013_test.jpg} &
    \includegraphics[width=0.27\textwidth]{images/Metrics/gt/img_2020_test.jpg} &
    \includegraphics[width=0.27\textwidth]{images/Metrics/gt/img_2004_test.jpg} \\
    \rotatebox{90}{\parbox{2.7cm}{\centering\small\textbf{Predictions}\\\footnotesize(random colors)}} &
    \includegraphics[width=0.27\textwidth]{images/Metrics/preds/img_2013_test_pinet_finetuned.jpg} &
    \includegraphics[width=0.27\textwidth]{images/Metrics/preds/img_2020_test_yolino_finetuned.jpg} &
    \includegraphics[width=0.27\textwidth]{images/Metrics/preds/img_2004_test_clrnet_finetuned.jpg} \\[4pt]
    \rotatebox{90}{\parbox{2.7cm}{\centering\small\textbf{easy}}} &
    \includegraphics[width=0.27\textwidth]{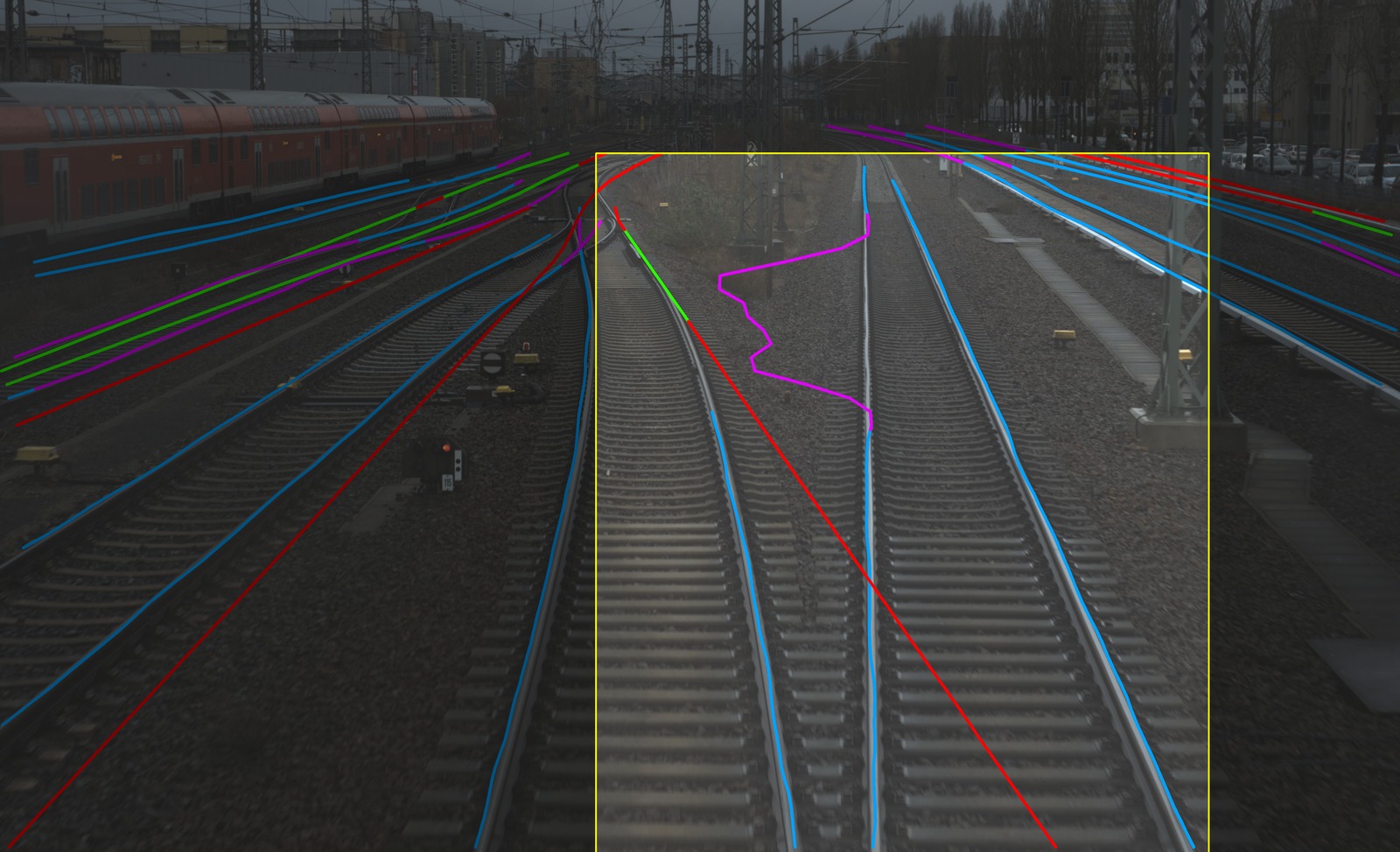} &
    \includegraphics[width=0.27\textwidth]{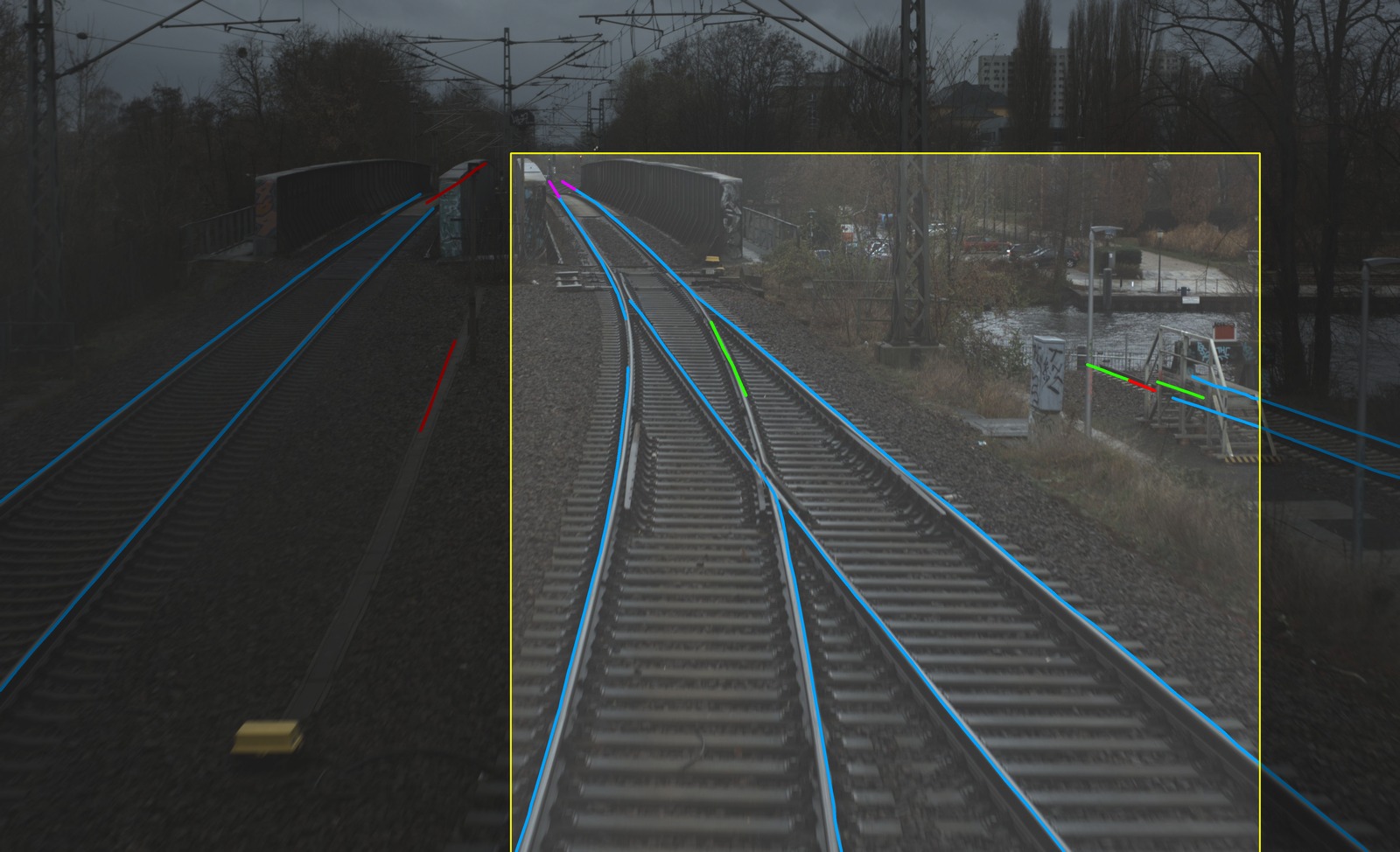} &
    \includegraphics[width=0.27\textwidth]{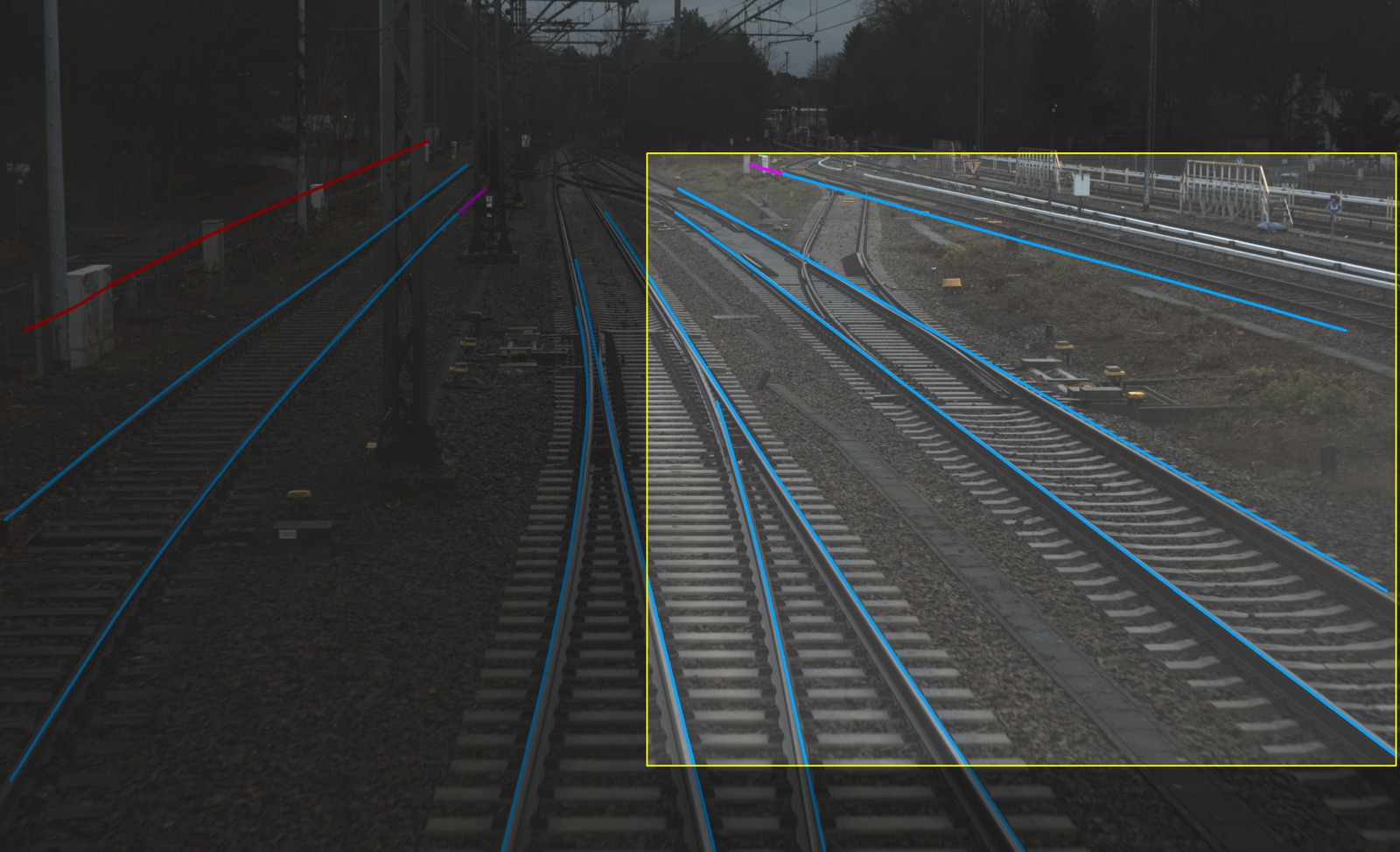} \\
    \rotatebox{90}{\parbox{2.7cm}{\centering\small\textbf{moderate}}} &
    \includegraphics[width=0.27\textwidth]{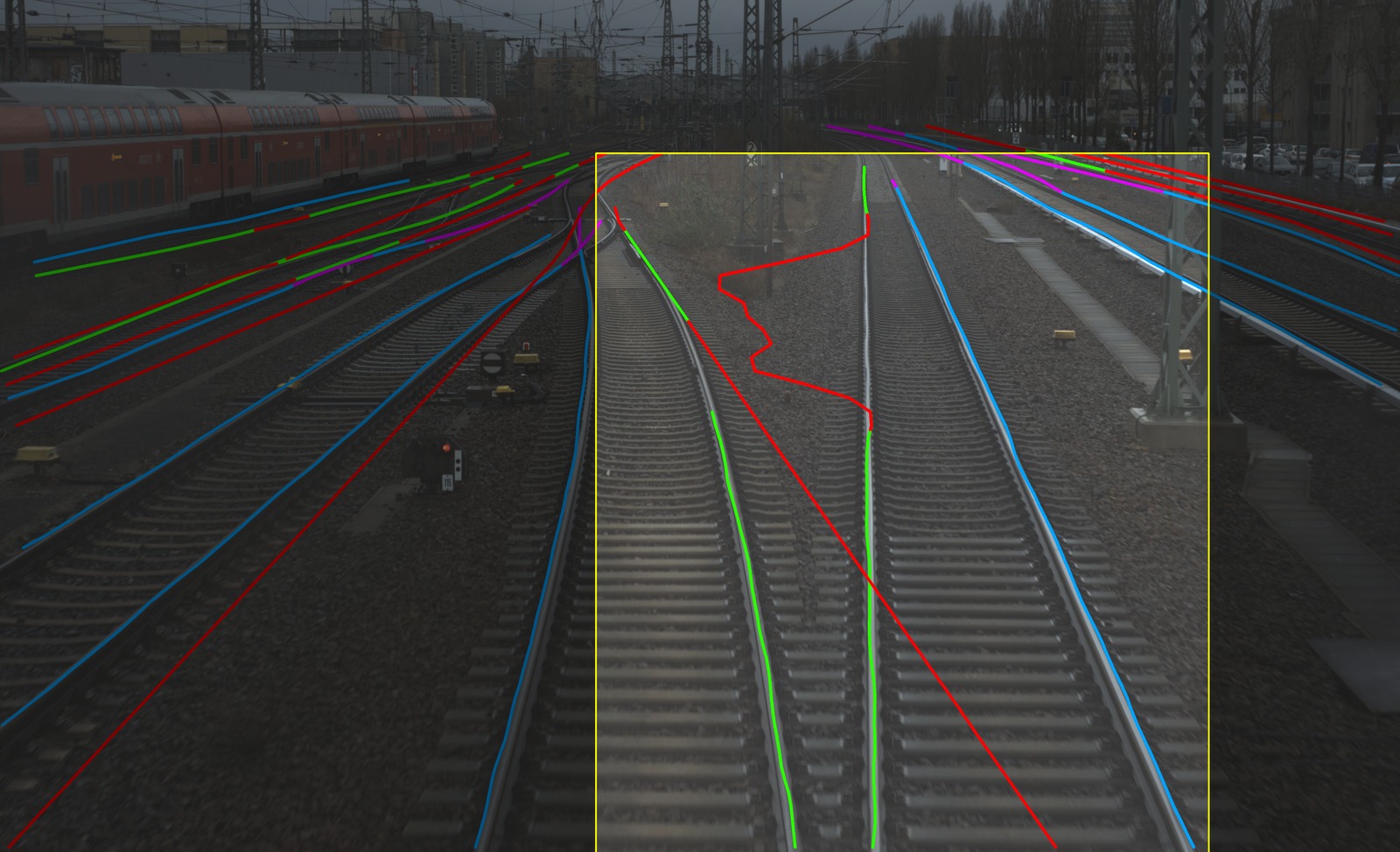} &
    \includegraphics[width=0.27\textwidth]{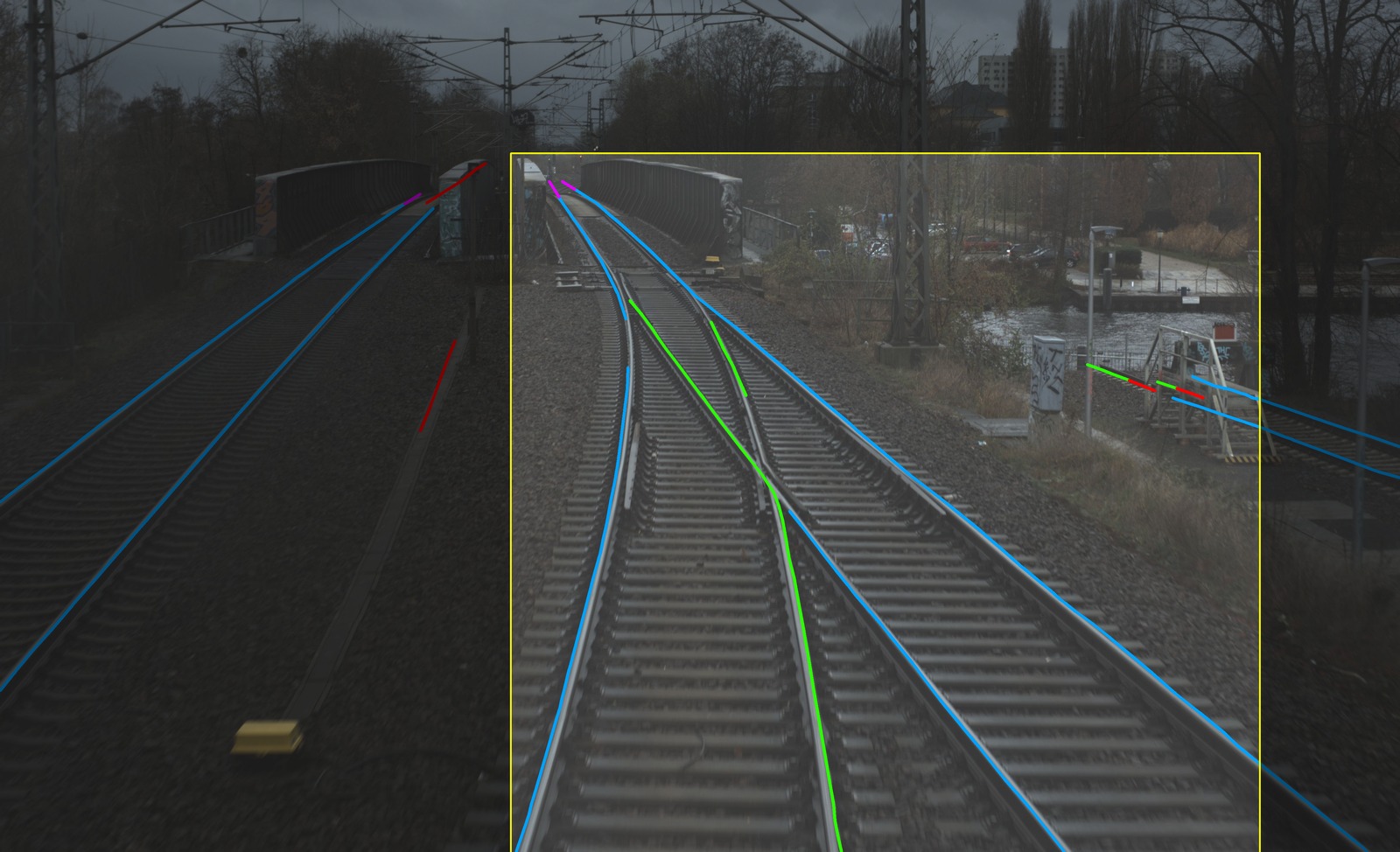} &
    \includegraphics[width=0.27\textwidth]{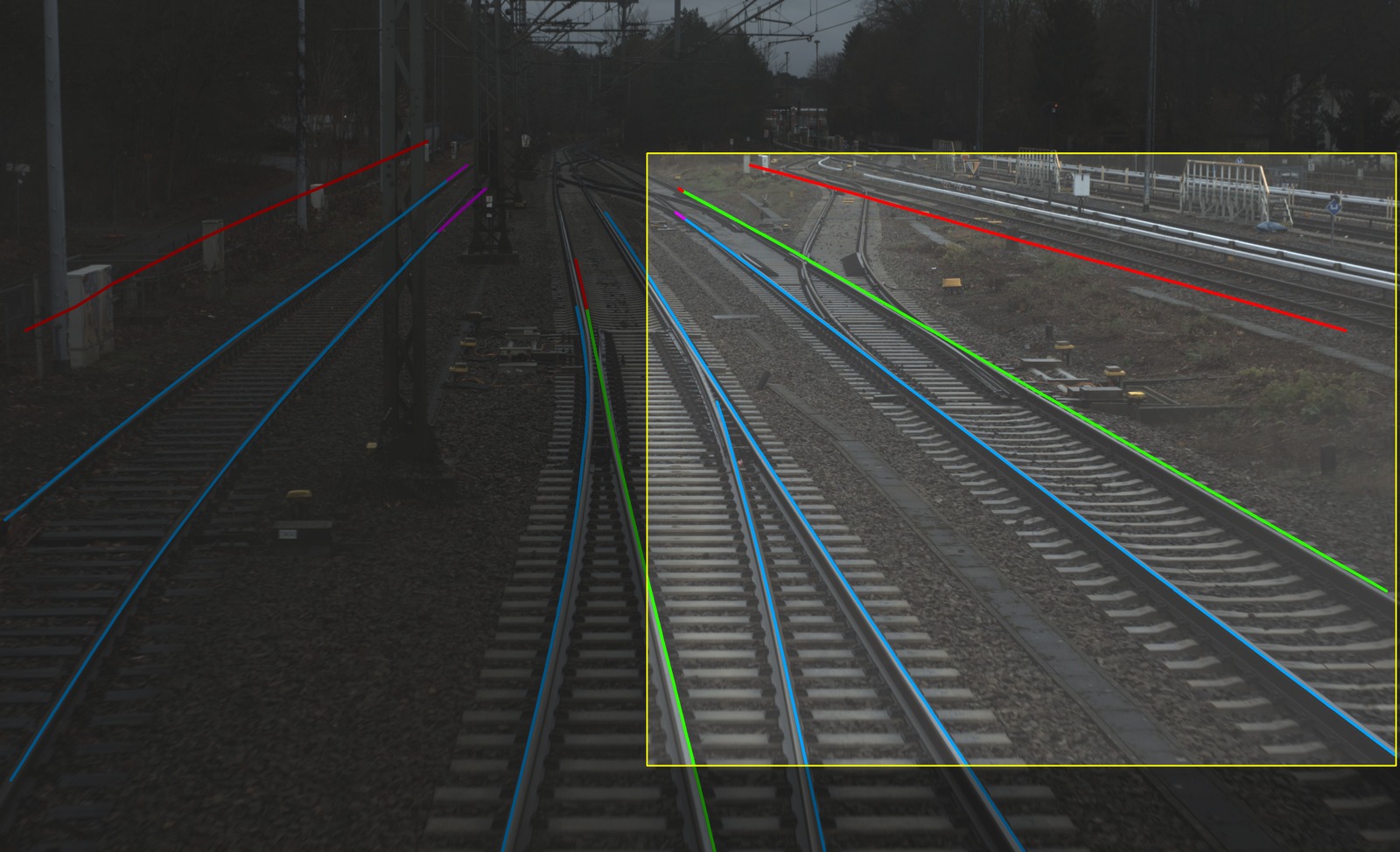} \\
    \rotatebox{90}{\parbox{2.7cm}{\centering\small\textbf{hard}}} &
    \includegraphics[width=0.27\textwidth]{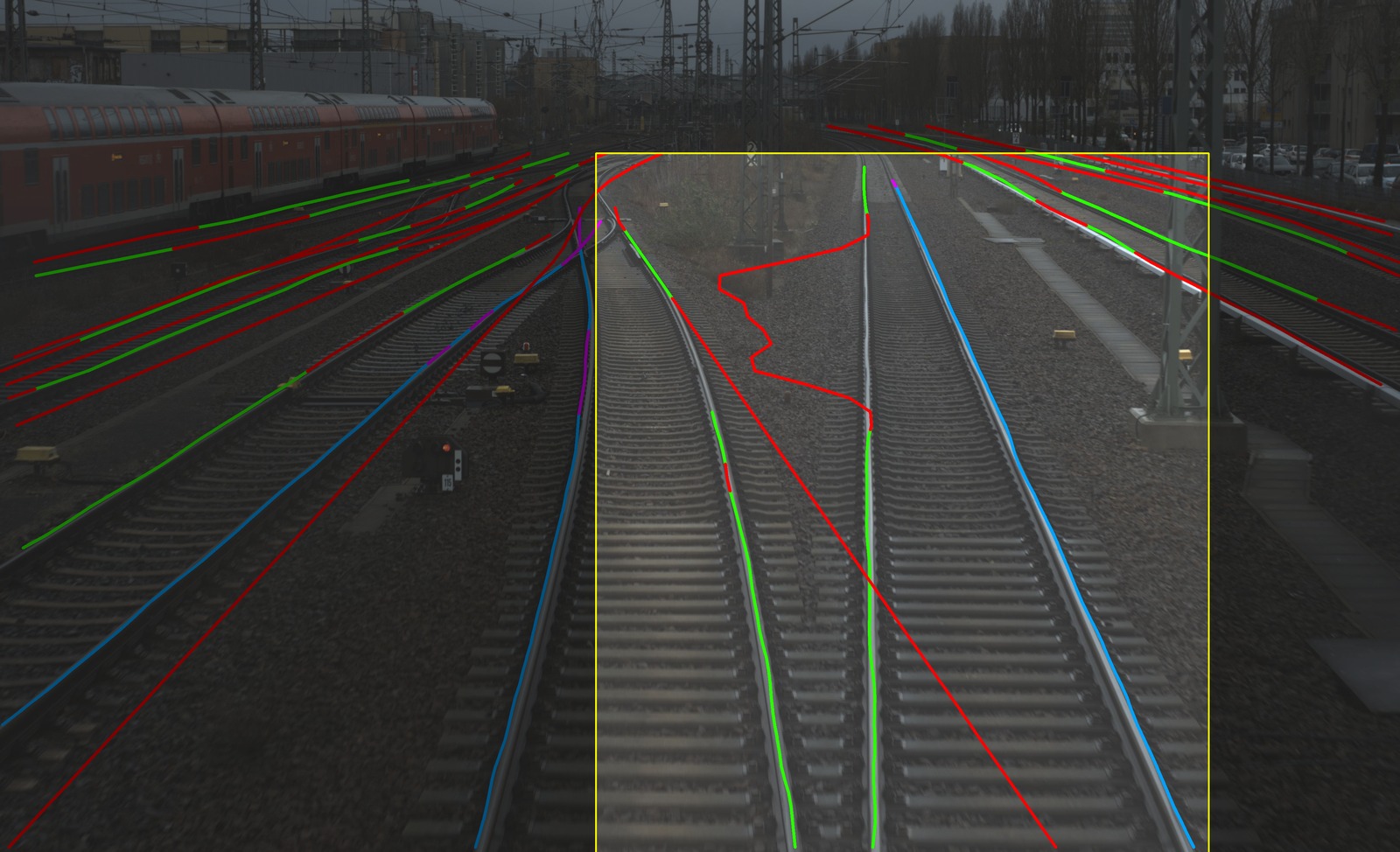} &
    \includegraphics[width=0.27\textwidth]{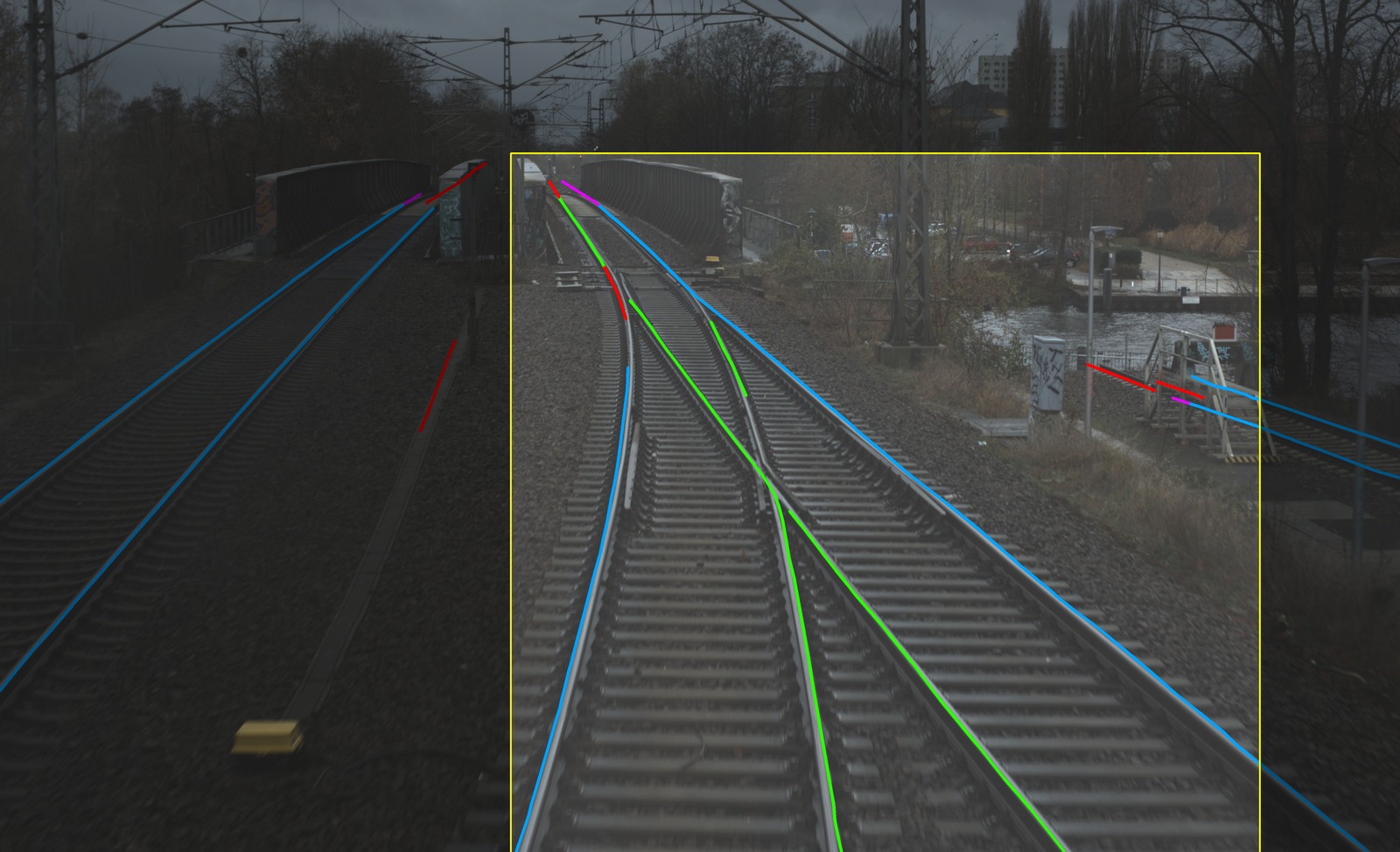} &
    \includegraphics[width=0.27\textwidth]{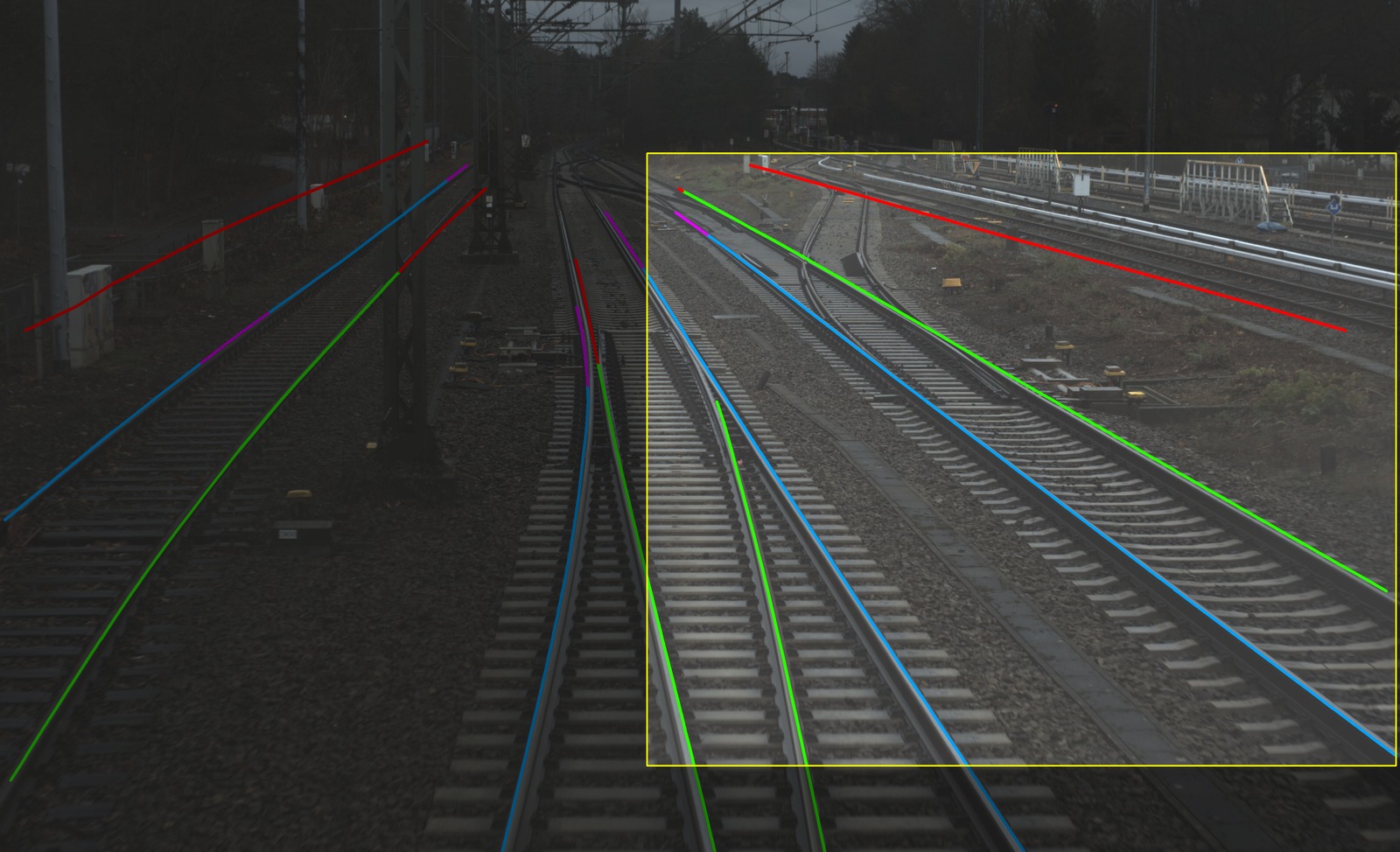} &
    \metriclegend{CULane}
  \end{tabular}
  \caption{Ground truth (first row) and predicted rails (rows 2--5) from finetuned PINet, YOLinO, and CLRNet, evaluated with \textbf{LineAP} and \textbf{CULane} in three difficulty levels: easy (LineAP@20 vs. CULane@5), moderate (LineAP@7 vs. CULane@1), hard (LineAP@3 vs. CULane@0.5). 
  \textcolor[HTML]{00a2ff}{Blue}: both metrics agree that the prediction is correct.
  \textcolor[HTML]{ff0000}{Red}: both metrics agree that the prediction is incorrect.
  \textcolor[HTML]{2bff00}{Green}: identified as correct by LineAP only.
  \textcolor[HTML]{ff04ea}{Pink}: identified as correct by CULane only.
  Best viewed in color.}
  \label{fig:metrics-lineap-vs-culane}
\end{figure*}

\begin{figure*}[!tb]
  \centering
  \setlength{\tabcolsep}{2pt}
  \renewcommand{\arraystretch}{0}
  \begin{tabular}{ccccc}
    & \multicolumn{1}{c}{\small\textbf{PINet (finetuned)}} &
    \multicolumn{1}{c}{\small\textbf{YOLinO (finetuned)}} &
    \multicolumn{1}{c}{\small\textbf{CLRNet (finetuned)}} \\[4pt]
    \rotatebox{90}{\parbox{2.7cm}{\centering\small\textbf{Ground Truth}\\\footnotesize(random colors)}} &
    \includegraphics[width=0.27\textwidth]{images/Metrics/gt/img_2013_test.jpg} &
    \includegraphics[width=0.27\textwidth]{images/Metrics/gt/img_2020_test.jpg} &
    \includegraphics[width=0.27\textwidth]{images/Metrics/gt/img_2004_test.jpg} \\
    \rotatebox{90}{\parbox{2.7cm}{\centering\small\textbf{Predictions}\\\footnotesize(random colors)}} &
    \includegraphics[width=0.27\textwidth]{images/Metrics/preds/img_2013_test_pinet_finetuned.jpg} &
    \includegraphics[width=0.27\textwidth]{images/Metrics/preds/img_2020_test_yolino_finetuned.jpg} &
    \includegraphics[width=0.27\textwidth]{images/Metrics/preds/img_2004_test_clrnet_finetuned.jpg} \\[4pt]
    \rotatebox{90}{\parbox{2.7cm}{\centering\small\textbf{easy}}} &
    \includegraphics[width=0.27\textwidth]{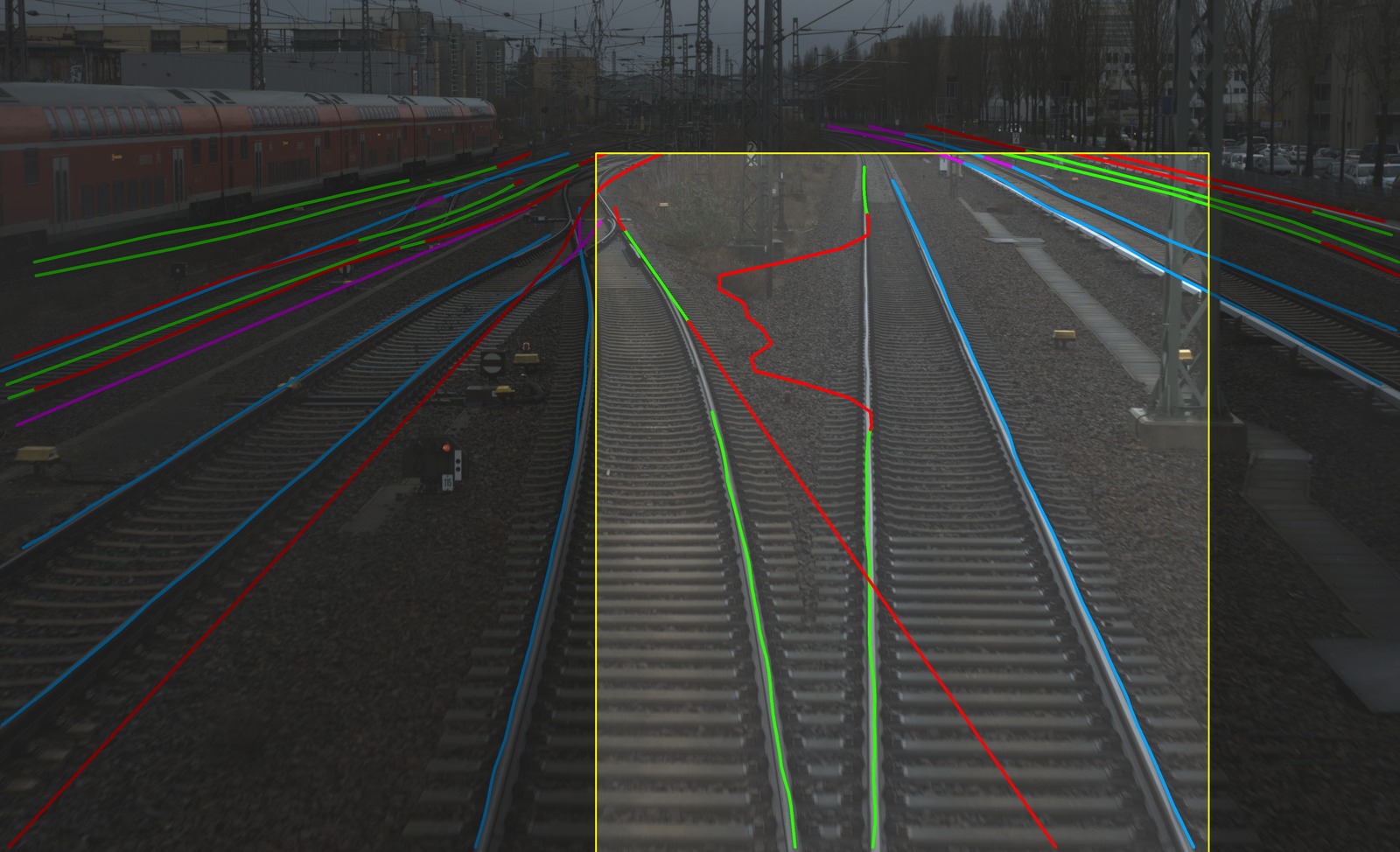} &
    \includegraphics[width=0.27\textwidth]{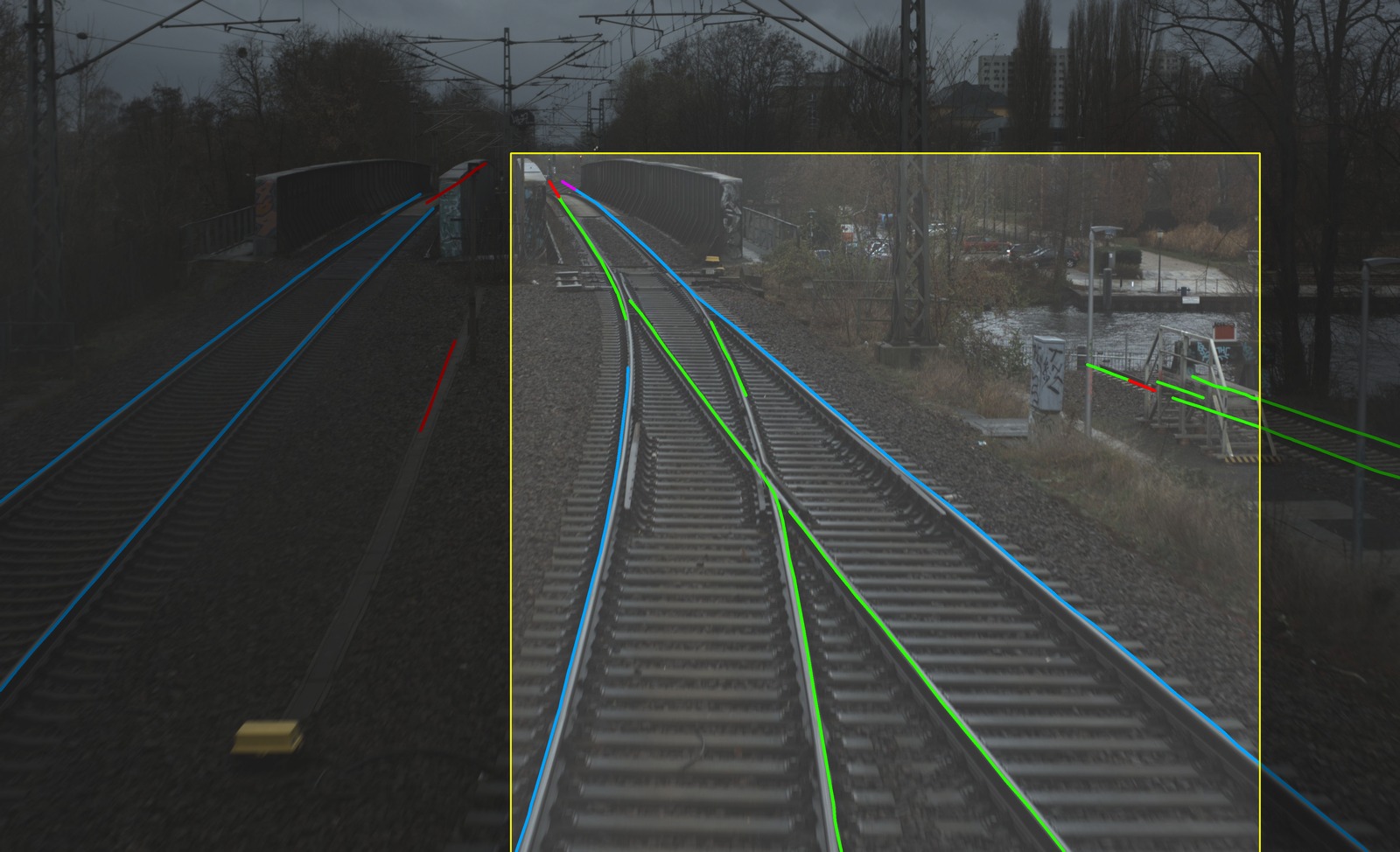} &
    \includegraphics[width=0.27\textwidth]{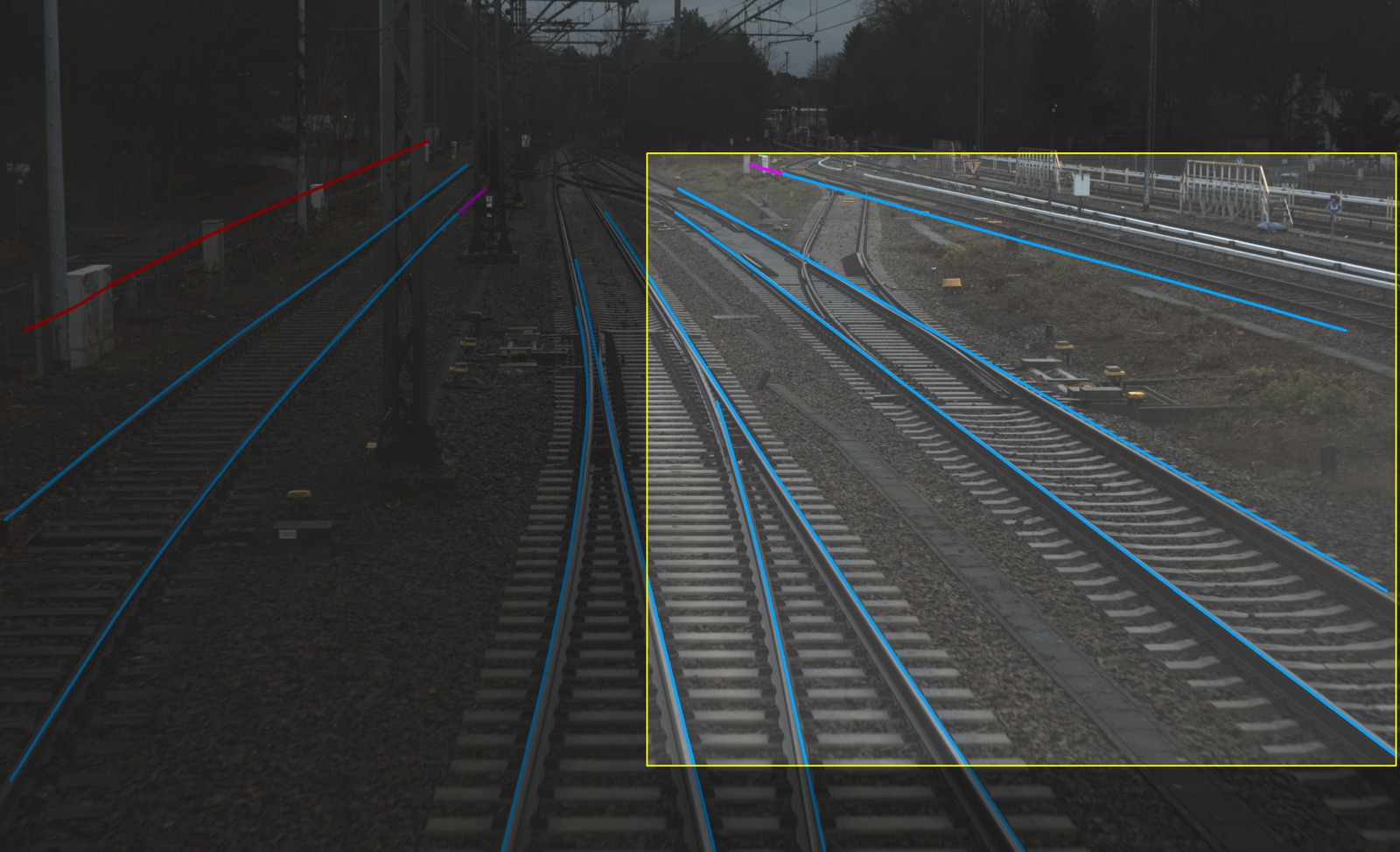} \\
    \rotatebox{90}{\parbox{2.7cm}{\centering\small\textbf{moderate}}} &
    \includegraphics[width=0.27\textwidth]{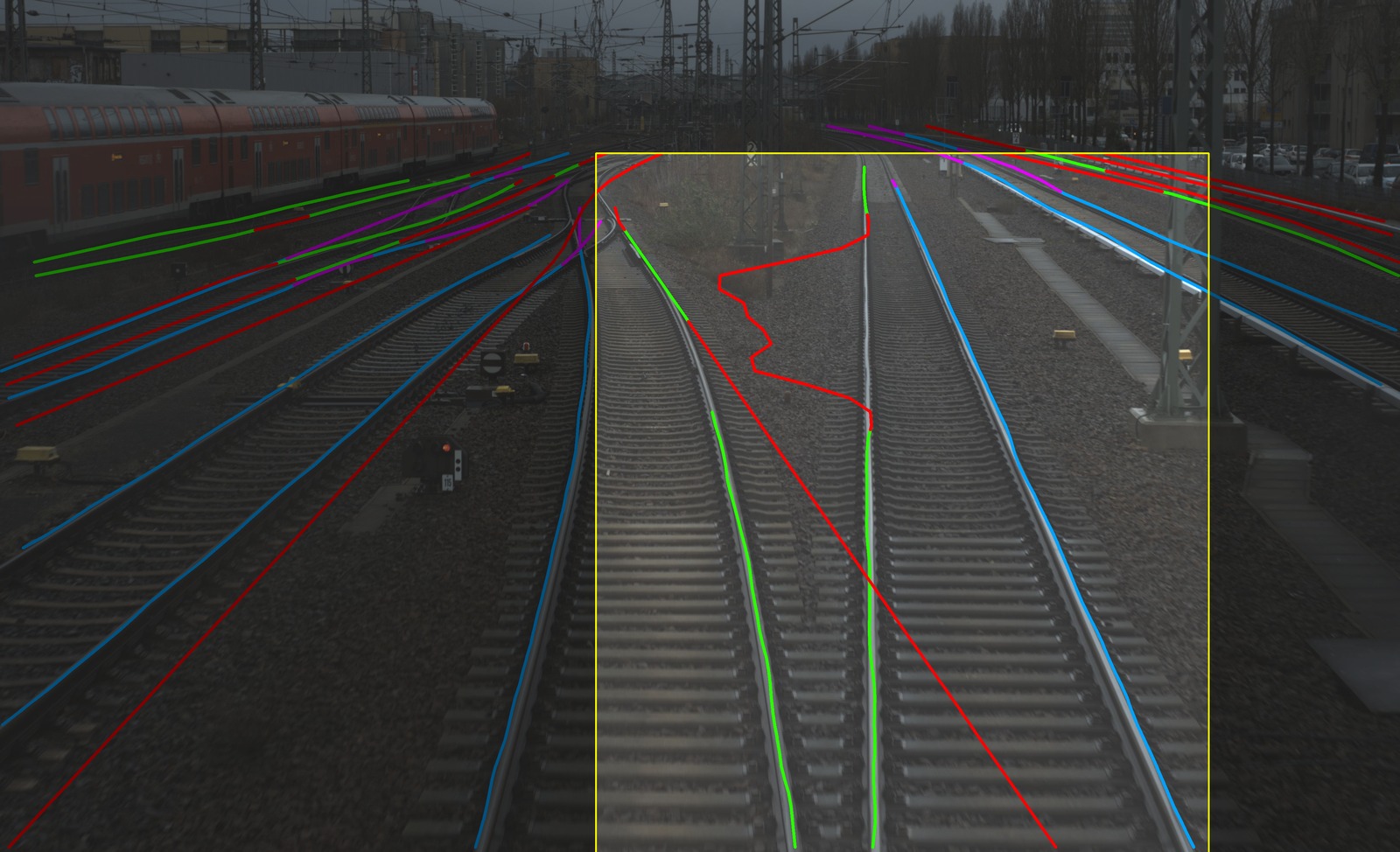} &
    \includegraphics[width=0.27\textwidth]{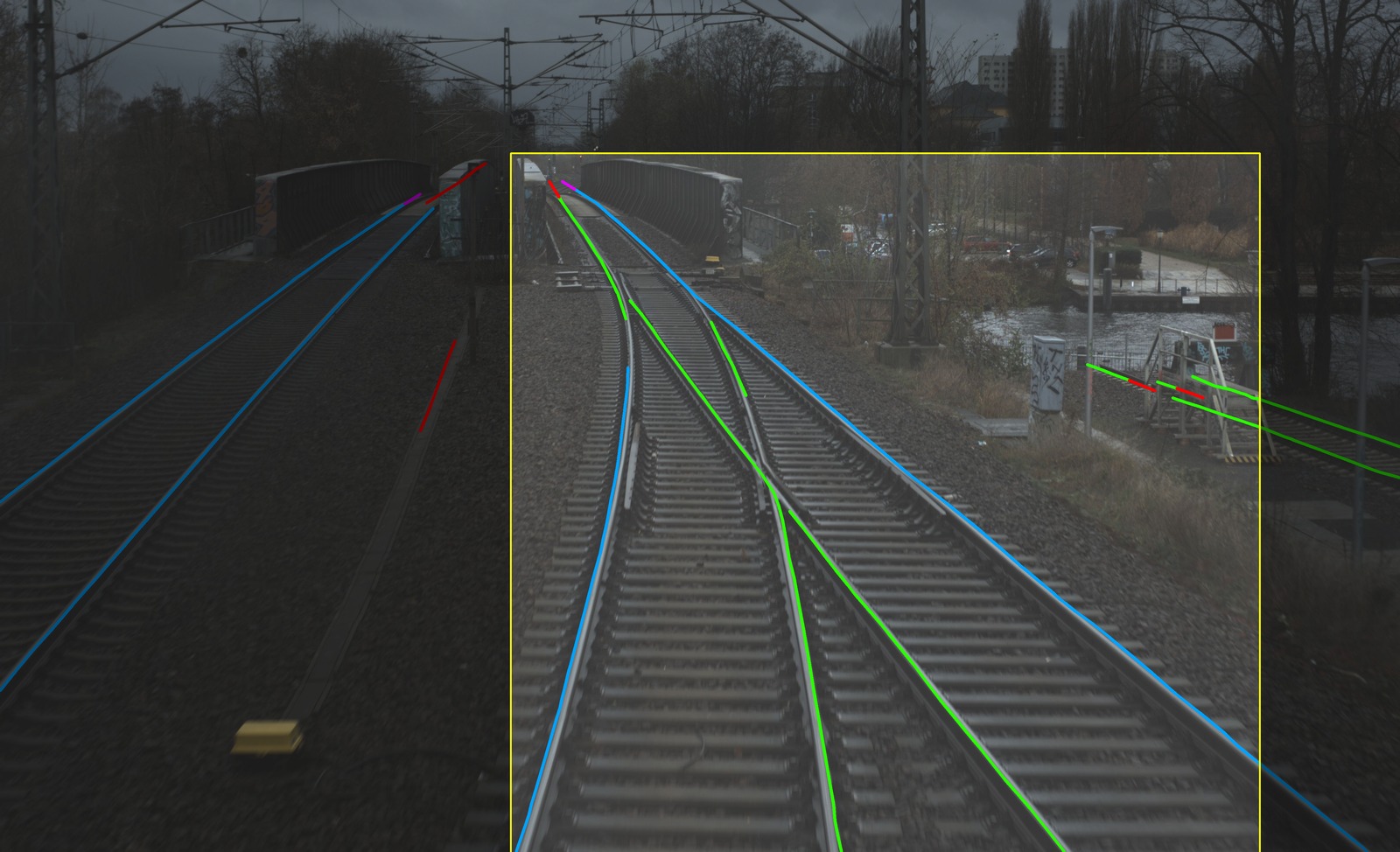} &
    \includegraphics[width=0.27\textwidth]{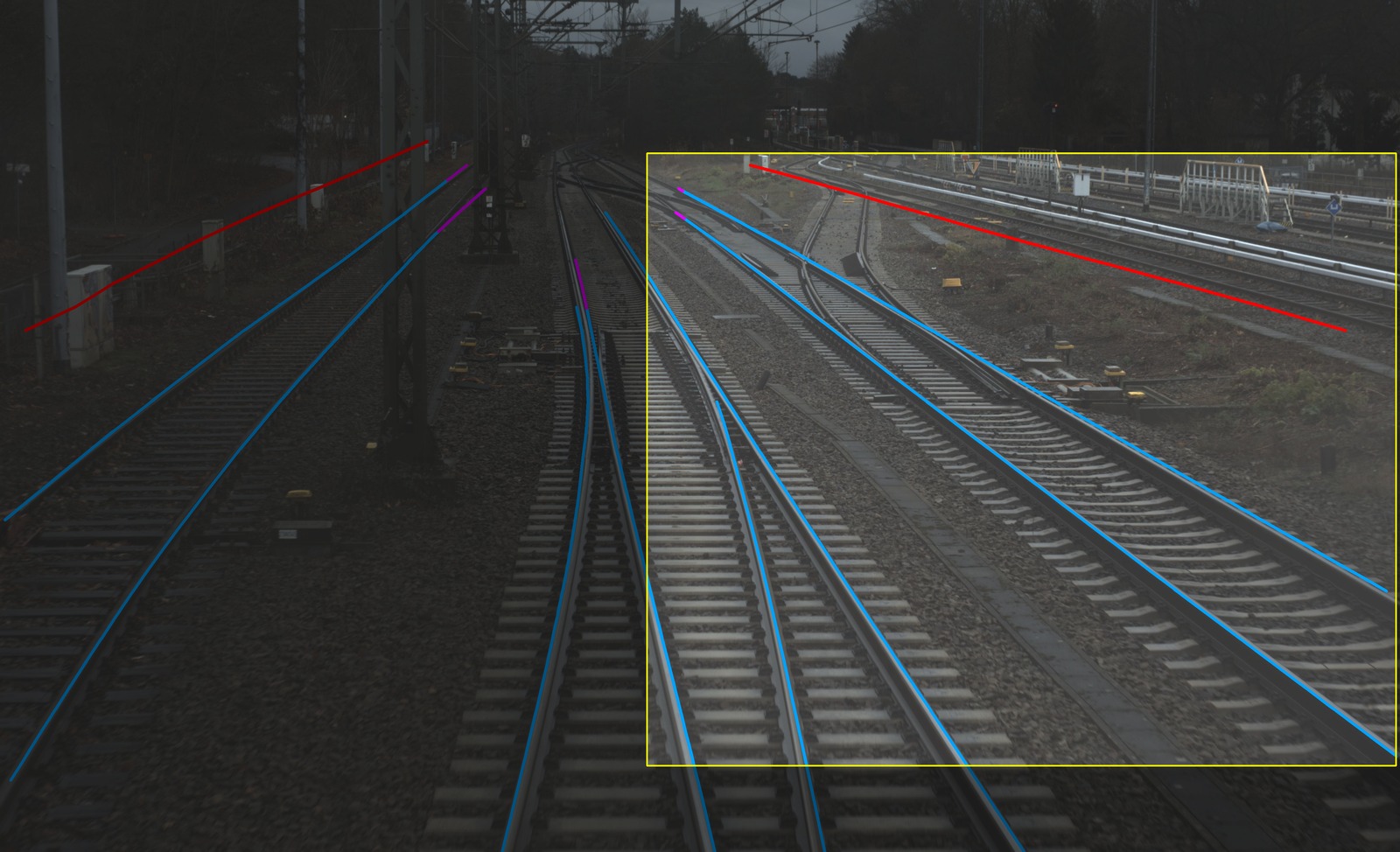} \\
    \rotatebox{90}{\parbox{2.7cm}{\centering\small\textbf{hard}}} &
    \includegraphics[width=0.27\textwidth]{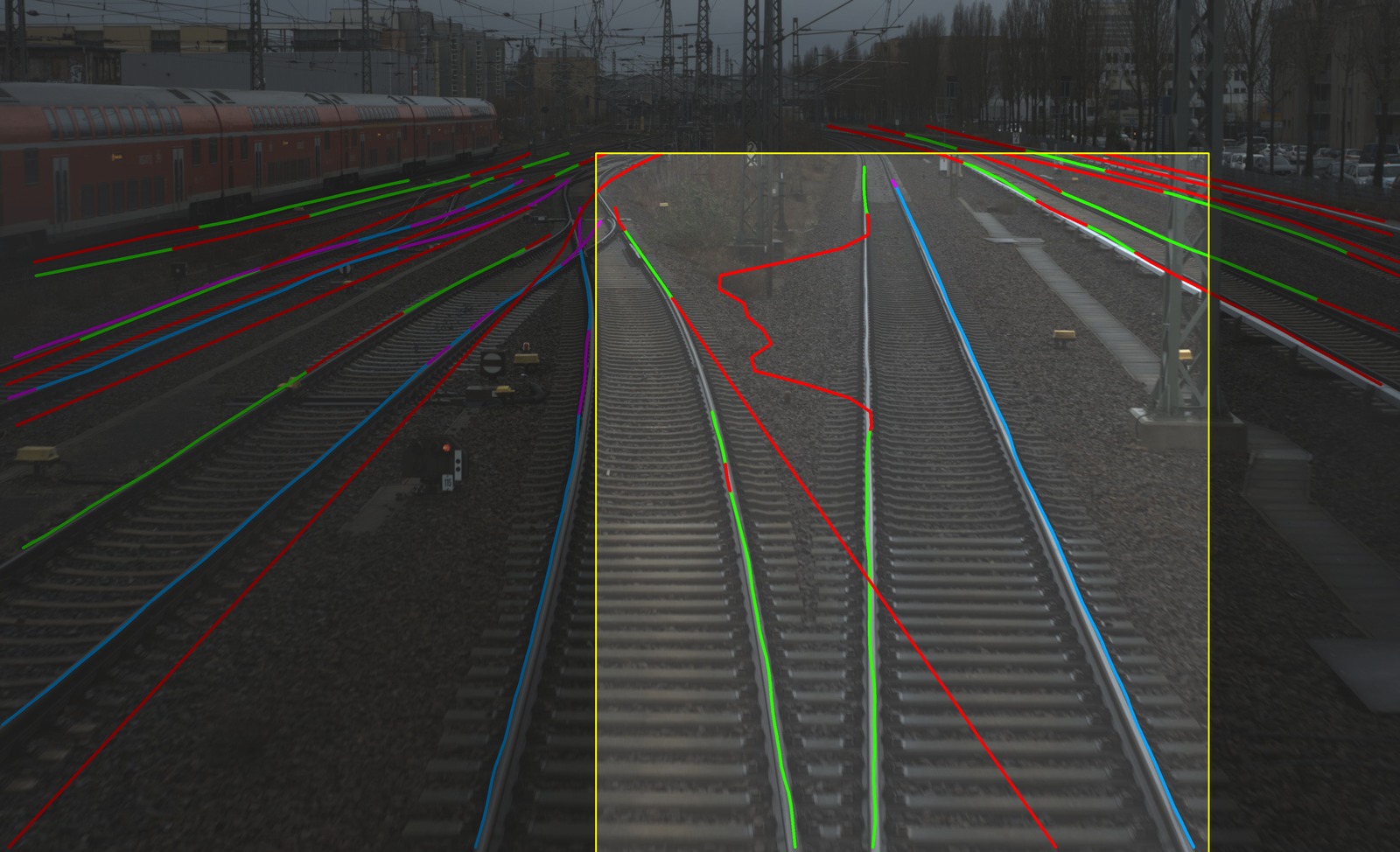} &
    \includegraphics[width=0.27\textwidth]{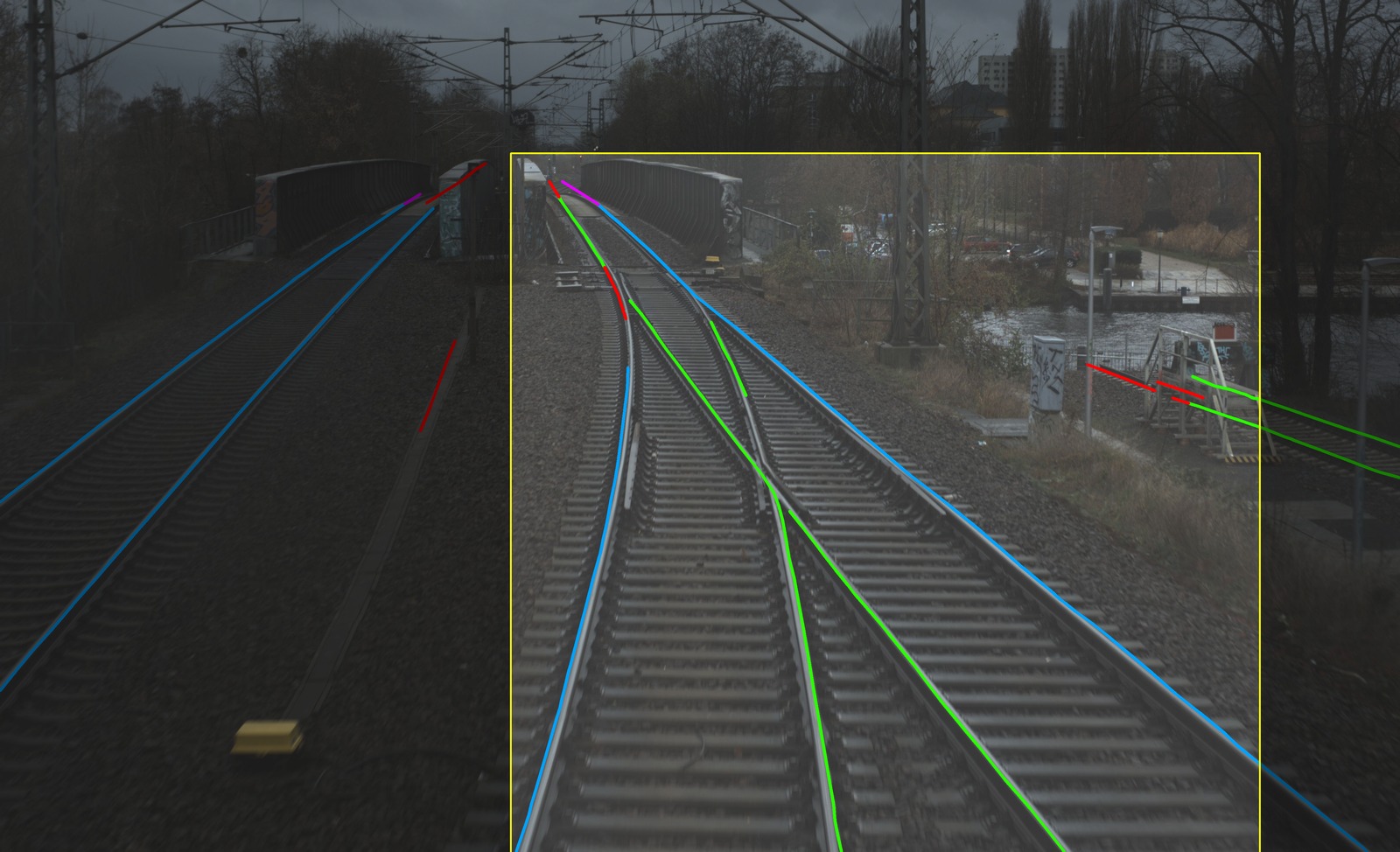} &
    \includegraphics[width=0.27\textwidth]{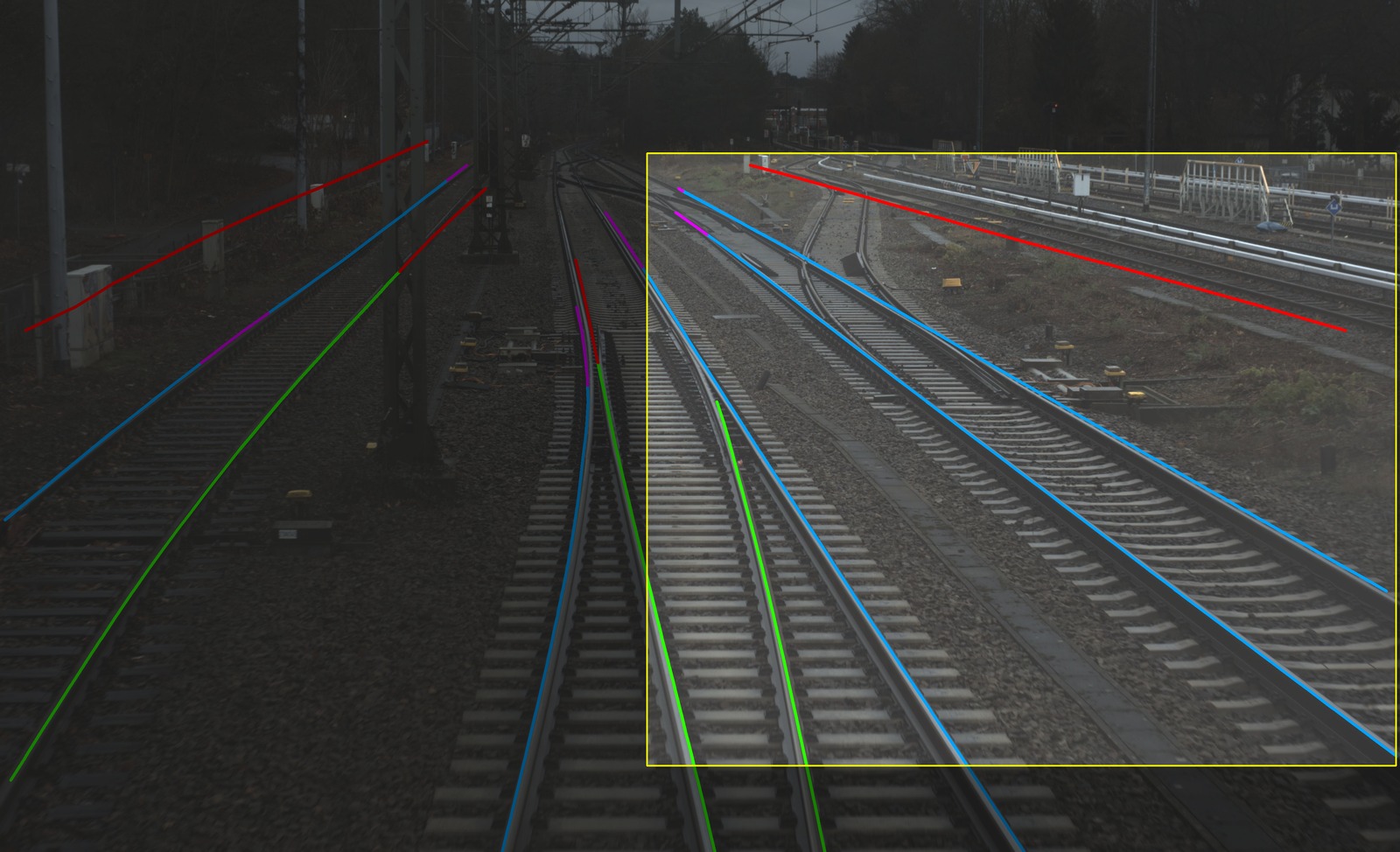} &
    \metriclegend{TuSimple}
  \end{tabular}
  \caption{Ground truth (first row) and predicted rails (rows 2--5) from finetuned PINet, YOLinO, and CLRNet, evaluated with \textbf{LineAP} and \textbf{TuSimple} in three difficulty levels: easy (LineAP@20 vs. TuSimple@2), moderate (LineAP@7 vs. TuSimple@0.7), hard (LineAP@3 vs. TuSimple@0.3). 
  \textcolor[HTML]{00a2ff}{Blue}: both metrics agree that the prediction is correct.
  \textcolor[HTML]{ff0000}{Red}: both metrics agree that the prediction is incorrect.
  \textcolor[HTML]{2bff00}{Green}: identified as correct by LineAP only.
  \textcolor[HTML]{ff04ea}{Pink}: identified as correct by TuSimple only.
  Best viewed in color.}
  \label{fig:metrics-lineap-vs-tusimple}
\end{figure*}

The qualitative results in Figures~\ref{fig:metrics-lineap-vs-culane} and~\ref{fig:metrics-lineap-vs-tusimple} show that the CULane and TuSimple metrics -- similar to ChamferAP -- are not able to properly differentiate between partially correct or incomplete predictions and geometrically incorrect predictions where the predicted lines are in the wrong location or orientation.

\end{document}